\documentclass{article}

\usepackage{arxiv}

\usepackage[T1]{fontenc}    
\usepackage{hyperref}       
\usepackage{bm}
\usepackage{cite}
\usepackage{graphics,fancyhdr,graphicx,subfigure}
\usepackage{subfigure}
\usepackage{amsthm,amsmath,amsfonts,latexsym,amssymb}
\usepackage{lineno}
\usepackage[ruled,linesnumbered]{algorithm2e}
\usepackage{multirow,booktabs}
\usepackage{listings}
\usepackage[table]{xcolor}
\usepackage{lscape}
\usepackage{tabularx}

\DeclareMathOperator*{\argmin}{\arg\!\min}

\def\equationautorefname~#1\null{%
  Eq.~(#1)\null
  }
\def\subfigureautorefname~#1\null{%
  Fig.~#1\null
}

\definecolor{listinggray}{gray}{0.9}
\definecolor{lbcolor}{rgb}{0.9,0.9,0.9}
\definecolor{Darkgreen}{RGB}{0,100,0}

\title{Transfer learning enhanced physics informed neural network for phase-field modeling of fracture}

\author{
  Somdatta Goswami \\
  Institute of Structural Mechanics\\
  Bauhaus Universit{\"a}t-Weimar\\
  99423-Weimar, Germany \\
  \texttt{somdatta89@gmail.com} \\
   \And
  Cosmin Anitescu \\
  Institute of Structural Mechanics\\
  Bauhaus Universit{\"a}t-Weimar\\
  99423-Weimar, Germany \\
  \texttt{cosmin.anitescu@uni-weimar.de} \\
  \And
  Souvik Chakraborty \\
  Center for Informatics and Computational Science\\
  University of Notre Dame\\
  Notre Dame - 46556, U.S.A. \\
  \texttt{csouvik41@gmail.com} \\
  \And
  Timon Rabczuk \\
  Division of Computational Mechanics\\
  Ton Duc Thang University\\
  Ho Chi Minh City, Viet Nam\\
  \texttt{timon.rabczuk@uni-weimar.de} \\
}

\begin{document}
\maketitle

\begin{abstract}
In this work, we present a new physics informed neural network (PINN)
algorithm for solving brittle fracture problems.
While most of the PINN algorithms available in the literature minimize the residual of the 
governing partial differential equation, the proposed approach takes a different path by minimizing the variational energy of the system.
Additionally, we modify the neural network output
such that the boundary conditions associated with the problem are exactly satisfied.
Compared to conventional residual based PINN, the proposed approach has two major advantages.
First, the imposition of boundary conditions is relatively simpler and more robust. 
Second, the order of derivatives present in the functional form of the variational energy is of lower order than in the residual form used in conventional PINN  and hence, training the network is faster.
To compute the total variational energy of the system, an efficient scheme that takes as input a geometry described by spline based CAD model and employs Gauss quadrature rules for numerical integration has been proposed.
Moreover, we note that for obtaining the crack path, the proposed PINN has to be trained at each load/displacement step, which can potentially make the algorithm computationally inefficient.  
To address this issue, we propose to use the concept `transfer learning' wherein, instead of re-training the complete network, we only  re-train the network {\it partially} while keeping the weights and the biases corresponding to the other portions fixed.
With this setup, the computational efficiency of the proposed approach is significantly enhanced.
The proposed approach is used to solve four fracture mechanics problems.
For all the examples, results obtained using the proposed approach match closely
with the results available in the literature.
For the first two examples, we compare the results obtained using the proposed approach with the 
conventional residual based neural network results. For both the problems, the proposed approach is found to yield better accuracy compared to conventional residual based PINN algorithms.
\end{abstract}


\section{Introduction}
\label{sec:intro}
The artificial neural network (ANN) is a class of machine learning tools, which is inspired by the structure and behaviour of biological neural systems. 
Since its introduction, it has proven to be a powerful and versatile tool for problems where the solutions are not clearly known, or there has been insufficient information given for the relationship between the inputs and the output.
It is capable of determining complex linear and non-linear relationships. 
ANN recognizes the patterns in a series of input and output values and using the acquired `knowledge' it then predicts the unknown output values for a given set of input values. 
However, despite its excellent performance in the domain of image processing and computer science, there are two major issues when it comes to the application of ANN in the engineering fields.
First, in the course of analyzing complex engineering systems, data acquisition is often computationally expensive.
Consequently, we may have access to a limited amount of training samples (i.e., we work in the {\it small data} regime).
Secondly, and perhaps more importantly, ANN trained from training data cannot ensure that the physics of the problem will be satisfied. Therefore, the essential physical laws associated with an engineering problem may not be satisfied.

Coming to our rescue is the physics informed/constrained neural networks (PINN). 
Over the past two years, a number of studies on PINN can be found in the literature \cite{Raisi2017Physics, Raissi2018DeepHP, Raissi2018HiddenFM, Zhu2019physics, Geneva2019modeling, Karumuri2019simulator}. 
In these methods, the neural networks are trained to solve supervised learning problems while respecting any given law of physics described by general non-linear partial differential equations.
In this paper, we propose a new PINN algorithm for studying the growth and propagation of fracture in brittle materials.
The proposed approach differs from the existing PINNs on several aspects.
First, unlike most of the PINN algorithms available in the literature, we do not minimize the residual of the governing differential equations; instead, we propose to minimize the variational energy of the system.
While crack nucleation may depend on stress, the propagation of cracks requires a certain energy, the fracture energy that represents the surface energy of a solid. 
Hence, energy criteria are used in the study of fracture using the phase-field approach \cite{Bourdin2000,Sun2005}.
One major advantage of the proposed variational energy formulation resides in the fact that it requires derivatives one order lower than in the conventional residual minimization approach \cite{Weinan2018}. Hence, this approach is computationally efficient.

Secondly, in almost all the available PINN methods, either trapezoidal rule or Monte Carlo integration is used for computing the integral by sampling the domain with either randomly or uniformly spaced points.
In this setup, a large number of integration points are required to obtain accurate results.
This, in turn, increases the computational cost of the approach.
To address this issue, we utilize the Gauss-Legendre quadrature rules.
However, directly generating Gauss points within the whole domain is not efficient for integrating non-smooth functions, which are common in modeling fracture. Therefore, motivated from finite element analysis \cite{azevedo2006hybrid, zienkiewicz1977finite} and isogeometric analysis \cite{cottrell2009isogeometric, Miehe2010},
we divide the computational domain into a number of elements and then, the Gauss points are generated within each element.

Moreover, in almost all the PINN methods developed 
over the last two years, the boundary conditions 
are enforced by considering a boundary-loss term in
the loss function. To strike a balance between the 
boundary-loss and the residual loss function, a 
penalty parameter has to be introduced with the 
boundary-loss term. This approach has two major disadvantages.
First, the boundary terms and the energy/residual component for the interior are often conflicting in nature (as one increases the other decreases).
This makes the optimization problem difficult to solve.
Secondly, the penalty parameter in this approach has to be modulated manually.
This also complicates the optimization problem as the selection of proper penalty parameters is tedious and time-consuming.
To address this issue, we propose to modify the neural network output so that the boundary conditions are exactly satisfied. As a consequence, no component corresponding to the boundary loss is needed in the loss function of the proposed approach.
This significantly simplifies the objective function to be minimized.

To accurately model the geometry, we propose to use non-uniform rational basis spline (NURBS) patches, as it allows us to exactly model complex geometries. 
Fracture analysis exhibits varying material properties in a local zone. 
Subsequently, when trying to capture the local quantities of interest, more integration points are required around the crack path. The geometry is therefore refined using the quad/oct-tree subdivision scheme for two and three-dimensions, respectively. 

Lastly, we note that we intend to apply the developed framework for studying fracture propagation and growth and hence,
we need to train the neural network at every load/displacement step.
This potentially can make the algorithm computationally expensive.
To address this issue, we propose to use the concept of `transfer learning' wherein, second step onward, we only retrain the weights and biases associated with the last layer.
The weights and biases corresponding to the other layers are kept fixed at previously trained values.
With this setup, the training phase is significantly accelerated.
Moreover, 
because of the increased robustness of neural network optimization algorithms, 
a larger load/displacement increment can be used for obtaining the 
crack path.
This also contributes towards obtaining the crack path in an efficient manner.

The novelty of this work is two-fold. First, as discussed above, an enhanced PINN is proposed in this work. Second, in this work, we have used the developed PINN for studying the growth and propagation of fracture. To the best of our knowledge, this is the first instance where PINN has been used for solving fracture growth and propagation problem.

The remainder of the paper is organized as follows. In \autoref{sec:concept_fracture}, we discuss the problem statement for phase-field modeling of brittle fracture using PINN. The details of the proposed approach are presented in \autoref{sec:PINN}.
Implementation of the proposed approach for solving fracture mechanics problems using phase field method is discussed in \autoref{sec:phase_pinn}.
The concept and implementation of transfer learning are also discussed in this section.
Numerical examples illustrating the performance of the proposed approach are presented in \autoref{sec:numericals}. Finally, \autoref{sec:conclusion} presents the concluding remarks and possibilities future work.

\section{Phase-field modeling for fracture}
\label{sec:concept_fracture}
Phase-field modeling is an effective way to model fracture by assuming the process zone has a finite width which is controlled by a length scale parameter ($l_0$). 
A sharp crack topology is recovered in the limit as $l_0 \to 0$ \cite{Bourdin2000}. In this approach, the effects associated with crack formation such as stress release are incorporated into the constitutive model. 
A continuous scalar parameter ($\phi$) is used to track the fracture pattern. The cracked region is represented by $\phi = 1$ while the undamaged portion is given by $\phi = 0$. 
Modeling fracture using the phase-field method involves the solving for the vector-valued elastic field, $\bm{u}$ and the scalar-valued phase-field, $\phi$. The equilibrium equation for the elastic field for an isotropic model, considering the evolution of crack, can be written as:
\begin{equation}\label{eq:degradaed_Disp_eq}
    -\nabla\cdot g(\phi)\bm{\sigma} = \bm{f} \text{ on } \Omega,
\end{equation}
where $\bm{\sigma}$ is the Cauchy stress tensor, $\bm{f}$ is the body force and $g(\phi)$ represents the monotonically decreasing stress-degradation function. A common form of the degradation function, as used in the literature, for isotropic solids is \cite{Miehe2010}:
\begin{equation}\label{eq:degradation_func}
    g(\phi) = (1-\phi)^2.
\end{equation}
The elastic field is constrained by Dirichlet and Neumann boundary conditions: 
\begin{equation}\label{eq:Disp_boundary}
    \begin {split}
        g(\phi)\bm{\sigma}\cdot \bm{n} &= \bm{t}_N \text{ on } \partial\Omega_{N} \\
        \bm{u} &= \bm{\overline u} \text{ on } \partial \Omega_{D},\\
    \end{split}
\end{equation}
where $\bm{t}_N$ is the prescribed boundary forces and $\bm{\overline u}$ is the prescribed displacement for each load step. The Dirichlet and Neumann boundaries are represented by $\partial\Omega_{D}$ and $\partial\Omega_{N}$, respectively. 

On the other hand, the governing equation for the phase-field is written as:
\begin{equation}\label{eq:phasefield_eq}
       \frac{G_c}{l_0}\phi - G_{c}l_{0}\nabla^{2}\phi = -g'(\phi)H(\bm{x},t) \text{ on } \Omega,
\end{equation}
where $G_c$  represents the critical energy release rate (property of material) and $H(\bm{x},t)$ is the strain-history function. With the evolving damage, only the tensile component of the principal stress degrades while the compressive component remains unaffected \cite{Miehe2010a}. Hence, the strain energy functional is decomposed into the tensile ($\Psi^{+}_{0}$) and compressive ($\Psi^{-}_{0}$) components as: 
\begin{equation}\label{eq:strain_decom}
    \Psi({\bm{\epsilon}}) = \Psi^{+}_{0}(\bm{\epsilon}) + \Psi^{-}_{0}(\bm{\epsilon}),
\end{equation}
where
\begin{equation}\label{eq:decom_def}
    {\Psi^{\pm}_0}(\bm{\epsilon}) = \frac{\lambda}{2}\left\langle {\text{tr}(\bm{\epsilon})}\right\rangle ^{2}_{\pm}  + \mu \text{tr}(\bm{\epsilon}^2_{\pm}).
\end{equation}
tr$(\cdot)$ in \autoref{eq:decom_def} denotes the trace of the tensor and $\lambda$ and $\mu$ are the Lam\'e constants.
$H(\bm{x},t)$ contains the maximum positive tensile energy in the history of deformation of the system and is defined as:
\begin{equation}\label{eq:history_field}
    H(\bm{x},t) = {\max_{s \in [0,t]}}\Psi^{+}_{0}(\bm{\epsilon}(\bm{x},s)),
\end{equation}
where $\bm{x}$ is the integration point. The strain-history functional ensures monotonically increasing values of $\phi$ and prevents the crack from healing \cite{Miehe2010}. 
The advantage of using the local history functional approach is that an initial history functional can be used to define initial cracks in the system \cite{Miehe2010a}. The initial strain-history function ($H(\bm{x},0)$) could be defined in terms of $d(\bm{x},l)$, which is the closest distance from any point ($\bm{x}$) on the domain to the line ($l$), which represents the discrete crack \cite{Borden2012}. In particular, we set
\begin{equation}\label{eq:initial_history_field}
    H(\bm{x},0) = \left\{ {\begin{array}{l l}
  {\frac{BG_c}{2l_0}(1 - \frac{2d(\bm{x},l)}{l_0})}&{d(\bm{x},l) \leqslant \frac{l_0}{2}} \\ 
  0&{d(\bm{x},l) > \frac{l_0}{2}}
\end{array}} \right.,
\end{equation}
where $B$ is a scalar parameter that controls the magnitude of the scalar history field and is calculated as:
\begin{equation}\label{eq:scalarB}
    B = \frac{1}{1-\phi}  \; \; \; \text{for} \;\;\phi < 1.
\end{equation}
The phase-field is assumed to satisfy homogeneous Neumann-type boundary conditions on the entire boundary:
\begin{equation}\label{eq:Phase_boundary}
    \nabla\phi\cdot \bm{n} = 0 \text{ on } \partial\Omega.
\end{equation}
The displacement field, $\bm u$ and the phase-field, $\phi$ can be computed by solving Eqs. (\ref{eq:degradaed_Disp_eq}) and (\ref{eq:phasefield_eq}) subjected to the boundary conditions defined in Eqs. (\ref{eq:Disp_boundary}) and (\ref{eq:Phase_boundary}).

In the energy method, the solution is obtained by minimization of the total variational energy of the system, $\mathcal{E}$ \cite{A.Griffith1921, Bourdin2000, Borden2014}. The problem statement can be written as:
\begin{equation}\label{eq:ProbStatement_energy}
\begin{split}
    \text{Minimize:}\;\;\;\; \mathcal{E} &= \Psi_e + \Psi_c,\\
    \text{subject to:}\;\;\;\; \bm{u} &= \bm{\overline u} \text{ on } \partial \Omega_{D},\\
\end{split}
\end{equation}
where $\Psi_e$ is the stored elastic strain energy, $\Psi_c$ is the fracture energy and $\bm{\overline u}$ is the prescribed displacement on the Dirichlet boundary, $\partial \Omega_{D}$. 
Using the variational approach, the traction-free Neumann boundary conditions are automatically satisfied. 
In \autoref{eq:ProbStatement_energy}, $\Psi_e$ and $\Psi_c$ are defined as:
\begin{equation}\label{eq:energyterms}
    \begin{split}
       \Psi_e &=  \int_{\Gamma}f_e(\bm{x}) d\Omega,\\
       \Psi_c &=  \int\limits_\Omega f_c(\bm{x}) d\Omega,\\
    \end{split}
\end{equation}
where
\begin{equation}\label{eq:indv_ent}
    \begin{split}
        f_e(\bm{x}) &=  g(\phi) \Psi_{0}^+(\bm{\epsilon}) + \Psi_{0}^-(\bm{\epsilon}),\\
       f_c(\bm{x}) &=  \frac{G_c}{2l_0} \left(\phi^2 +l_0^2 |\nabla\phi|^2 \right) + g(\phi)H(\bm{x},t),\\
    \end{split}
\end{equation}
where $g\left(\phi\right)$ and $\Psi_0^{\pm}$  are defined in Eqs. (\ref{eq:degradation_func}) and (\ref{eq:decom_def}) respectively. 
$G_c$, as already stated, is the critical energy release rate and $l_0$ is the length scale parameter.

The two fields could either be solved simultaneously using the monolithic-solution scheme \cite{Heister,Vignollet2014} or they could be solved one at a time using the staggered-solution scheme \cite{Miehe2010,Miehe2010a}. In the monolithic scheme, $\Psi_e$ and $\Psi_c$ are simultaneously minimized (by directly minimizing $\mathcal E$) to obtain the displacement field and the phase-field.
On the other hand, in the staggered scheme, we repeatedly cycle between the minimization of $\Psi_e$ and the minimization of $\Psi_c$ until a self-consistent solution is obtained.
However, there is no guarantee that the staggered scheme will achieve self-consistency.
Even if the solution converges, the number of cycles may be significantly large. Hence, the staggered solution scheme is computationally more expensive as compared to the monolithic scheme.

In this work, we only focus on the monolithic scheme. The objective is to develop a PINN for solving phase-field based brittle fracture problem using the monolithic scheme.

\section{Physics informed neural network}
\label{sec:PINN}
In this section, we first provide a brief description on the anatomy of deep neural network.
Thereafter, we discuss various components of the proposed PINN approach, which form a platform for their implementation in the proposed approach.

\subsection{Deep neural network architecture}
\label{subsec:dnn}
Deep neural networks are distinguished from the conventional shallow neural networks by the number of hidden layers present in the network.
In conventional shallow networks, we have one input layer, one output layer, and a hidden layer.
On the other hand, in a deep neural network, we have more than one hidden layers, the intuition being, more hidden layers will be more expressive and hence, will provide results that are more accurate.
In this work, we have used a deep, fully connected feed-forward neural network. 
Considering that the network consists of $L$ hidden layers, where the $0$-th layer denotes the input layer and $(L+1)$-th layer is the output layer, the 
weighted input, $z^l_i$ into a $i$-th neuron on layer, $l$, is a function of weight, $W^l_{ij}$ and bias, $b^{l-1}_j$ and is represented as:
\begin{equation}\label{eq:weighted_input}
    z^l_i =  \sigma_{l-1}\left(\sum_{j=1}^{m_{l-1}}\left(W^l_{i,j}(z^{l-1}_j) + b^l_i\right)\right),
\end{equation}
where $\sigma_{l-1}\left( \cdot \right)$ denotes the activation function in layer $l$ and $m_{l-1}$ are the number of neurons in the layer $l-1$.
From the above concepts, the feed-forward algorithm for computing the output, $\bm{Y}^L$ is expressed as:
\begin{equation}\label{eq:feedforward}
\begin{split}
       \bm{Y}^L &= \sigma_L(\mathbf W^{L+1}\bm{z}^L + \bm b^L),\\
       \bm{z}^L & = \sigma_{L-1}\left(\mathbf{W}^{L}\bm{z}^{L-1} + \bm b^L\right),\\
       \bm{z}^{L-1} & = \sigma_{L-2}\left(\mathbf{W}^{L-1}\bm{z}^{L-2} + \bm b^{L-1}\right),\\
        &\vdots\\
        \bm z^1 &= \sigma_0\left(\mathbf W^1\bm{x} + \bm b^1\right),\\
\end{split}
\end{equation}
where $\bm{x}$ is the input of the neural network. 
\autoref{eq:feedforward} can be represented in a compressed form as $Y = \mathcal N (x;\bm{\theta})$, where $\bm{\theta} = \left(\mathbf W, \bm b \right)$ includes both the weights and biases of the neural network, $\mathcal N$. 
For putting the neural network to use, we need to learn the weights, $W^l_{ij}$ and biases, $b^{l}_j$.
Conventionally, this is achieved by first collecting data, $\mathcal D = \left\{\bm x, \bm Y \right\}_{i=1}^{N_t}$, and then minimizing a loss-function.
Common loss-functions used in literature includes the $l_2$-loss function and the $l_1$-loss function \cite{Rojas1996neural}. 
\begin{equation}\label{eq:L1_loss}
\begin{split}
    l_1 &= \sum_{i =1}^n\left| Y_i- \mathcal N(x_i; \bm{\theta})\right|,\\
    l_2 &= \sum_{i =1}^n\left(Y_i- \mathcal N (x_i; \bm{\theta})\right)^2,\\
\end{split}
\end{equation}
where $Y_i$ and $\mathcal N (x_i; \bm{\theta})$ are the target value and the corresponding predicted value, respectively and $x_i$ denotes the sample point.
One primary bottleneck of neural networks (both deep and shallow) rests in the fact that a large number of training data is required.
Unfortunately, in engineering problems, data collection either by using numerical or physical experiments is often expensive and time-consuming.
Moreover, engineering systems are governed by certain physical laws, and with neural networks (or in fact, most data-driven techniques) it cannot be guaranteed that these physical laws will be satisfied.

Next, we present a PINN algorithm
where we compute the weights and biases associated with the neural networks based on the physics of the problem (defined by a non-linear partial differential equation).

\subsection{Physics informed neural network -- An energy based approach}
\label{subsec:pinn}
Without loss of generality, we consider the physics of a problem is defined by a generic one dimensional time-independent differential equation of the form:
\begin{subequations}\label{eq:ps}
\begin{equation}\label{eq:generic_de}
    \mathcal F \left(u,u_x,\ldots, u_{x\cdots x}, x, f\left(x\right) \right) = 0,
\end{equation}
\vspace*{-\baselineskip}
\begin{equation}\label{eq:gen_bc}
    u\left(x_D\right) = u_D,
\end{equation}
\end{subequations}
where $x_D$ represents a Dirichlet boundary point, $u$ represents the dependent variable to be computed, $u_x$, $u_{xx}$ and $u_{x\cdots x}$
represents the first order, second order and higher order derivative with respect to the independent variable $x$. In 
\autoref{eq:generic_de}, $f\left(x\right)$ represents the source term and 
\autoref{eq:gen_bc} represents the Dirichlet boundary condition.
Since the method we are about to propose is based on the energy principle,
the homogenous Neumann boundary conditions are automatically satisfied.

Let us assume the variational energy of \autoref{eq:generic_de} is represented as
\begin{equation}\label{eq:tot_energy}
    \mathcal V_e = \int_{\Omega}\mathcal G \left(u,u_x,\ldots, u_{x\cdots x}, x, f\left(x\right) \right) d\Omega,
\end{equation}
where $\Omega$ represents the problem domain and $\mathcal G$ is a differentiable functional.
With this, solution to \autoref{eq:generic_de} can be obtained by solving the following optimization problem:
\begin{equation}\label{eq:opt_ve}
\begin{split}
    & u^* = \argmin_{u} \mathcal V_e \left(u \right)\\
    \text{subject to  }\;\;\;\;\;\;\;\;\;\; & u\left(x_D\right) = u_D
\end{split}
\end{equation}
In the proposed PINN approach, we utilize a similar concept as discussed in \autoref{eq:opt_ve}.

\begin{figure}[htbp!]
    \centering
    \subfigure[Proposed PINN]{
    \includegraphics[width = 0.95\textwidth]{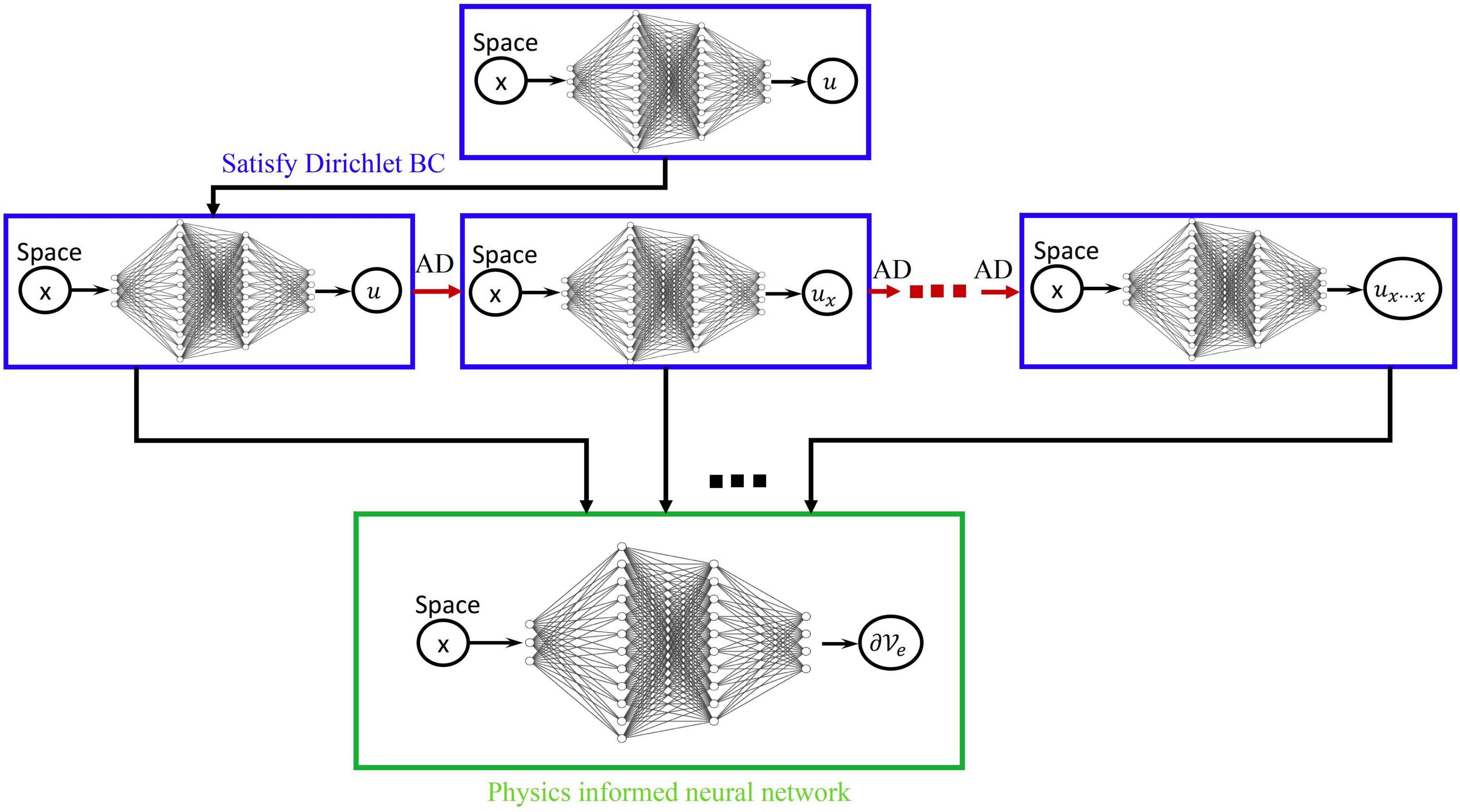}}
    \subfigure[Computational graph]{
    \includegraphics[width = 0.75\textwidth]{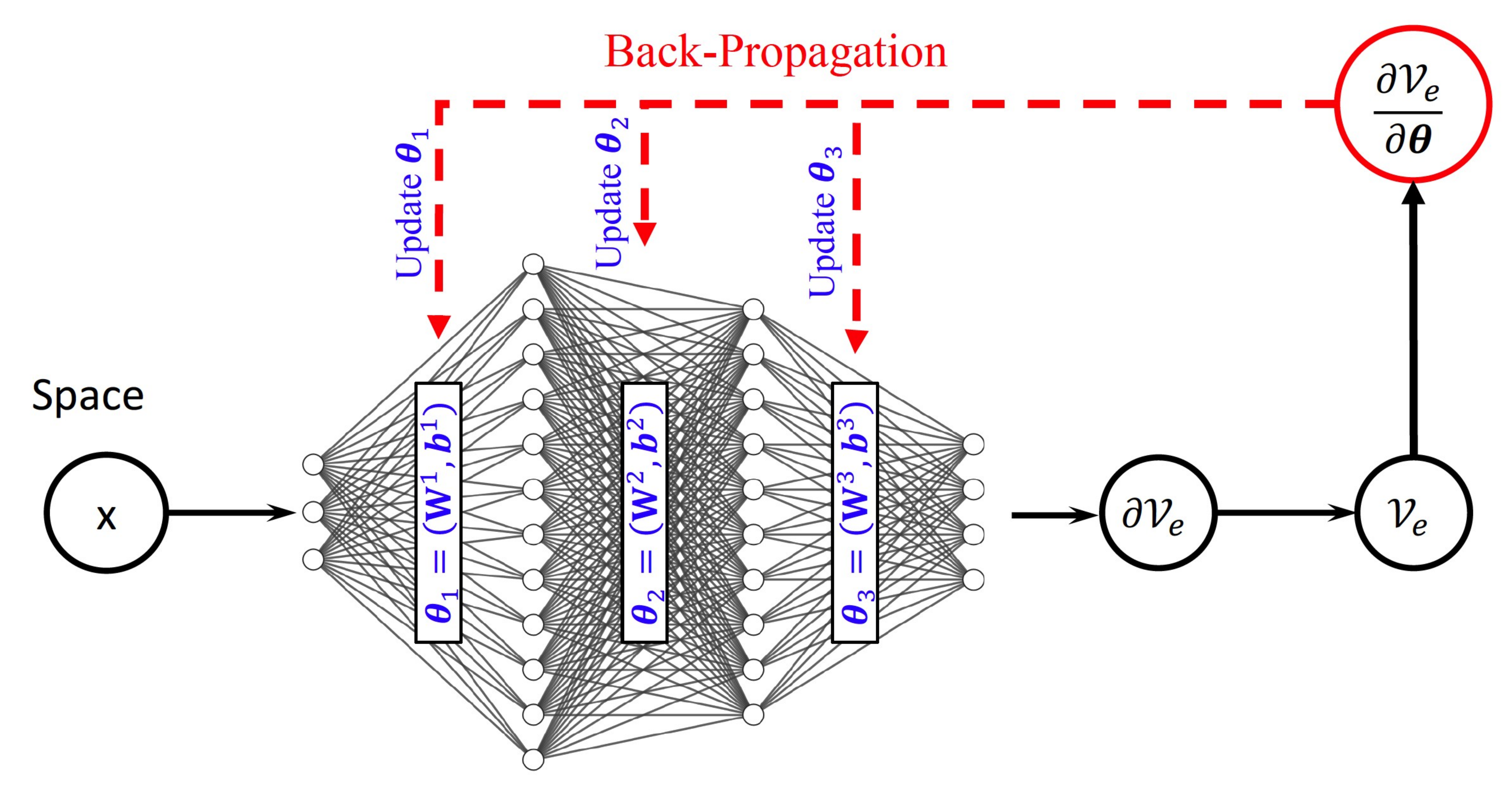}}
    \caption{(a) Schematic representation of the proposed physics informed neural network. For computing the derivatives, automatic differentiation (AD) has been used. All the neural networks share the same parameters $\bm{\theta}$. $\mathcal V_e$ represents the variational energy (b) Computational graph illustrating how the physics informed neural network is trained. The parameter $\bm {\theta}$ includes both weights and biases. For training, we have used the ADAM optimizer followed by L-BFGS.}
    \label{fig:schematic_rep_pinn}
\end{figure}

The steps involved in the proposed approach are as follows:
\begin{itemize}
\item First, we construct a neural network, $\mathcal N \left( x; \bm {\theta}\right)$
with parameters, $\bm{\theta}$ which includes both the weights and the biases.
As already stated in \autoref{subsec:dnn}, whether $\mathcal N$ is deep or shallow depends on the number of hidden layers.
\item Next, we modify the neural network outputs in such a way so that the Dirichlet boundary conditions are exactly satisfied.
To that end, we set,
\begin{equation}\label{eq:auto_bc}
    \begin{split}
    u \approx u_{NN} &= \tilde{u}_D + B\cdot\mathcal{N}\left( x; \bm {\theta}\right),\\
    & = \hat{\mathcal{N}}\left( x; \bm {\theta}\right),
    \end{split}
\end{equation}
where $\tilde{u}_D$ is a function chosen such that $\tilde{u}_D = u_D$ and $B=0$ on the Dirichlet boundary. As an example, if the boundary condition demands $u=0$ at $x=0$, we can set
\begin{equation}
    u \approx u_{NN} = x\mathcal N\left(x;\bm{\theta}\right).
\end{equation}
\item In the third step, we compute the derivatives present in the expression of the variational energy.
\begin{equation}\label{eq:pinn}
    \begin{gathered}
        u_x \approx \frac{\partial u_{NN}}{\partial x} = \hat{\mathcal N}^{x} \left(x; \bm{\theta} \right),\\
        u_{xx} \approx \frac{\partial^2 u_{NN}}{\partial x^2} = \hat{\mathcal N}^{xx} \left(x; \bm{\theta} \right), \\
        \vdots \\
        u_{x\cdots x} \approx \frac{\partial^n u_{NN}}{\partial x^n} = \hat{\mathcal N}^{x \cdots x}\left(x; \bm{\theta} \right).
    \end{gathered}
\end{equation}
We note that all the derivatives shown in \autoref{eq:pinn} are also neural networks with the same parameters $\bm {\theta}$.
The only difference resides in the fact that the form of the activation function has changed due to differentiation.
From computational point-of-view, we emphasize that the differentiation is carried out by using the automatic differentiation and hence, no manual calculations are needed.
\item In the fourth step, we utilize $u$ obtained from \autoref{eq:auto_bc} and the derivatives of $u$ obtained from \autoref{eq:pinn} to compute the energy variation.
\begin{equation}\label{eq:pinnve}
\begin{split}
    \partial \mathcal V_e & = \mathcal G \left( u, u_x, \ldots, u_{x\cdots x}, x, f\left(x\right) \right) \\
    & = \mathcal G \left(\hat{\mathcal N}\left(x;\bm{\theta}\right), \hat{\mathcal N}^x\left(x;\bm{\theta}\right), \ldots, \hat{\mathcal N}^{x\cdots x}\left(x;\bm{\theta}\right), x, f\left(x\right) \right).\\
\end{split}
\end{equation}
Since, $u$ and its derivatives are represented by neural networks,
it can be inferred that the energy variation $\partial \mathcal V_e$ is also represented by a neural network (see \autoref{eq:pinnve}).
More importantly, the fact that $\partial \mathcal V_e$ is a neural network, is based on the physics of the problem and hence, the name {\it physics informed neural network}.
The total variational energy of the system is computed by \autoref{eq:tot_energy}
\item Finally, we minimize the variational energy obtained in \autoref{eq:tot_energy} to compute the parameters, $\bm{\theta}$.
\begin{equation}\label{eq:var_en_pinn}
    \bm{\theta}^* = \argmin_{\bm{\theta}} \mathcal V_e.
\end{equation}
We note that unlike \autoref{eq:opt_ve}, the optimization problem in \autoref{eq:var_en_pinn} is an unconstrained optimization problem.
This is because we have already satisfied the boundary conditions using the modification in \autoref{eq:auto_bc}.
\end{itemize}
A schematic representation of the proposed framework and the computational graph for training the proposed PINN are shown in \autoref{fig:schematic_rep_pinn}.

\subsection{Numerical integration and geometrical modeling}
\label{subsec:NURBS}
In \autoref{subsec:pinn}, we presented details about the proposed PINN.
However, to use the proposed PINN in practice, we need to compute the integral in \autoref{eq:opt_ve}.
In this section, we discuss how the integral in \autoref{eq:tot_energy} can be computed in an efficient and accurate fashion.

An obvious option for computing the integral is to either use the trapezoidal rule or use Monte Carlo integration.
However, in general a large number of integration points are required. A more accurate alternative to the trapezoidal rule is to use the Gauss-Legendre rule.
However, the Gauss points generated are generally more dense near the boundary and hence, results obtained will be inaccurate in case there exist local features where the solution is not smooth (e.g. in the presence of cracks). In this work, we divide the problem domain into elements and subsequently, generate the Gauss points within each element. However, such an approach has two major challenges. First, it is essential that the geometry/domain of the problem is properly modeled. While this is trivial for a regular shaped domain, the challenge arises when the problem domain is irregular (e.g. a plate with a hole).
Second, even if we are able to model the problem domain accurately, generating uniform elements is unlikely to work. For example, finer meshes will be required at the vicinity of the crack. In this  section, we propose strategies to address both the issues discussed here.

Primarily, it is essential to obtain an accurate geometric description of the problem. 
An obvious option is to use piecewise polynomials commonly used in finite element analysis.
However, these functions cannot represent curved boundaries exactly, which can lead to errors related to the geometry during analysis. 
An alternative is to utilize NURBS to model the problem geometry.
NURBS are a generalization of B-splines and are represented using piecewise rational functions defined in parametric form, which makes it capable to represent free-form curves such as circles, cylinders, etc. 
Desired complex geometries are obtained by projective 
transformation of B-splines entities. For example 
circles are created by transformation of piecewise quadratic curves. 
Because of the above mentioned qualities, we model the geometry using NURBS patches in the proposed approach.

\begin{figure}
    \centering
    \subfigure[]{
    \includegraphics[width = 0.3\textwidth]{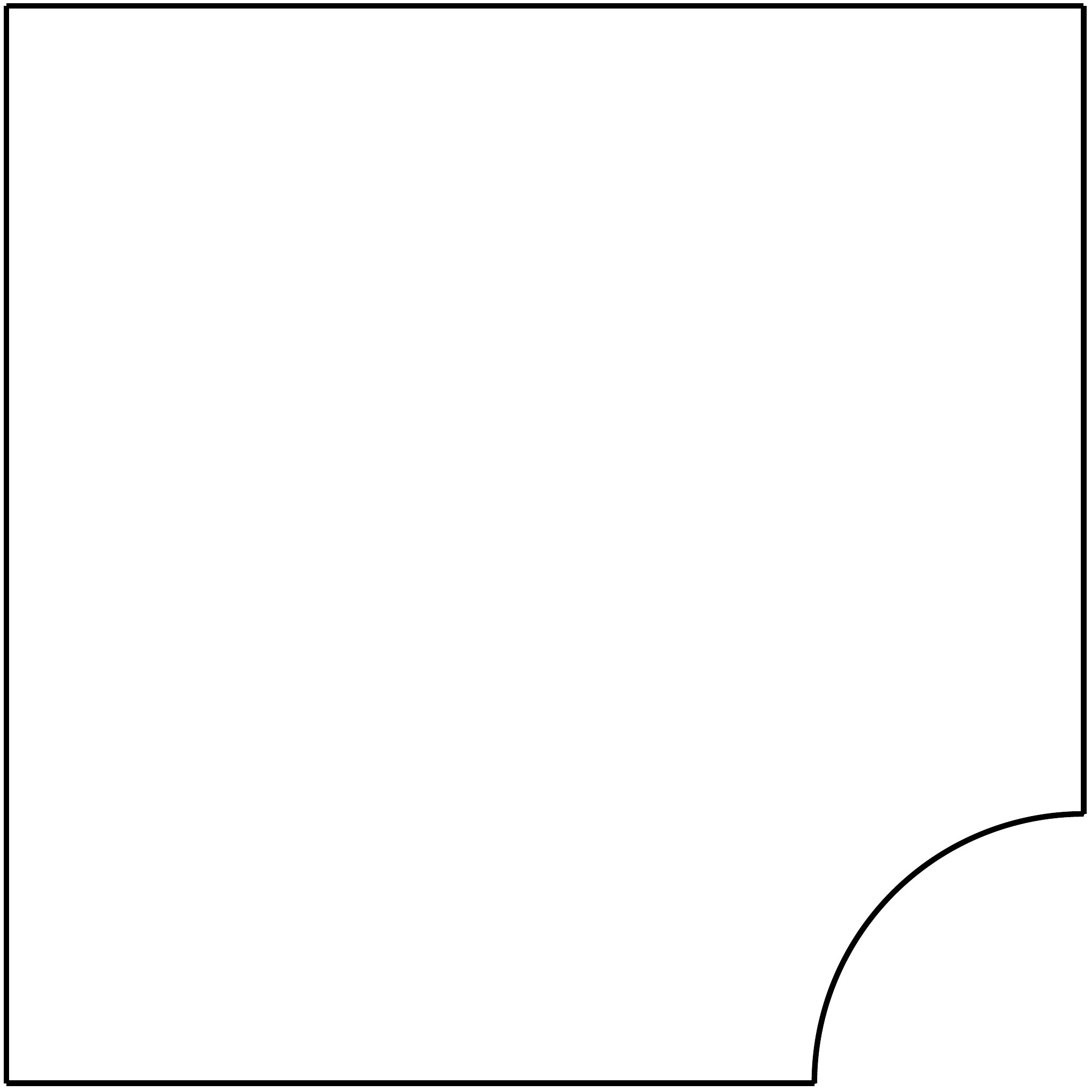}}
    \subfigure[]{
    \includegraphics[width = 0.3\textwidth]{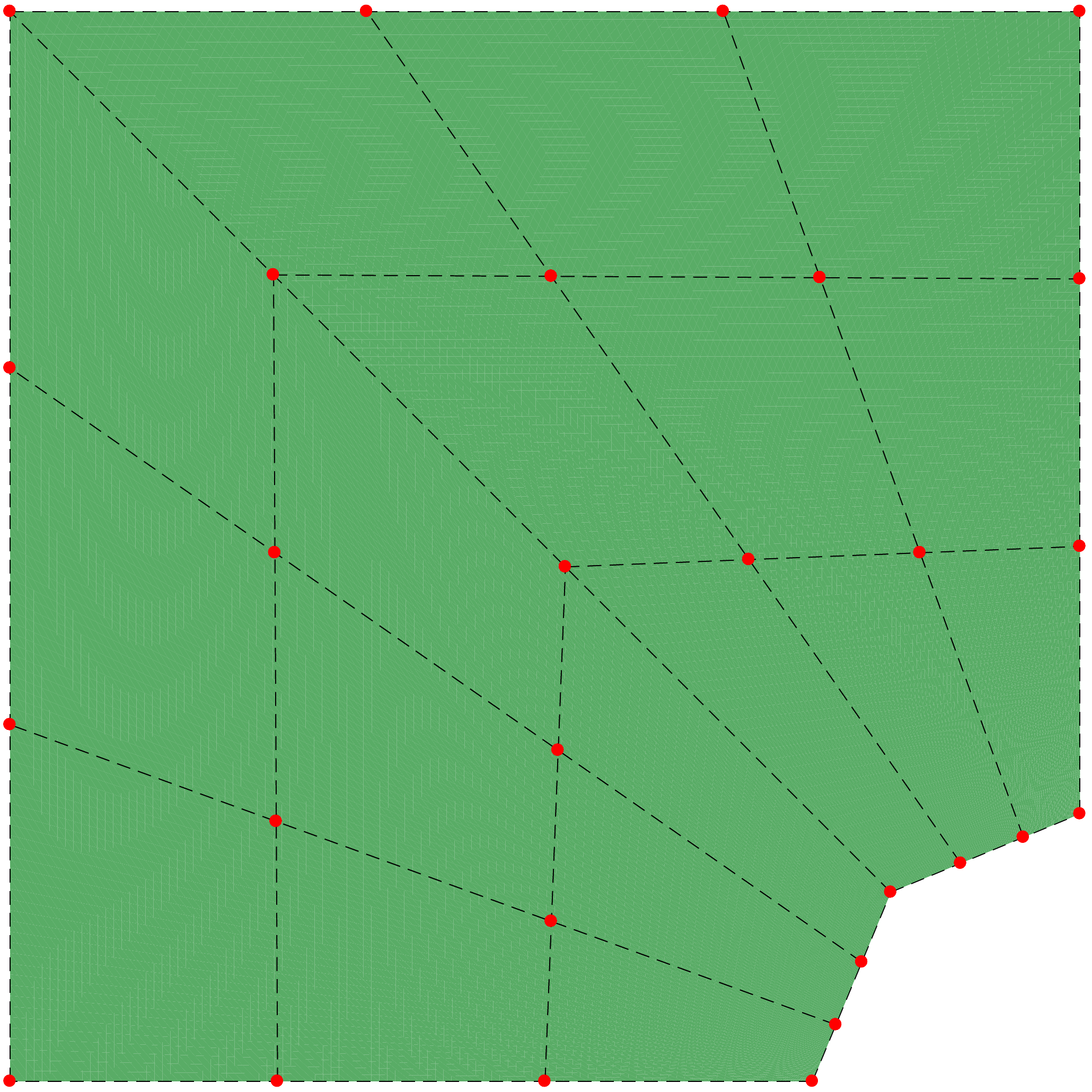}}
    \subfigure[]{
    \includegraphics[width = 0.3\textwidth]{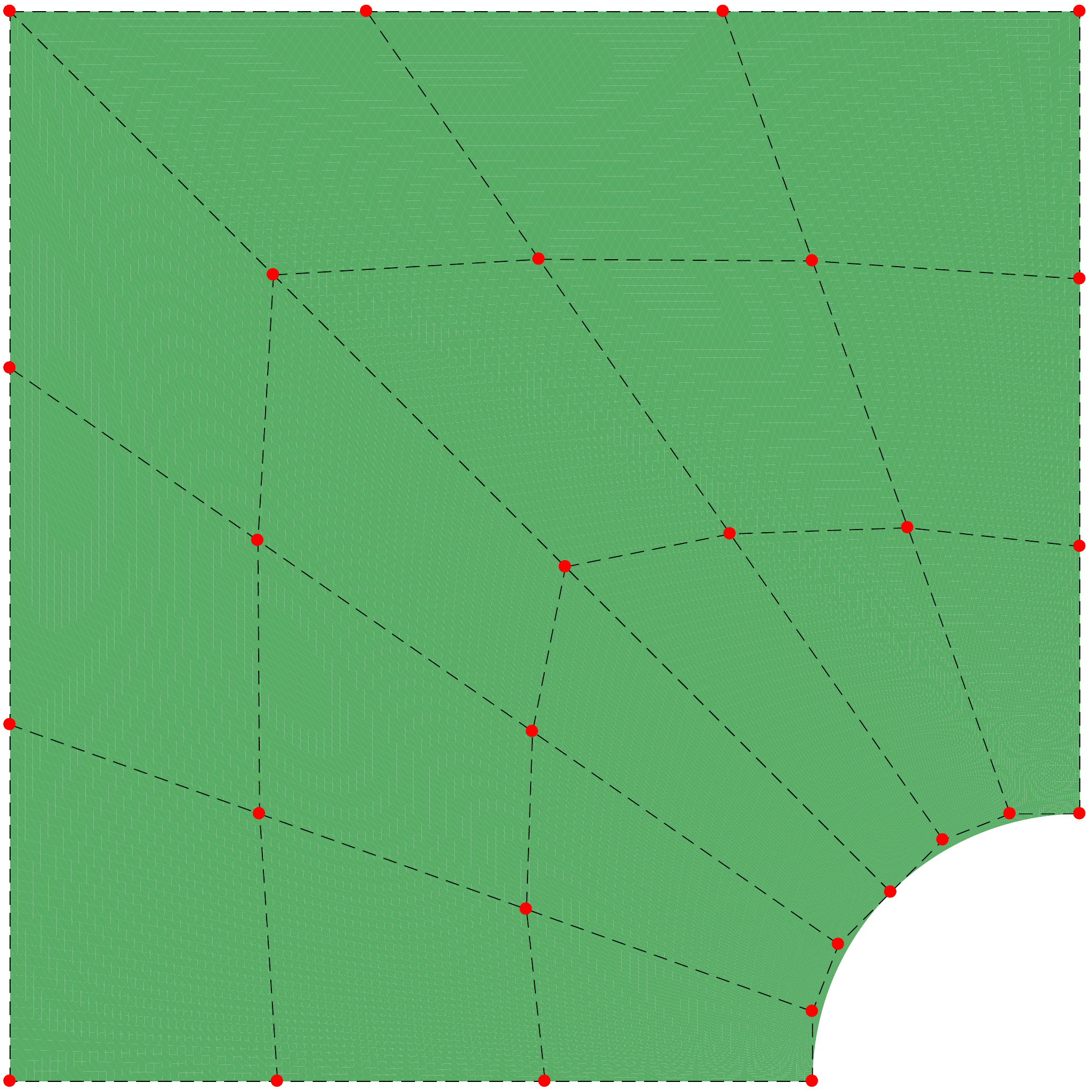}}
    \caption{(a) This is the original geometry to be modeled. (b) Modeling the geometry using the piecewise polynomial function. (c) Modeling the geometry using NURBS patch for bi-quadratic polynomial. The control mesh is shown using the dotted lines. The control points are shown in red and the modeled geometry is shown as the green patch.}
    \label{fig:Nurbs}
\end{figure}

In NURBS based modeling, there are two meshes, the control mesh and the physical mesh. 
The surface is expressed in terms of control mesh, in which each point, known as the control point, acts as an attractor of the resulting surface. 
The control mesh does not conform to the actual geometry, rather it forms a scaffold of the geometry. 
On the other hand, the physical mesh is a representation of the original geometry, which is decomposed into knot spans determined by a knot vector. A knot vector is an increasing set of parameter space coordinates, where each entry is called a knot. The initial set of knot vectors is denoted by the set of vertices ($\Xi^{i}$) corresponding to the spatial direction in the parameter space, $\Omega = \left[0,1\right]^d$:
\begin{equation}\label{eq:knot_vector}
    \Xi^{i} = \{{\xi_0^i},{\xi_1^i},{\xi_2^i},\ldots,{\xi_{n_i}^i}\},\text{    }i = 1,\ldots,d, 
\end{equation}
where $0={\xi_0^i}\leq{\xi_1^i}\leq{\xi_2^i}\leq,\ldots\leq{\xi_{n_i}^i}=1$. 
In \autoref{eq:knot_vector}, $n_i$ is the number of 
elements in each parametric direction. The sets 
$\Xi^{i}$ determine the initial tensor product mesh on level 0. A univariate rational basis function is 
defined as:
\begin{equation}
    R_{i,p}(\xi) = \frac{w_i N_{i,p}(\xi)}{\sum_{\hat i=1}^{n}w_{\hat i} N_{\hat {i},p}(\xi)}, \;\;\; 1\leq i \leq p+1,
\end{equation}
where $N_{i,p}(\xi)$ are the basis functions of the B-spline curve, $p$ is the degree of the polynomial function and $w_i$ is the weight associated with the control point. The uni-variate NURBS curve is given by:
\begin{equation}
    C(\xi) = \sum_{i = 1}^{n_{cp}}  R_{i,p}(\xi)B_i,
\end{equation}
where $B_i$ is the set of control points for the B-spline curve with knot vector $\Xi$ and $n_{cp}$ denotes the total number of NURBS control points. Tensor product generalizations of uni-variate B-splines are used to create multivariate B-splines. The tensor product mesh on the initial level of refinement ($\mathbb{T}_0$), can be written for a two-dimension as:
\begin{equation}\mathbb{T}_0 = \{\textit{E}_{0,k_m} = [\xi_{k_{1}-1}^{(1)},\xi_{k_1}^{(1)}]\times[\xi_{k_{2}-1}^{(2)},\xi_{k_2}^{(2)}], k_1 = 1,\ldots,n_1 \text{ and } k_2 = 1,\ldots,n_2\},
\end{equation}
where $E_{0,k}$ denotes an element in the mesh at level 0 and  $k_m = (k_2-1)n_1+k_1$. 
\autoref{fig:Nurbs} shows an example of modeling a geometry using NURBS patches over piecewise polynomial.

\begin{figure}[t]
    \centering
    \includegraphics[width = 0.6\textwidth]{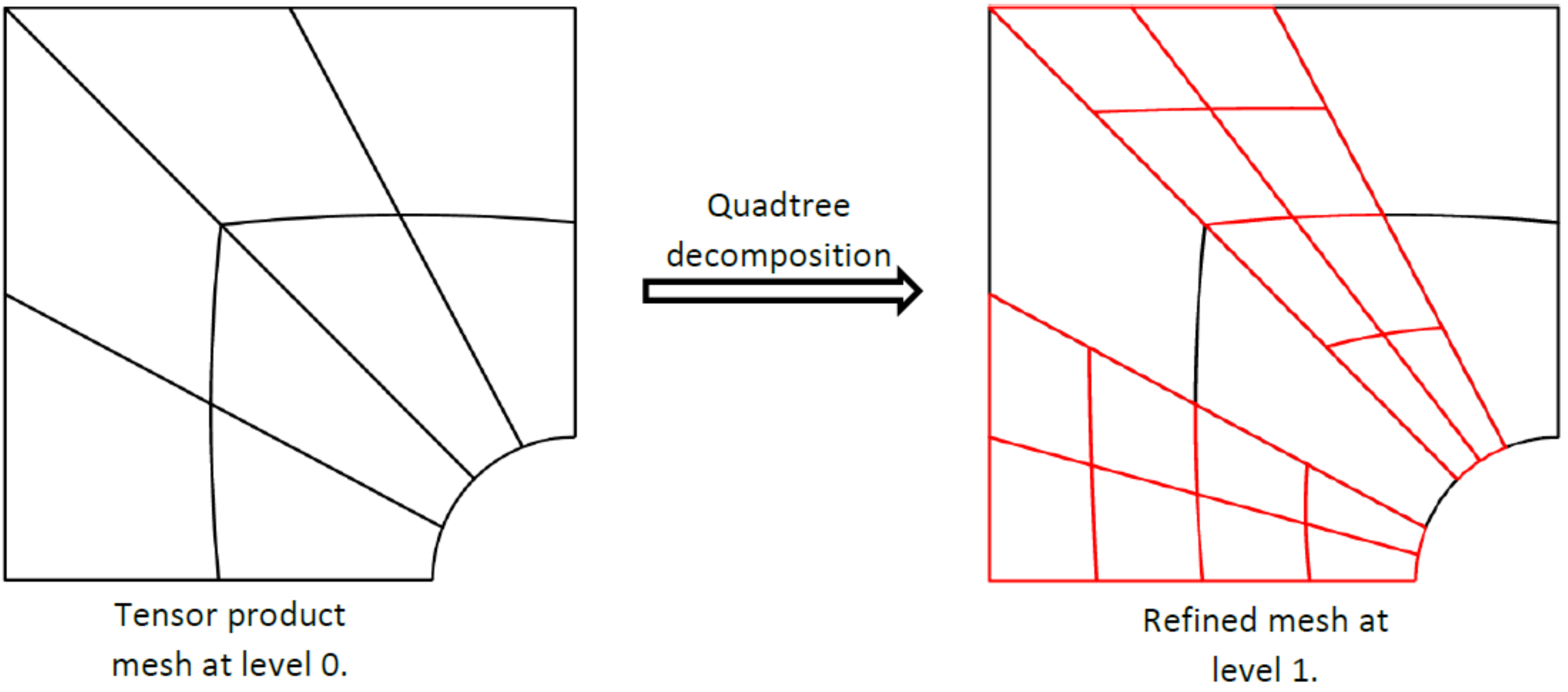}
    \caption{The initial tensor product mesh at level 0 is refined via cross-insertion technique using the quadtree decomposition approach. The elements marked in red on the right-side image shows the refined elements obtained using the quadtree decomposition.}
    \label{fig:quadtree}
\end{figure}

The initial tensor product mesh, $\mathbb{T}_0$ is locally refined along the crack path using quad/oct-tree decomposition of the domain. 
The local refinement is based on refinement of elements via `cross insertion'. The elements to be refined are subdivided into $2^d$ sub-cells, where $d$ denotes the number of spatial dimensions. \autoref{fig:quadtree} presents the local refinement of the domain using cross-insertion technique. For more information on modeling the geometry using NURBS and refinement using cross-insertion technique, we refer to \cite{piegl2012nurbs, Deng2008}.

After the geometrical model is generated using NURBS patches and is refined as desired (e.g., along the crack path), we generate the Gauss points within each element and their corresponding weights. The variational energy defined in \autoref{eq:tot_energy} is calculated at the Gauss points to approximate the integral.

\section{Phase-field modeling of fracture using PINN}
\label{sec:phase_pinn}
In the \autoref{sec:PINN}, we presented a new PINN where the parameters of the neural networks are trained based on the physics of a problem (defined in terms of a partial differential equation). In this section, we illustrate how the proposed PINN can be used for solving phase-field based brittle fracture problems defined in \autoref{sec:concept_fracture}.

In fracture mechanics, the primary goal is to obtain the crack path. This is generally achieved by applying a displacement/load increment until failure. In this work, we have considered displacement-controlled failure. We have also assumed a constant displacement step, $\Delta u$. With this setup, the proposed PINN is trained at each displacement increment and the strain-history function is updated at each increment. Before we start to train the network, we initialize the weights of the network randomly from a Gaussian distribution using the Xavier initialization technique \cite{glorot2010understanding}. Once the weights are initialized, we begin training the neural network. To that end, we first represent the displacement field, $\bm u$ and the phase-field, $\phi$ by using a deep neural network.
\begin{equation}\label{eq:primal_var}
    \left( \bm u, \phi \right) = \mathcal N\left(\bm x; \bm{\theta}\right)
\end{equation}
Without loss of generality, we assume that the neural network outputs already satisfy the boundary conditions. In case this is not true, we can modify the neural network outputs to exactly satisfy the boundary conditions (as described in \autoref{eq:auto_bc}). At this stage, the parameters $\bm{\theta}$ are unknown and the goal is to compute them based on the physics of the problem.

In order to compute the parameters of the neural network at the $i$-th displacement step,
we first follow the procedure described in \autoref{subsec:pinn} to generate the Gauss points, 
$\bm x^g$ and their corresponding weights, $w(\bm{x}^g)$.
In the next step, we use the automatic differentiation technique to compute the displacement gradients, $\nabla \bm u$ and the eigenvalues of the strain, $(\lambda_1, \ldots, \lambda_d)$, where $d$ is the number of spatial dimensions. 
The computed eigenvalues are then used to obtain $\Psi^+$ and $\Psi^-$.
\begin{subequations}\label{eq:decomposed_strain}
\begin{equation}
    \Psi^+ = \frac{\lambda}{8}\left( \lambda_s + \left| \lambda_s \right| \right)^2 + \frac{\mu}{4}\sum_{i=1}^d\left( \lambda_i + \left| \lambda_i \right| \right)^2,
\end{equation}
\begin{equation}
    \Psi^- = \frac{\lambda}{8}\left( \lambda_s - \left| \lambda_s \right| \right)^2 + \frac{\mu}{4}\sum_{i=1}^d\left( \lambda_i - \left| \lambda_i \right| \right)^2,
\end{equation}
\end{subequations}
where $\lambda_s = \sum_{i=i}^d \lambda_i$.
In the fourth step, we utilize $\Psi^+$ and $\Psi^-$ to compute $f_e\left(\bm x^g\right)$ (first equation in \autoref{eq:indv_ent}) and the history function, $H\left(\bm x^g, i \right)$ as:
\begin{equation}
    H\left(\bm x^g, i \right) = \max \left\{ \Psi^+, H\left(\bm x^g, i - 1 \right) \right\},\;\;\;\text{where}\;\; i>0.
\end{equation}
The crack is initialized at $i=0$ using \autoref{eq:initial_history_field}. In the next step, we use the automatic differentiation technique to obtain the phase-field gradients, $\nabla {\phi}$ and then use the gradients and $H\left(\bm x^g, i \right)$ to compute $f_c\left(\bm x^g\right)$ using the second equation in \autoref{eq:indv_ent}.

Next, we compute $\Psi_e$ and $\Psi_c$ by solving the integral problem in \autoref{eq:energyterms} using the Gauss-Legendre rule and then we approximate the total variational energy, $\mathcal V_e$ as defined in \autoref{eq:vare}.
\vspace*{-\baselineskip}
\begin{equation}\label{eq:vare}
\begin{split}
    \mathcal V_e &= \Psi_e + \Psi_c,\\
    \Psi_e & \approx \sum_{i = 1}^{N_{Pts}}f_e(\bm{x}^g_i)w(\bm{x}^g_i),\\
    \Psi_c & \approx \sum_{i = 1}^{N_{Pts}}f_c(\bm{x}^g_i)w(\bm{x}^g_i).\\
\end{split}
\end{equation}
Finally, we minimize $\mathcal V_e$ to compute the parameters $\bm {\theta}$. For optimization, we use the Adam (adaptive momentum) optimizer followed by a quasi-Newton method (L-BFGS).
We note that the above mentioned steps are to be repeated for each displacement step.
The overall framework of phase-field modeling of fracture using the proposed PINN is presented in \autoref{alg:PINN}.

\begin{algorithm}[ht]
\caption{PINN based phase-field approach.}
\label{alg:PINN}
\textbf{Initialize:} Provide displacement step ($\Delta u$), crack width ($l_0$), number of steps ($N_s$) and the neural network architecture.\\
Generate the geometry and obtain $\bm{x}^g$ and $w(\bm{x}^g)$ as per \autoref{subsec:NURBS}.\\
Generate coordinates of a fine prediction grid, $\bm{x}^*$ using uniformly spaced points (for visualization of the results).\\
$H(\bm{x}^g,0) \leftarrow 0$, $H(\bm{x}^*,0) \leftarrow 0$ and $\bm{\bar{u}}_0 \leftarrow 0$.\\
Initialize the neural network, $\mathcal{N}\left(x,\bm{\theta}\right)$.\\
Initialize the weights of the network using Xavier initialization technique \cite{glorot2010understanding}.\\
\For {$i= 0,\ldots,N_s$}{
$\bm{\bar{u}}_{i}$ = $\bm{\bar{u}}_{i-1}+\Delta u$.\\
Using \autoref{eq:primal_var}, represent the primal variables, $\bm u$ and $\phi$ as neural network. \\
Obtain the total variational energy loss using Eqs. (\ref{eq:decomposed_strain}) -- (\ref{eq:vare}).\\
Minimize the loss to compute the neural network parameters, $\bm{\theta}$. \\
Predict the history field, $H(\bm{x}^g,i)$ at $\bm{x}^g$ for $\bm{\bar{u}}_{i}$ and $H(\bm{x}^g,i-1)$.\\
Predict $\{u^*, v^*, \phi^*, H(\bm{x}^*,i)\}$ corresponding to $\bm{x}^*$ using $\bm{\bar{u}}_{i}$ and $H(\bm{x}^*,i-1)$.
}
\end{algorithm}

\begin{figure}[htbp!]
    \centering
    \includegraphics[width = 0.75\textwidth]{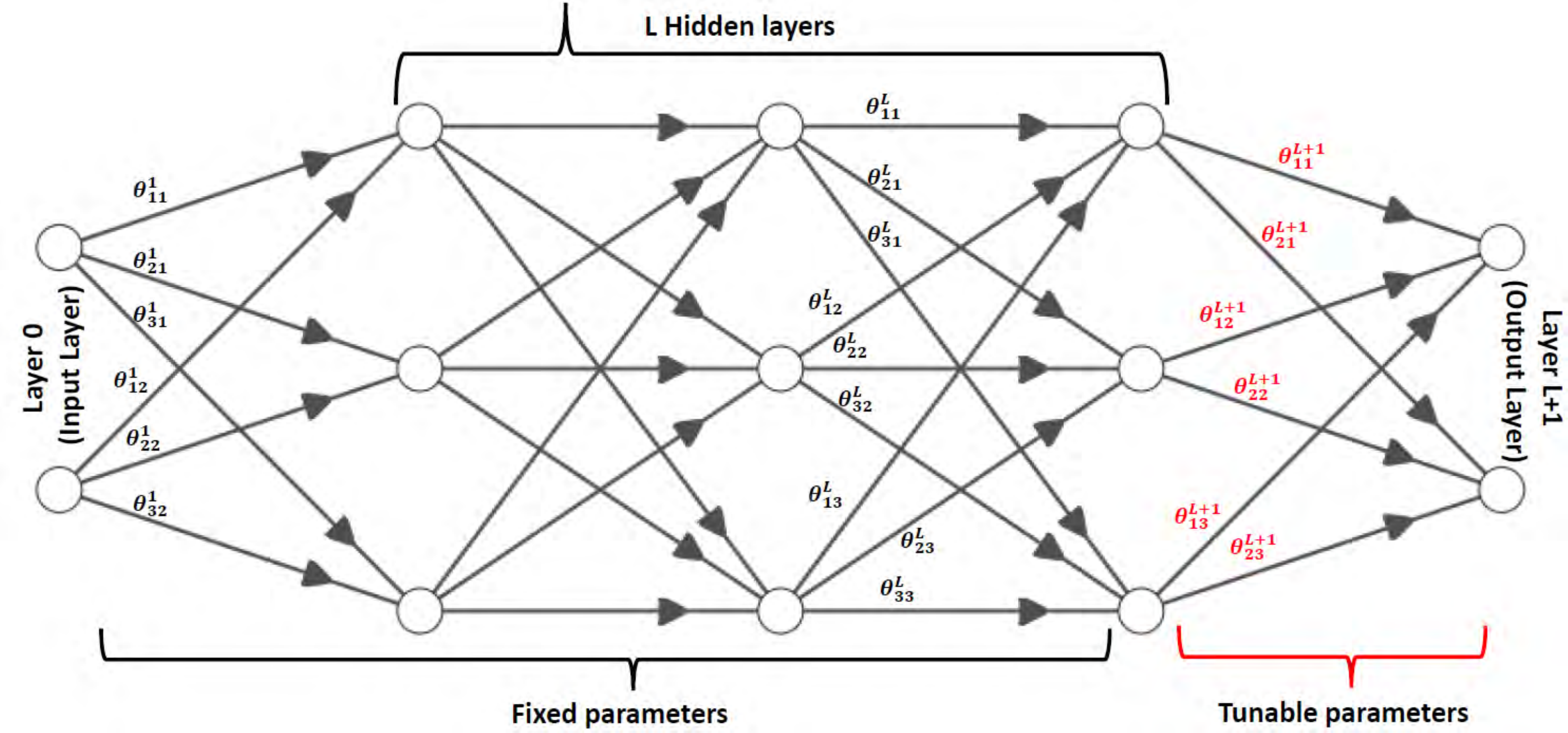}
    \caption{The parameters for each layer is represented as $\theta_{ij}^l$, which indicates the parameters corresponding to the $i$-th node on layer $l$ from the $j$-th node on layer $(l+1)$. For the first displacement step, all the weights are computed. However, for subsequent displacement steps, the parameters marked in black are fixed from the previous iterations, while the parameters shown in red are recomputed.}
  \label{fig:transfer_pinn}
\end{figure}

One major bottleneck associated with the application of the proposed PINN for phase-field based crack propagation problem resides in the fact that we need to train the model for each displacement-step.
Potentially, this can make the training phase computationally expensive. In order to address this issue, we propose to use the concept of `transfer learning'. Within this framework, we follow two simple steps. First, from second displacement-step onward, we only (retrain) the weights and biases corresponding to the last layer; weights and biases corresponding to all the other layers are kept fixed at the previously learned values. Second, even for the last layer, we start with the weights and biases of the $(i-1)$-th displacement step. The advantage of transfer learning is two-fold.
First, the number of iterations required to achieve a converged solution is much smaller.
Second, with this setup, the time required for each iteration is also substantially reduced.
A schematic representation of the proposed transfer learning scheme is shown in \autoref{fig:transfer_pinn}.

\section{Numerical examples}
\label{sec:numericals}
In this section, four well-known benchmark problems have been solved
to illustrate the performance of the proposed approach. 
The first example is of a one-dimensional elastic bar with a crack at the center, subjected to sinusoidal loading. 
An analytical solution for this problem is available. Hence, it is possible to validate the results obtained from the proposed approach. 
As the second example, we have solved the `single-edge notched tension test' problem.
For both first and second examples, we illustrate the superiority of the proposed PINN over conventional residual based PINN approach.
We also illustrate how using Gauss-Legendre rule results in an efficient solution.
As the third example, we have selected the `perforated and notched asymmetric bending test' problem.
With this example, we illustrate how the proposed PINN performs for problems with complicated domain geometries.
Finally, in the last example, we have considered crack propagation in a cube subjected to tensile loading.
This example illustrates the performance of the proposed approach for three-dimensional problems.

For all the problems, the proposed PINN is trained by using a combination of Adam optimizer and second-order quasi-Newton method (L-BFGS).
The implementation has been carried out using the \texttt{TensorFlow} framework \cite{abadi2015tensorflow}.
For accelerating the training algorithm, transfer learning as discussed in \autoref{sec:phase_pinn} has been used.
Details on the network architecture, such as number of layers, number of neurons in each layer, the activation function used etc., have been provided with each example. 
\subsection{One-dimensional elastic bar with crack}
\label{subsec:1D_Elastic_Test}
\begin{figure}[htbp!]
    \centering
    \includegraphics[width = 0.7\textwidth]{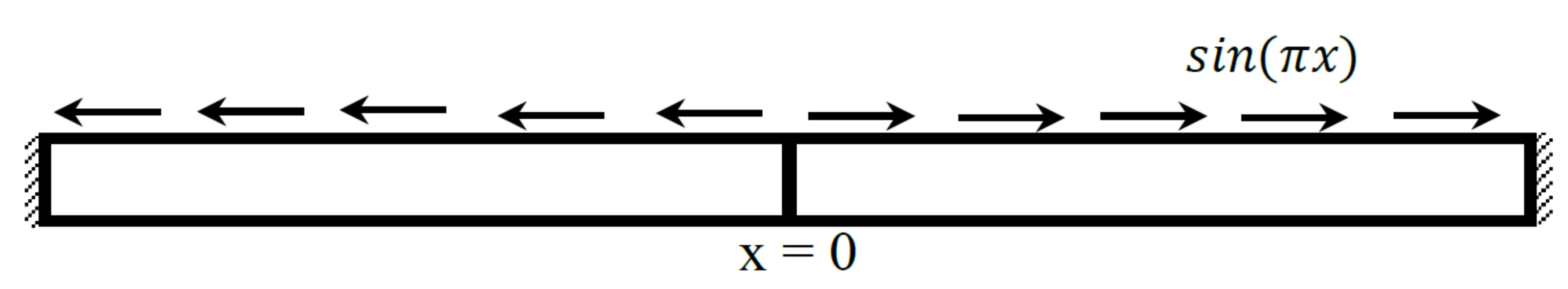}
    \caption{Geometrical setup of one-dimensional elastic bar with crack.}
  \label{fig:1D_setup}
\end{figure}

We consider a one-dimensional bar that is fixed at both the ends and is subjected to a sinusoidal load. The bar has a crack at the center. The geometrical setup is presented in \autoref{fig:1D_setup}. 
For simplicity $E$ is assumed to be unity and the strain ($\epsilon$) is assumed to be non-negative in the crack zone. Hence, the stress-strain relation is obtained by: 
\begin{equation}
    \sigma = g(\phi)\epsilon.
\end{equation}
$H_0(x)$ as stated in \autoref{eq:initial_history_field} is defined as:
\begin{equation}
    H(x,0) = \left\{ {\begin{array}{*{20}{c}}
  {1000}&{d(x,l) \leqslant l_0} \\ 
  0&{d(x,l) > l_0} 
\end{array}} \right.,
\end{equation}
where $l_0$ is considered to be 0.0125. 
In the fully cracked scenario, the analytical solutions for the displacement field, $u$ ($u_{ex}$) \cite{Schillinger2015} and phase-field, $\phi(x)$ ($\phi_{ex}$) \cite{Miehe2010} are given as:
\begin{subequations}
\begin{equation}{\label{eq:analyticalSoln_1D_u}}
    {u_{ex}} = \left\{ {\begin{array}{*{20}{c}}
  {\frac{1}{{{\pi ^2}}}\sin (\pi x) - \frac{{(1 + x)}}{\pi }}&{\text{  if  }x < 0} \\ 
  {\frac{1}{{{\pi ^2}}}\sin (\pi x) + \frac{{(1 + x)}}{\pi }}&{\text{  if  }x > 0} 
\end{array}} \right..
\end{equation}
\begin{equation}\label{eq:analyticalSoln_1D_phi}
    \phi_{ex} = \exp\left(\frac{-|x-a|}{l_0}\right),
\end{equation}
\end{subequations}
where the crack is located at $a$ units. The Dirichlet boundary conditions are:
\begin{equation}\label{eq:coupled1D_dirichlet}
    u(-1) = u(1) = 0,
\end{equation}
where $u$ is the solution of the elastic field in \textit{x}-axis.
The purpose of selecting this problem is two-fold.
First, we show that the proposed PINN can yield accurate results.
Second, we establish that the proposed PINN is more accurate as compared to the conventional residual based PINNs available in literature.

In order to solve this problem, we have considered a fully connected neural network with 3 hidden layers, comprising of 50 neurons in each hidden layer. For the first two layers, we have considered hyperbolic 
tangent, (\texttt{tanh}) activation function; whereas for the last layer, 
linear activation function has been considered.
We have subdivided the domain into three sections along the \textit{x}-axis; $lc$:[$-1.0$, $-2l_0$], $c$:[$-2l_0$, $2l_0$], 
$rc$:[$2l_0$, $1.0$], where $lc$ and $rc$ represent 
the left and right side of the crack zone, respectively and 
$c$ is the crack zone. 
This is done to generate more integration points in the 
vicinity of the crack.
In each of the three sections, 
$N_x = 112$ Gauss points have been generated.
In order to ensure that the neural network output 
exactly satisfies the Dirichlet boundary conditions, 
we have set
\begin{equation}
    u = [(x+1)(x-1)]\hat{u},
\end{equation}
where $\hat{u}$ is obtained from the neural network.
In the proposed PINN approach, we train the network by minimizing the total variational energy of the system as defined in \autoref{eq:vare}.
In order to quantify the accuracy of the results obtained using the proposed approach, we use the relative $\mathcal L_2$ error
\begin{equation}\label{eq:L2norm}
    \begin{split}
    \mathcal{L}_2^{rel,u} = \frac{\sqrt{\sum_{i = 1}^{N_{pred}}{(u(x_i) - u_{ex}(x_i))^2}dx}}{\sqrt{\sum_{i = 1}^{N_{pred}}{(u_{ex}(x_i))^2}dx}},\\
    \mathcal{L}_2^{rel,\phi} = \frac{\sqrt{\sum_{i = 1}^{N_{pred}}{(\phi(x_i) - \phi_{ex}(x_i))^2}dx}}{\sqrt{\sum_{i = 1}^{N_{pred}}{(\phi_{ex}(x_i))^2}dx}}
    \end{split}
\end{equation}

\begin{figure}[htbp!]
    \centering
    \subfigure[Comparison of $u_{exact}$ and $u_{comp}$.]{
    \includegraphics[width = 0.65\textwidth]{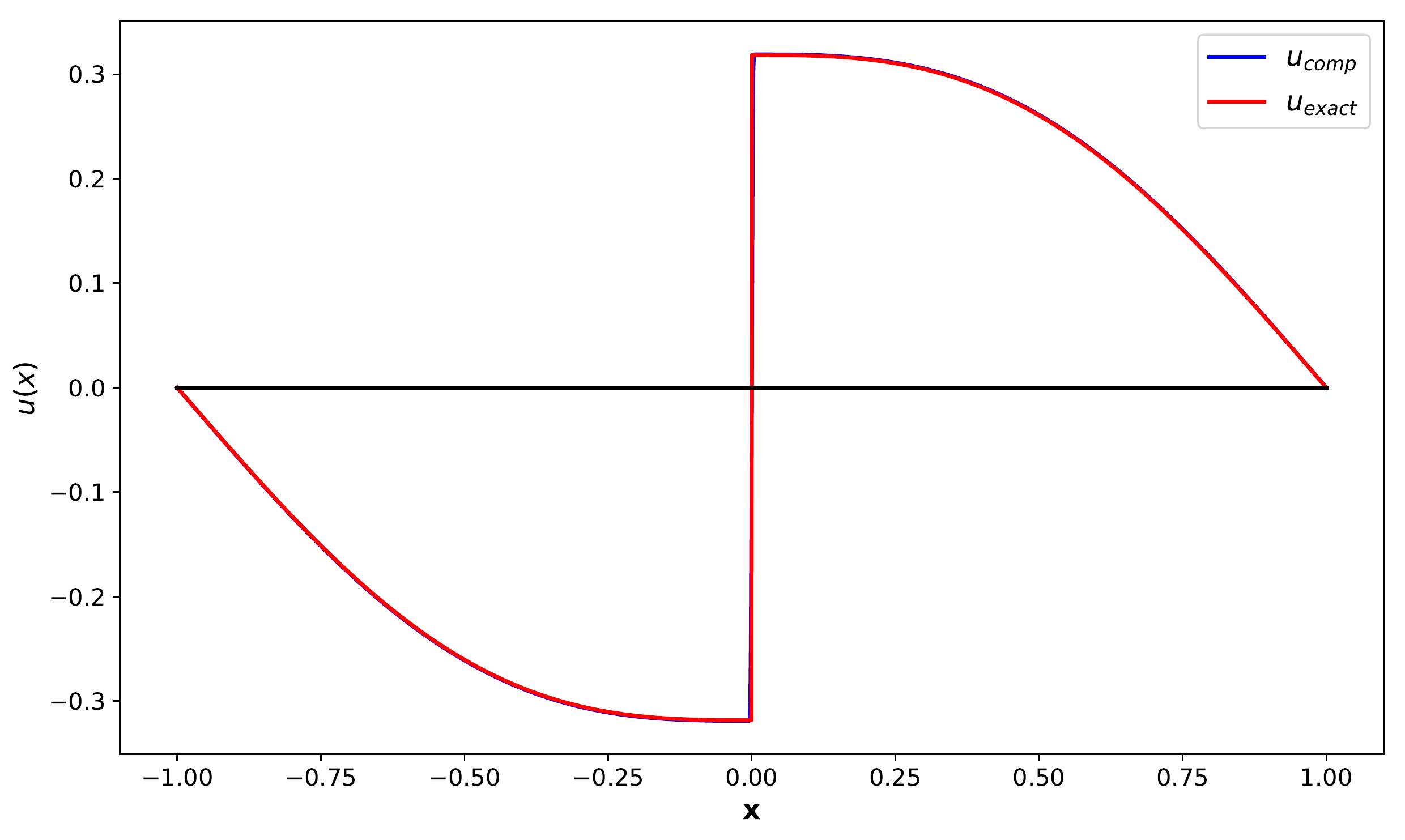}}
    \subfigure[Comparison of $\phi_{exact}$ and $\phi_{comp}$.]{
    \includegraphics[width = 0.4\textwidth]{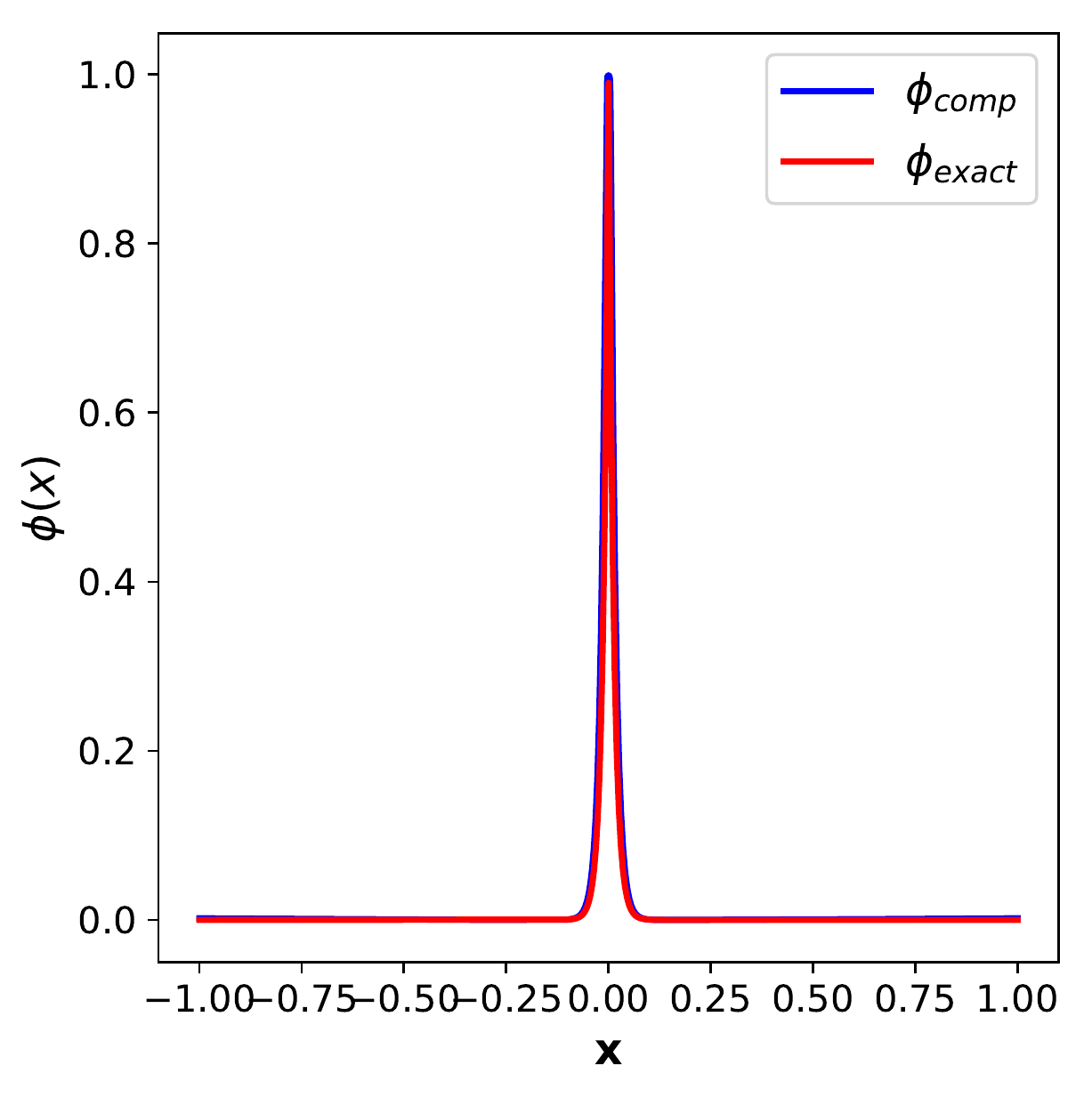}}
    \subfigure[Convergence of the loss function.]{
    \includegraphics[width = 0.4\textwidth]{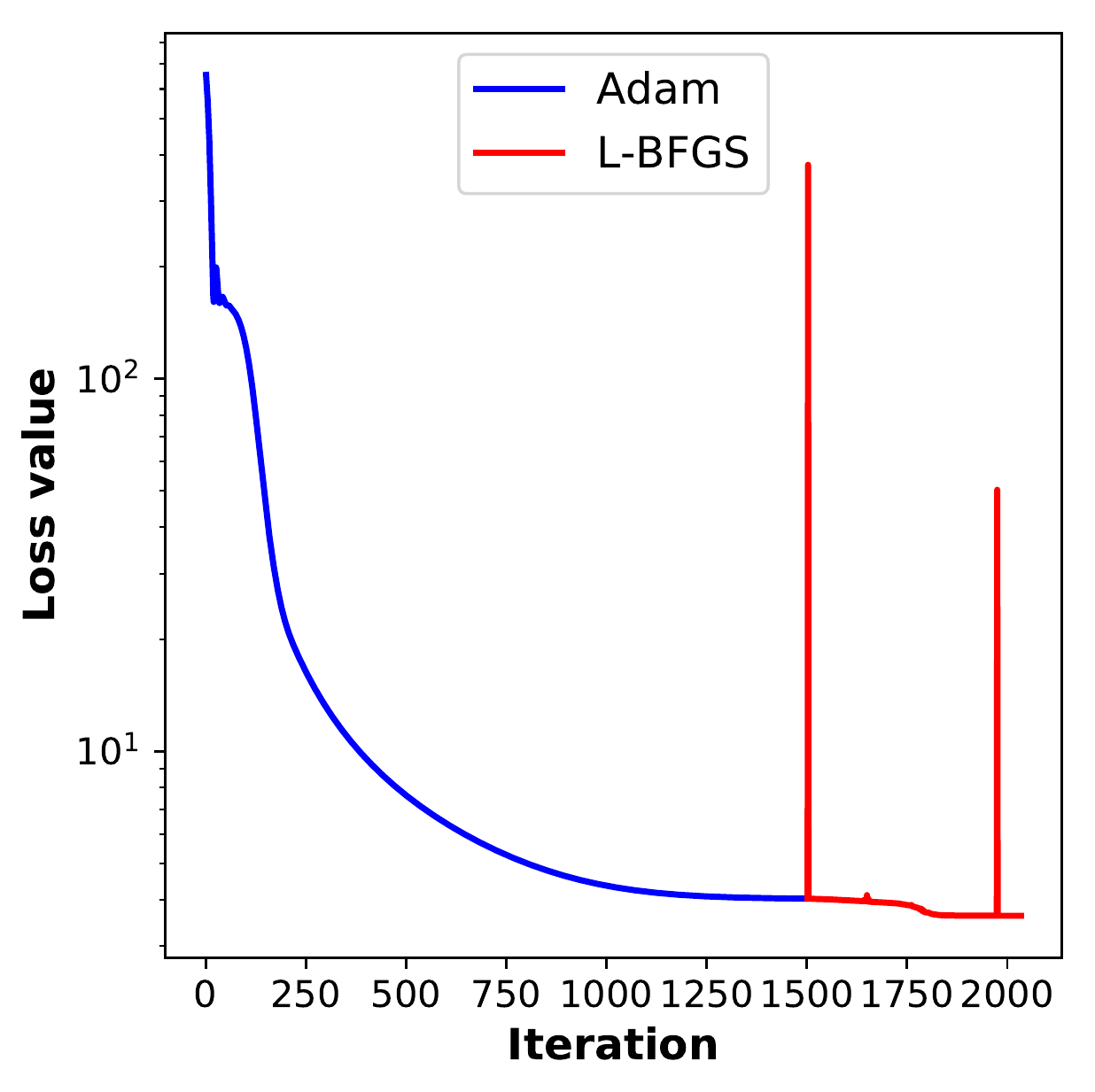}}
    \caption{Results for one-dimensional elastic bar with crack using variation energy based PINN (proposed approach).}
    \label{fig:prob1}
\end{figure}

Figs. \ref{fig:prob1}(a) and \ref{fig:prob1}(b) show the displacement field, $u$ and the 
phase-field, $\phi$ obtained using the proposed approach.
Visually, the results obtained using the proposed approach
overlaps with the analytical solutions obtained using Eqs. (\ref{eq:analyticalSoln_1D_u}) and (\ref{eq:analyticalSoln_1D_phi}).
To quantify the accuracy of the proposed PINN approach, the relative $\mathcal L_2$ error corresponding to $u$ and $\phi$ has been computed using \autoref{eq:L2norm}.
Corresponding to $u$ and $\phi$, a prediction error of 4.46\% and 3.61\%, respectively have been observed.
As for the computational cost,
the proposed approach requires only 336 ($3 \times 112$) Gauss points and 2100 (1500 + 600) iterations. The convergence history is shown in \autoref{fig:prob1}(c).

In order to illustrate the superiority of the proposed approach, 
we compare the results obtained using the proposed approach
with those obtained using the residual based PINN.
For a fair comparison, the neural network architecture and the number of integration points are kept same.
The residual based PINN is trained following the approach described in \cite{Raisi2017Physics}. It is observed that the residual based PINN fails to capture the sharp change in the gradient at the location of the crack. Consequently, the prediction error using the residual based PINN is 85.87\% for $u$ and 91.55\% for $\phi$. This illustrates the superiority of the proposed PINN over the residual based PINN.

Finally, we illustrate the advantage of using the Gauss-Legendre rule over generating uniformly distributed points to approximate the integrals.
To that end, we solve the problem using the uniformly distributed points, rather than Gauss points.
We observe that the neural network setup as discussed above (i.e., 3 hidden layers with 50 neurons each) with 336 integration points (since we generate uniformly distributed point, we refer them as integration points), fails to capture the variation of $u$ and $\phi$.
After several trials with different network architectures and numbers of integration points,
we observed that to obtain solutions of similar accuracy as reported earlier, 
5 hidden layers of 50 neuron each are required.
More importantly, the total number of integration points required are 8000 (1000+6000+1000).
This illustrates the computational advantage gained by using Gauss points over uniformly distributed integration points.
The prediction errors corresponding to the various case studies are shown in \autoref{tab:prob1}.

\begin{table}[htbp!]
    \centering
    \caption{Summary of results corresponding problem 1. We observe that the proposed approach yields the best results.}
    \label{tab:prob1}
    \begin{tabular}{llccc}
        \hline
        \multirow{2}{*}{\textbf{Methods}} & \multirow{2}{*}{\textbf{PINN Architecture}} & \multirow{2}{*}{\textbf{Integration points}} & \multicolumn{2}{c}{\textbf{Prediction error}} \\ \cline{4-5}
         & & & $\mathcal L_2^{rel,u}$ & $\mathcal L_2^{rel,\phi}$ \\ \hline
         VE-PINN$^*$ & $\left[1,50,50,50,2\right]$ & 336 & 4.46\% & 3.61\% \\
         R-PINN$^{\#}$ & $\left[1,50,50,50,2\right]$ & 336 & 85.87\% & 91.55\% \\
         VE-PINN2$^{\dagger}$ & $\left[1,50,50,50,50,50,2\right]$ & 8000 & 7.54\% & 3.58\% \\ \hline
         \multicolumn{5}{l}{\small $^*$VE-PINN = variational energy based PINN (proposed approach)} \\
         \multicolumn{5}{l}{\small $^{\#}$R-PINN = residual based PINN (conventional approach)} \\
         \multicolumn{5}{l}{\small $^{\dagger}$VE-PINN2 = Same as VE-PINN, but with uniformly distributed integration points} 
    \end{tabular}
\end{table}

\subsection{Single-edge notched tension example}
\label{sec:2D_tension_Test}
\begin{figure}[htbp!]
    \centering
    \subfigure[Geometrical setup and boundary conditions.]{
    \includegraphics[width = 0.45\textwidth]{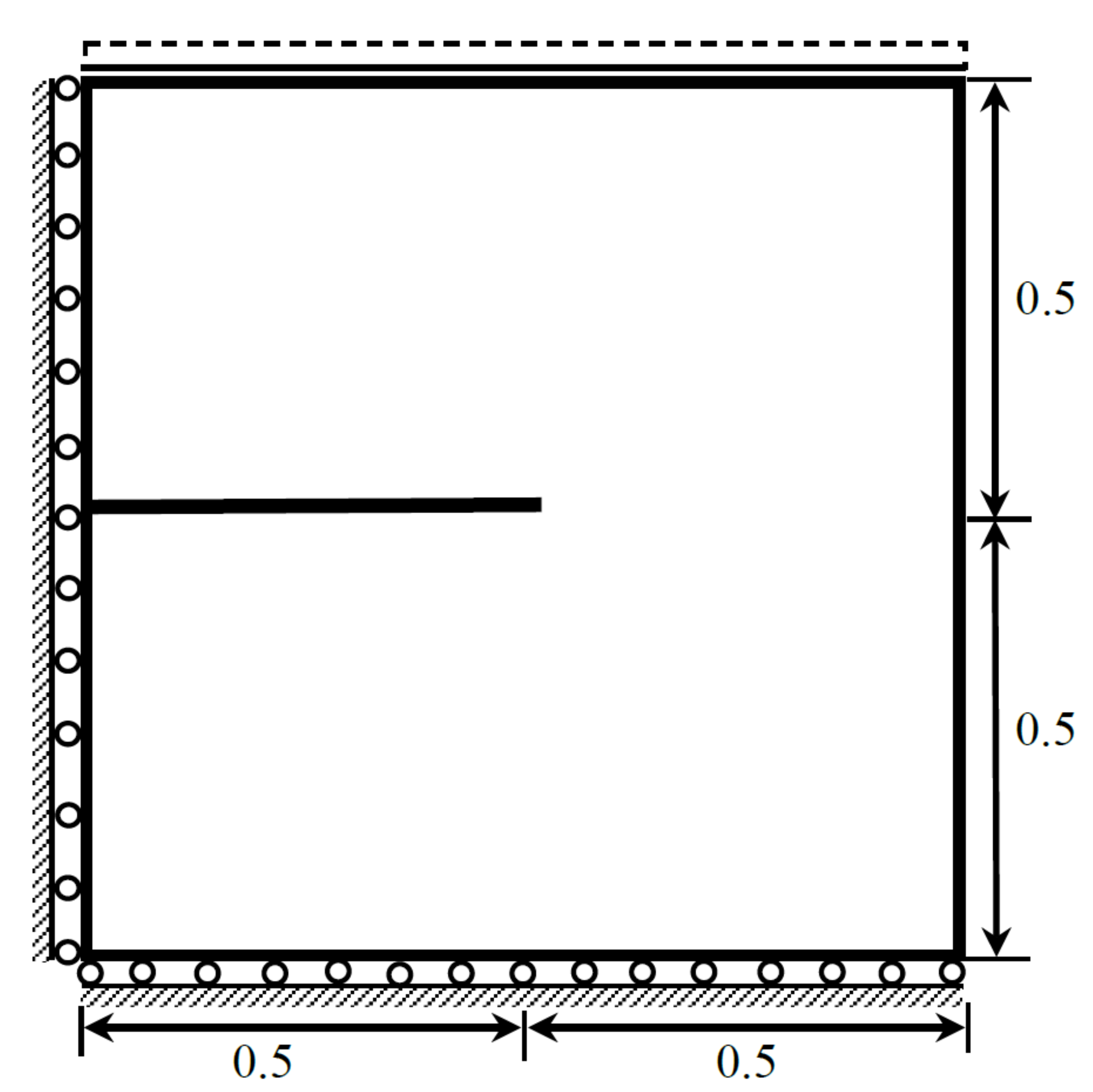}}
    \subfigure[Mesh for generating the Gauss points.]{
    \includegraphics[width = 0.42\textwidth]{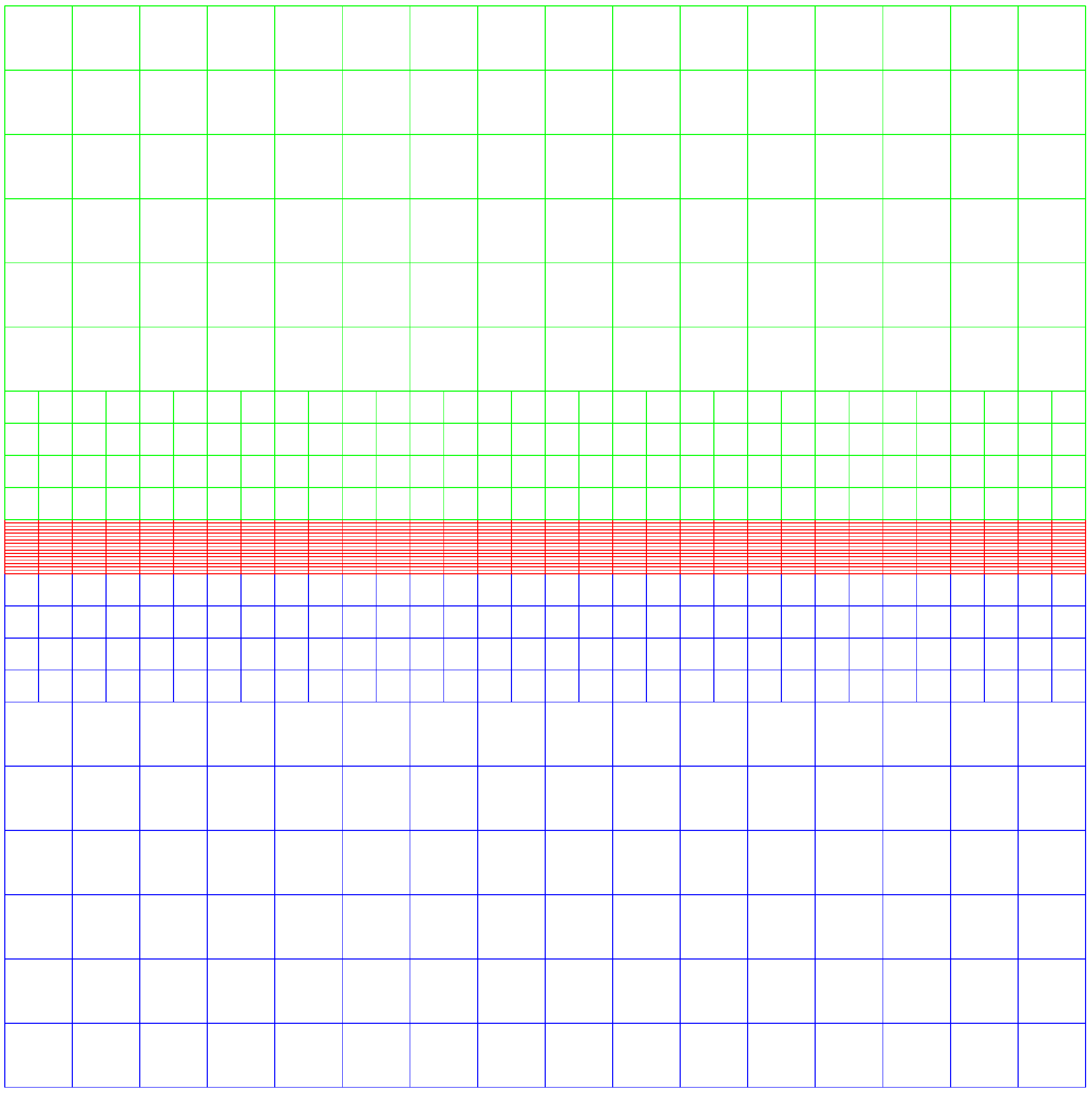}}
    \caption{Single-edge notch tension example.}
    \label{fig:setup}
\end{figure}

In this example, we consider a unit square plate with a 
horizontal crack from the midpoint of the left outer edge to 
the center of the plate. The geometric setup and boundary 
conditions of the problem are shown in 
\autoref{fig:setup}(a). The material 
properties of the plate are $\lambda = $ 121.15 kN/mm$^{2}$, $\mu = $ 80.77 kN/mm$^{2}$ and $G_c = 2.7 \times 10^{-3}$ kN/mm. In this example, we consider $l_0 = 0.0125$. The plate geometry, although simple, is generated using NURBS (see \autoref{subsec:NURBS}).
Using a quad-tree refinement, the plate is subdivided in three levels --
(a) level 0: we have $16\times6\times2$ elements,
(b) level 1: we have $32\times4\times2$ elements and
(c) level 2: we have $64 \times 8$ elements.
Overall, the domain has 960 elements.
\autoref{fig:setup}(b) presents the modeled mesh used for the
generation of Gauss points for training the deep neural network. 
The objective is to compute the crack path and the failure load of the system.

For obtaining the crack path for the single edge notched 
plate under tensile loading, we have used a fully connected neural network with 3 hidden layers of $50$ neurons each. Within each element, 64 Gauss points have been generated.
Similar to previous example, for the first two layers \texttt{tanh} activation function and for the last layer linear activation function have been used.
The crack is initiated using the strain history functional defined in \autoref{eq:initial_history_field}. The Dirichlet boundary conditions are:
\begin{equation}
    u(0,y) = v(x,0) = 0, \;\;\; v(x,1)= \Delta v,
\end{equation}
where $u$ and $v$ are the solutions of the elastic field in \textit{x} and \textit{y}-axis, respectively. 
For obtaining the crack path, a constant displacement 
increment of $\Delta v$ = $0.5\times 10^{-3}$ mm has been applied.
To exactly satisfy the Dirichlet boundary conditions, the neural network outputs for the elastic field are modified as:
\begin{equation}
\begin{split}
    u &= [x(1-x)]\hat{u},\\
    v &= [y(y-1)]\hat{v} + y\Delta v, 
\end{split}
\end{equation}
where $\hat{u}$ and $\hat{v}$ are obtained from the neural network.

\begin{figure}[htbp!]
    \centering
    \subfigure[]{
    \includegraphics[width = 0.3\textwidth]{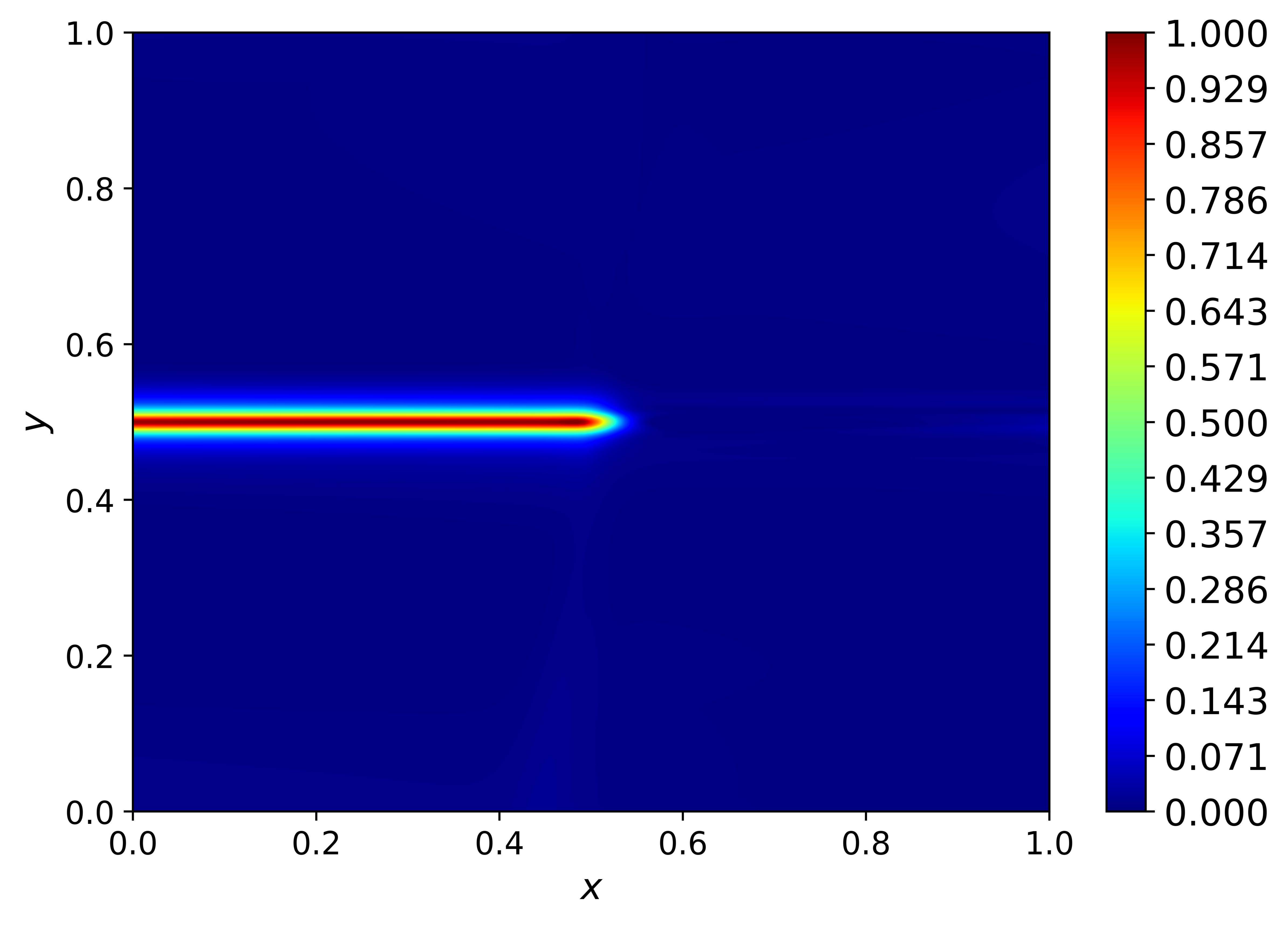}}
    \subfigure[]{
    \includegraphics[width = 0.3\textwidth]{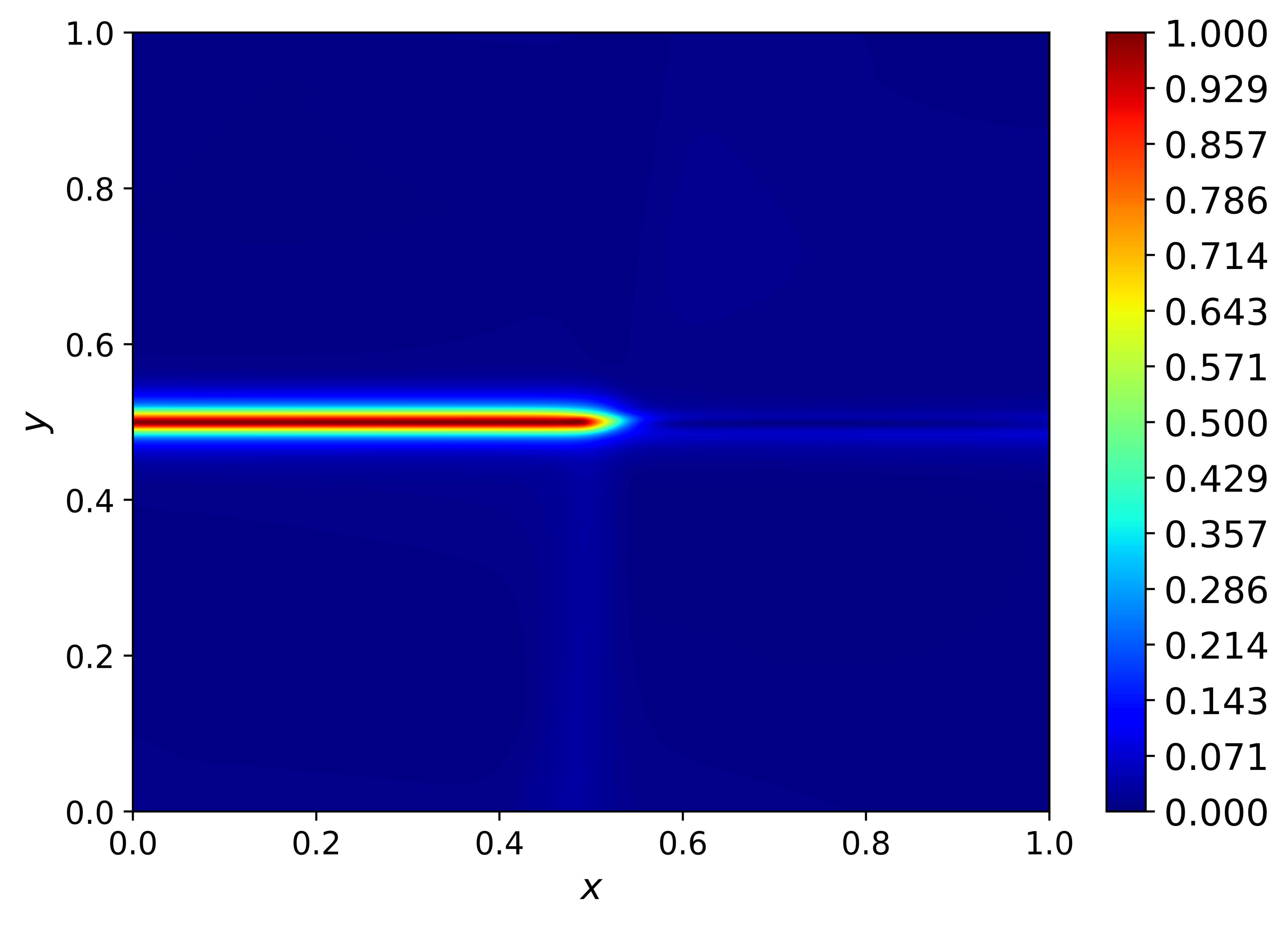}}
    \subfigure[]{
    \includegraphics[width = 0.3\textwidth]{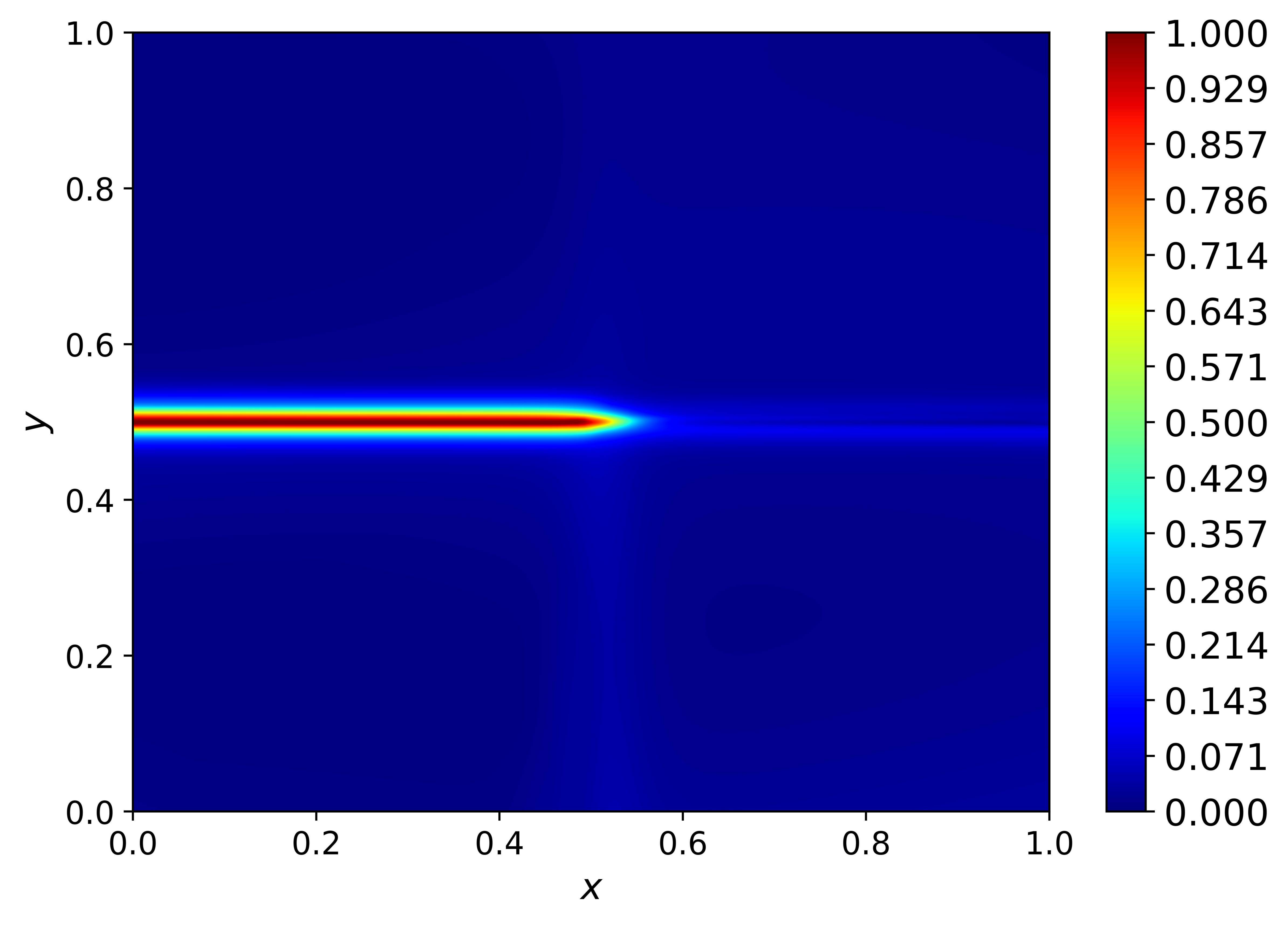}}
   \subfigure[]{
   \includegraphics[width = 0.3\textwidth]{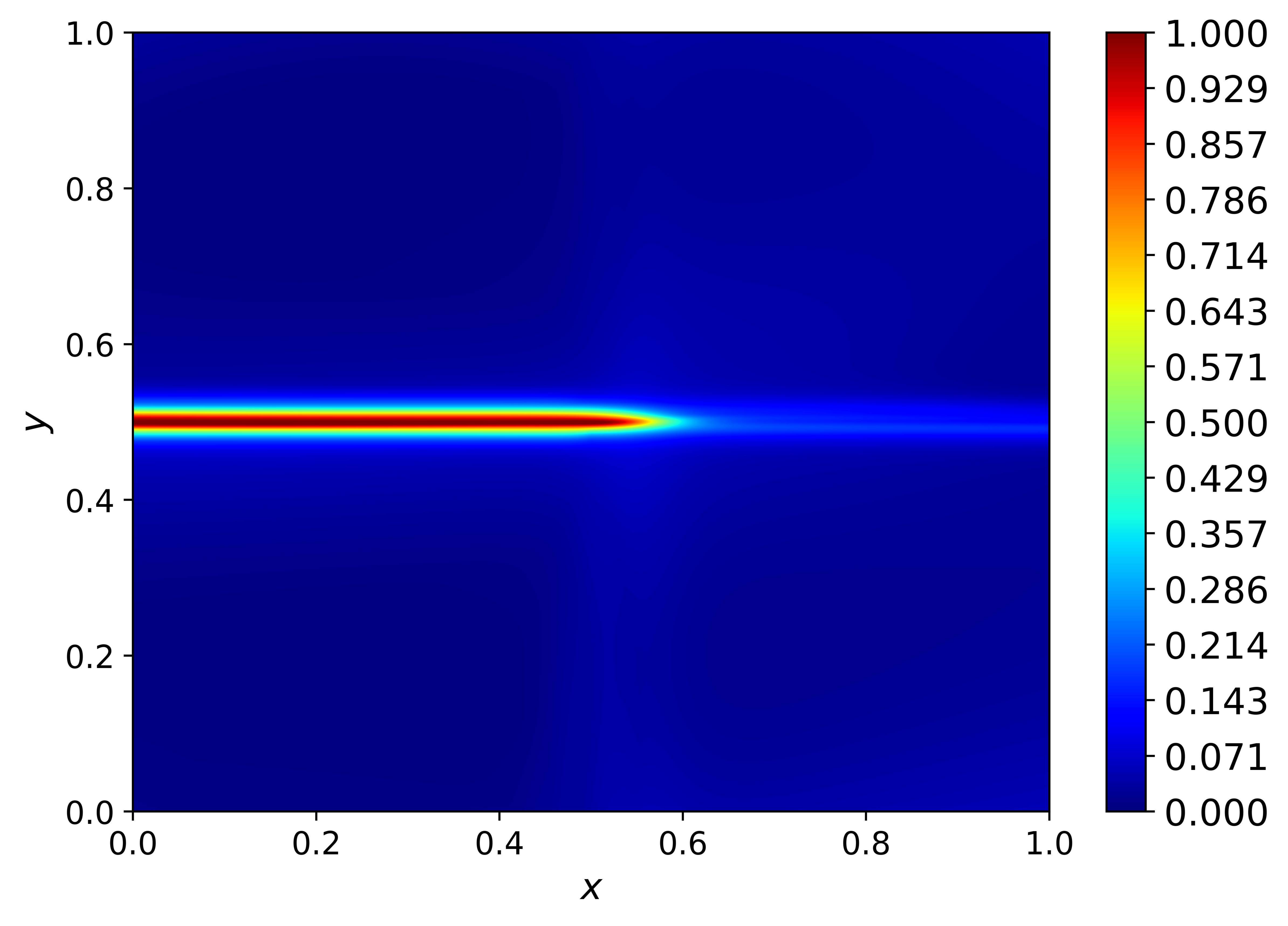}}
   \subfigure[]{
   \includegraphics[width = 0.3\textwidth]{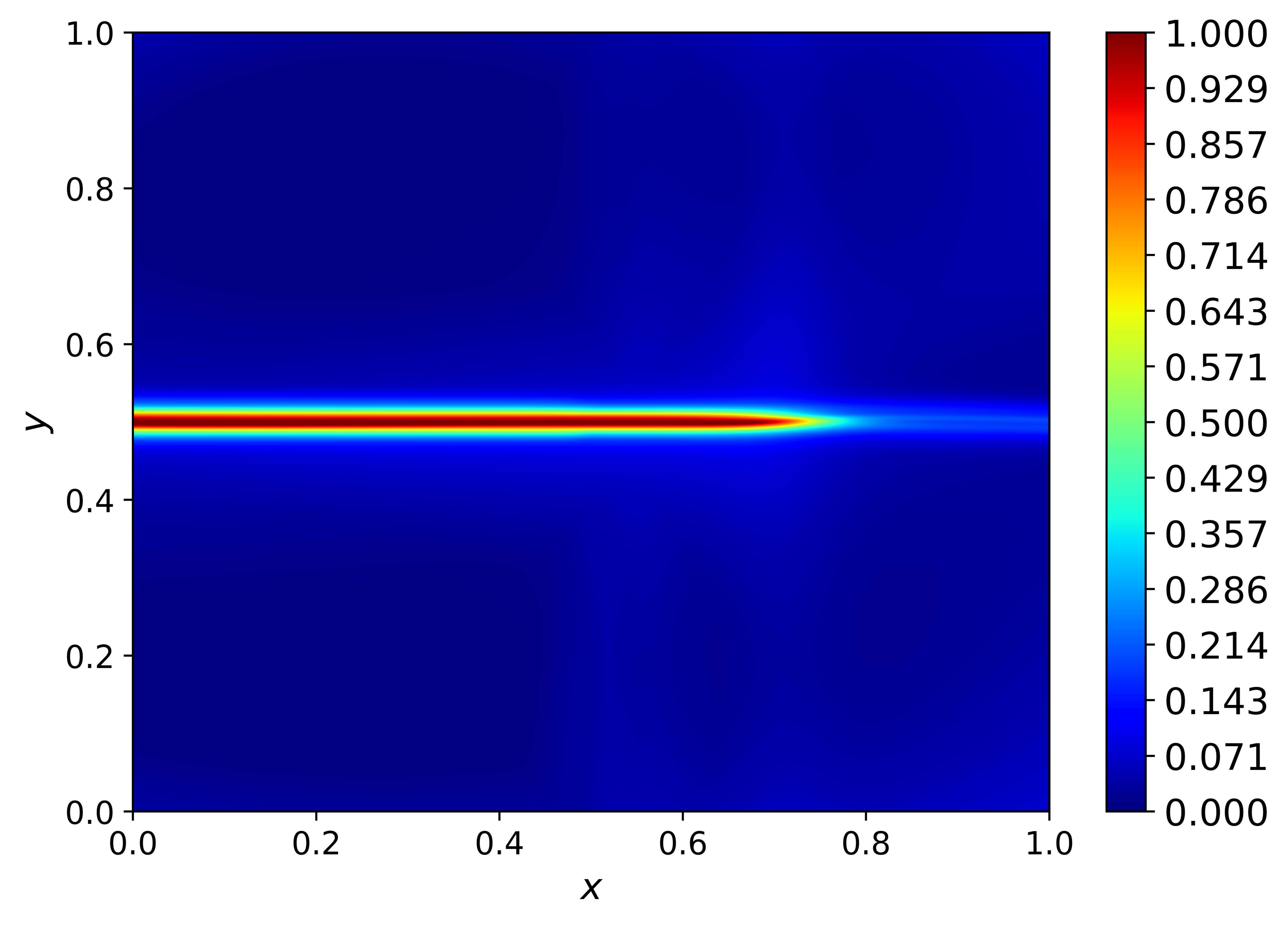}}
   \subfigure[]{
   \includegraphics[width = 0.3\textwidth]{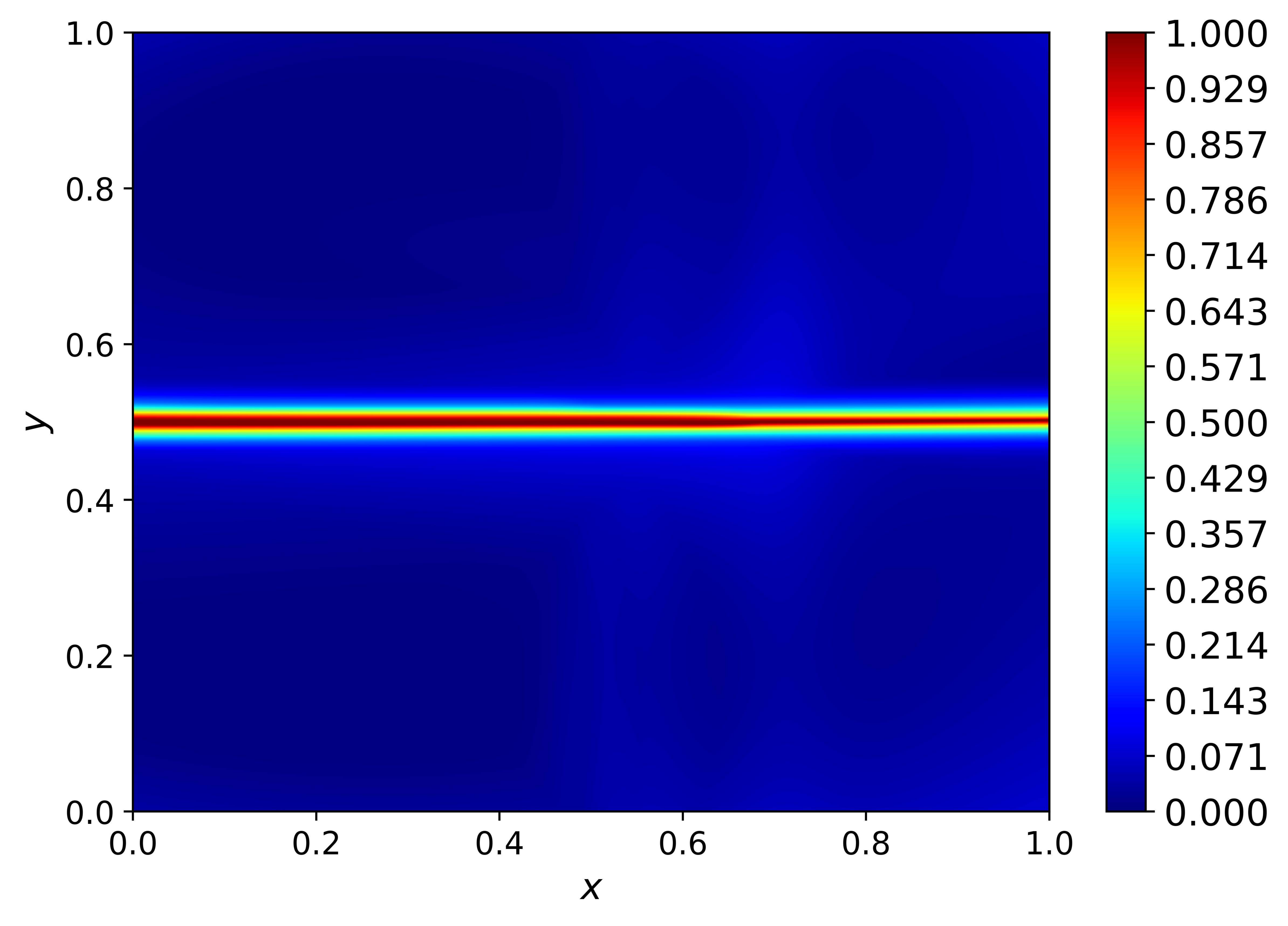}}
    \caption{Predicted crack pattern for prescribed displacement of (a) $1\times10^{-3}$, (b) $2\times10^{-3}$, (c) $3\times10^{-3}$, (d) $4\times10^{-3}$, (e) $4.5\times10^{-3}$ and (f) $5\times10^{-3}$.}
   \label{fig:phasefieldPlots}
\end{figure}

\begin{figure}[t]
    \centering
    \subfigure[]{
    \includegraphics[width = 0.26\textwidth]{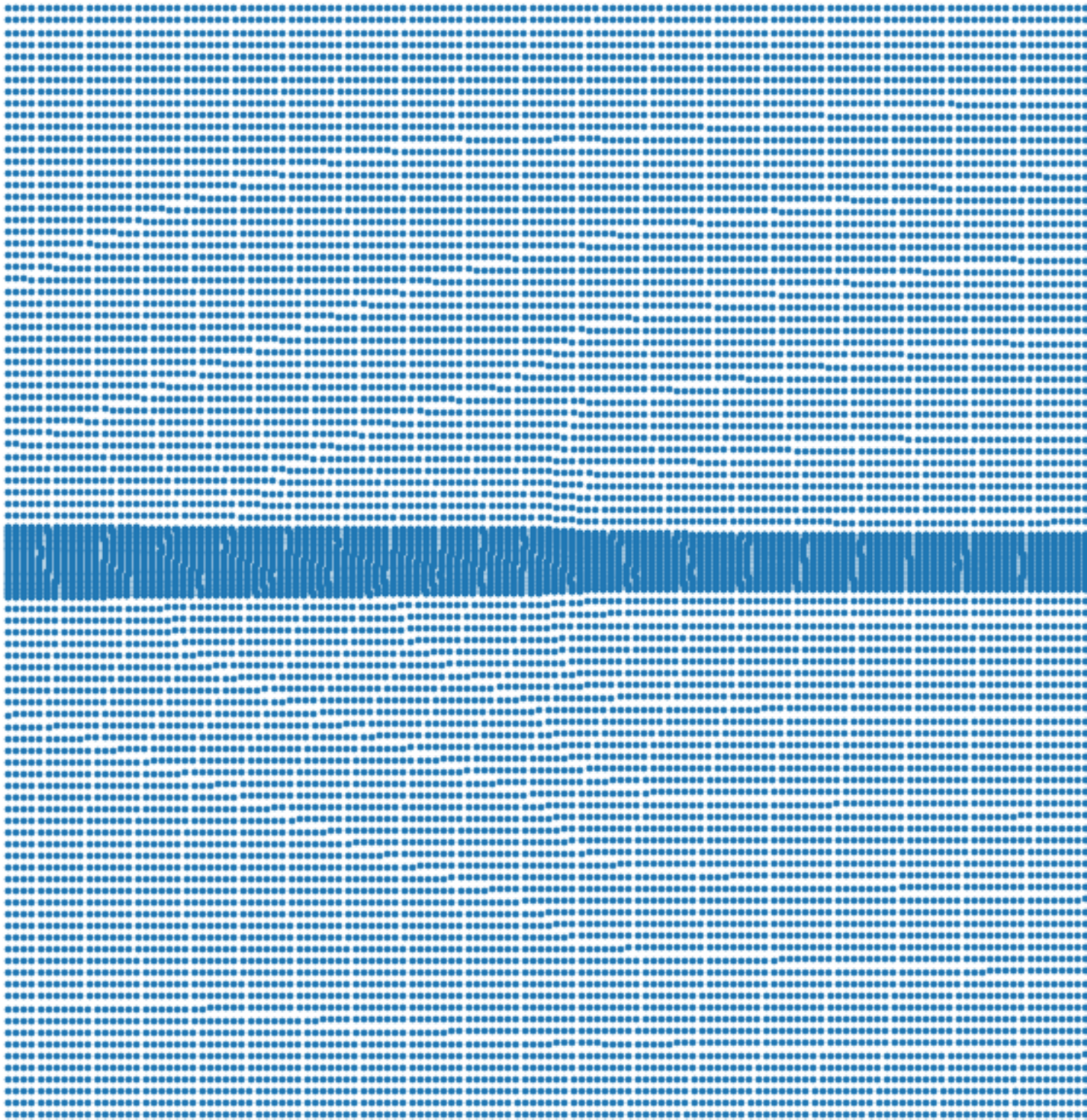}}
    \subfigure[]{
    \includegraphics[width = 0.26\textwidth]{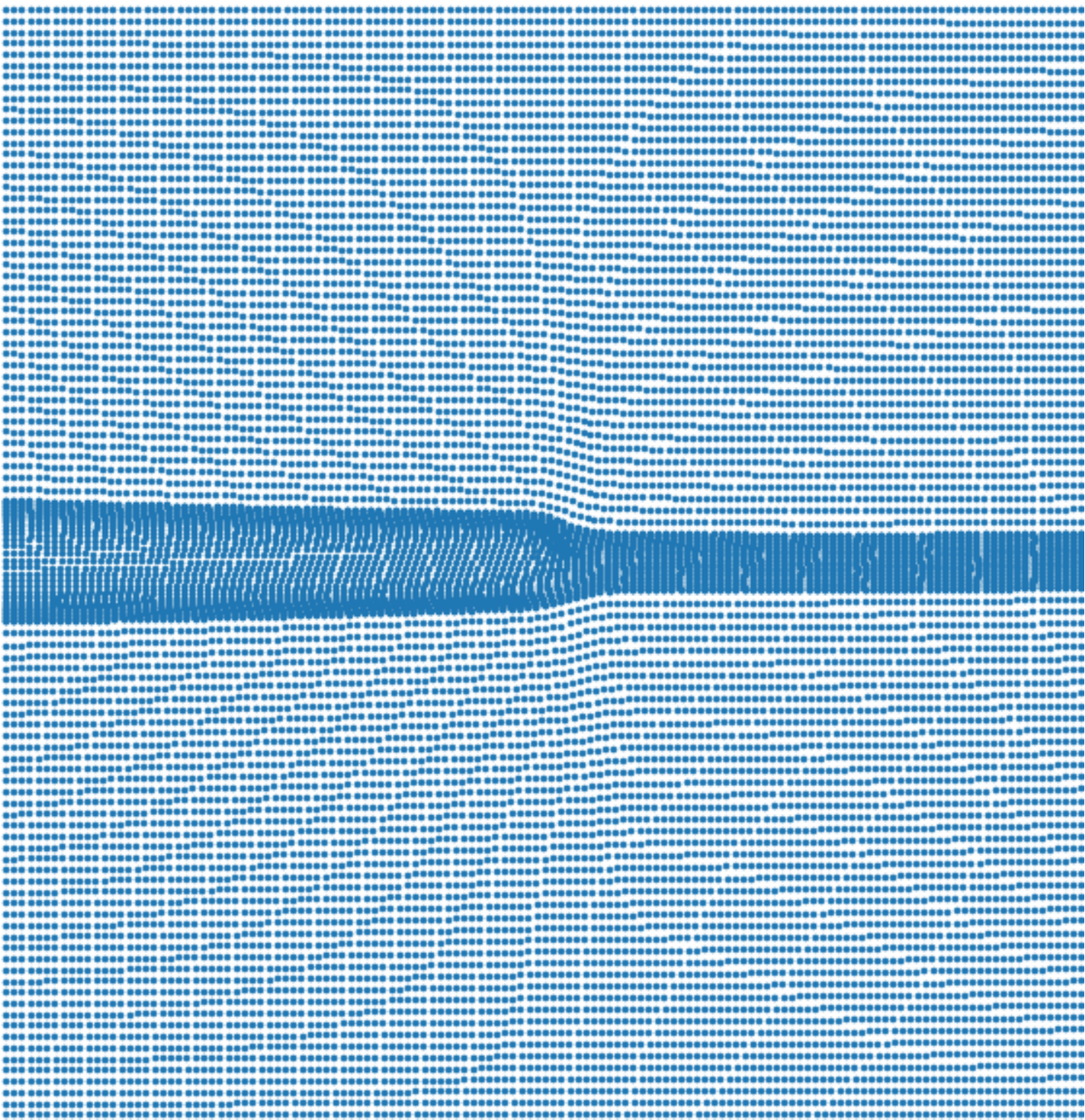}}
    \subfigure[]{
    \includegraphics[width = 0.26\textwidth]{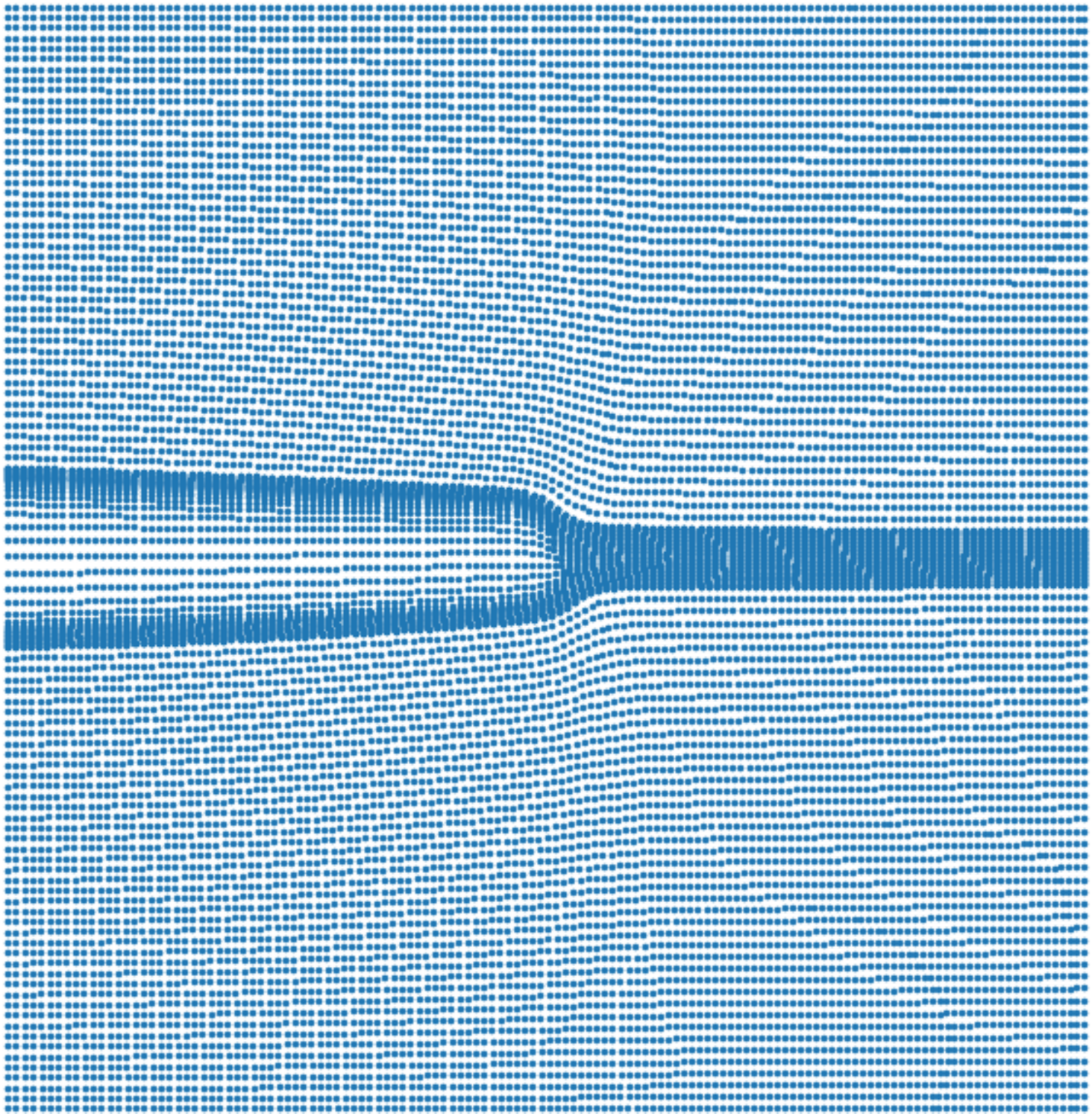}}
    \subfigure[]{
    \includegraphics[width = 0.26\textwidth]{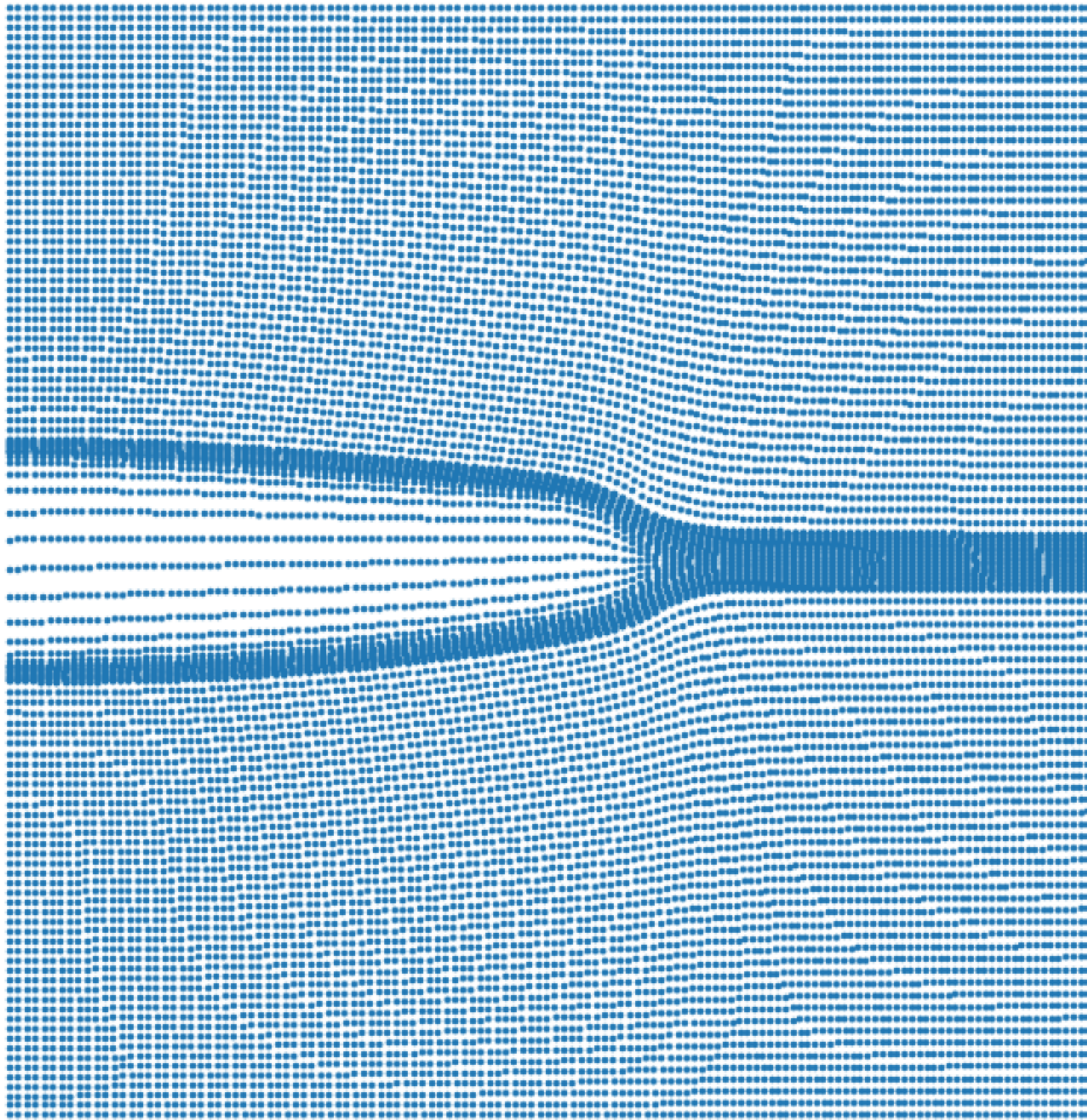}}
    \subfigure[]{
    \includegraphics[width = 0.26\textwidth]{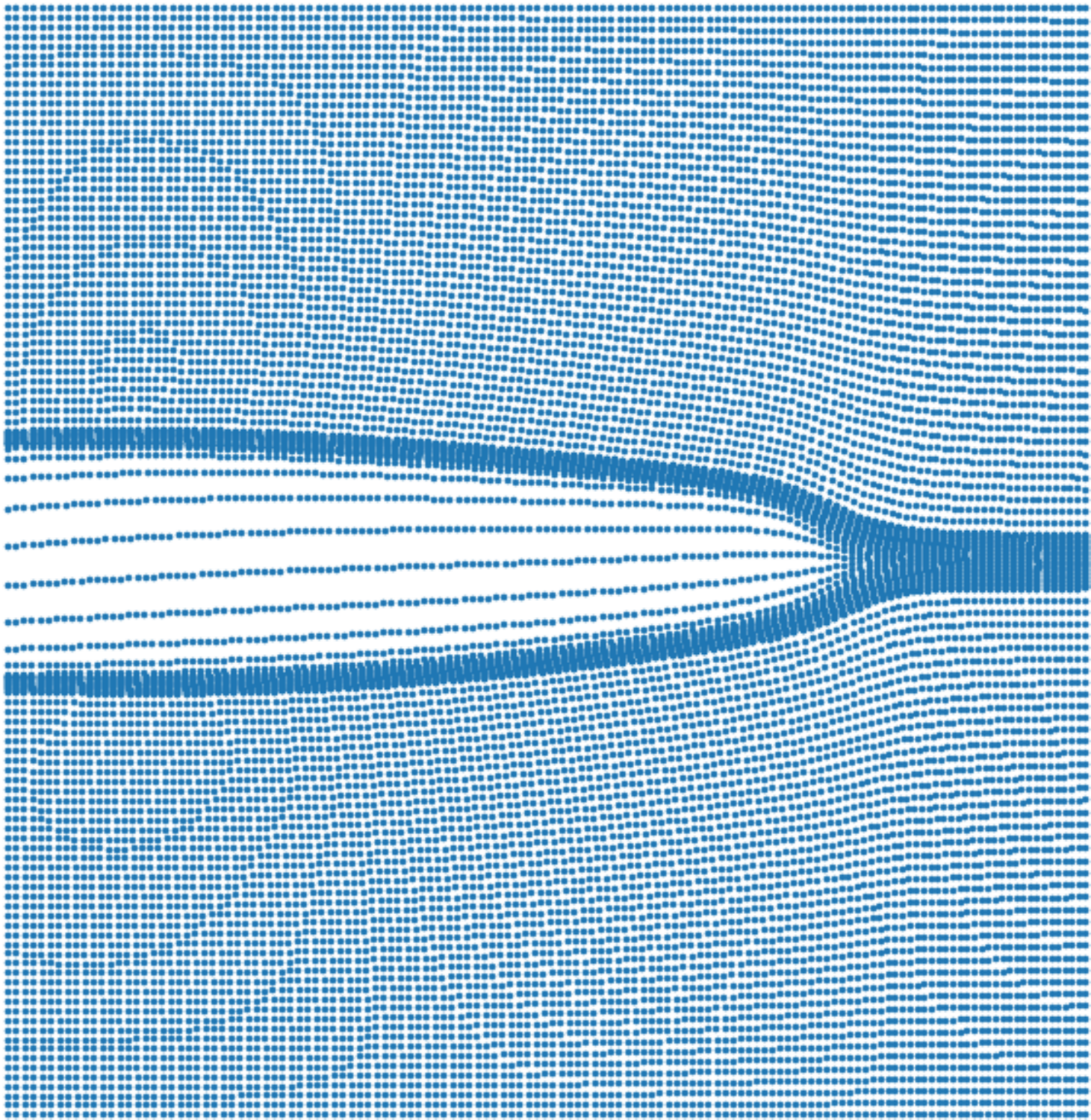}}
    \subfigure[]{
    \includegraphics[width = 0.26\textwidth]{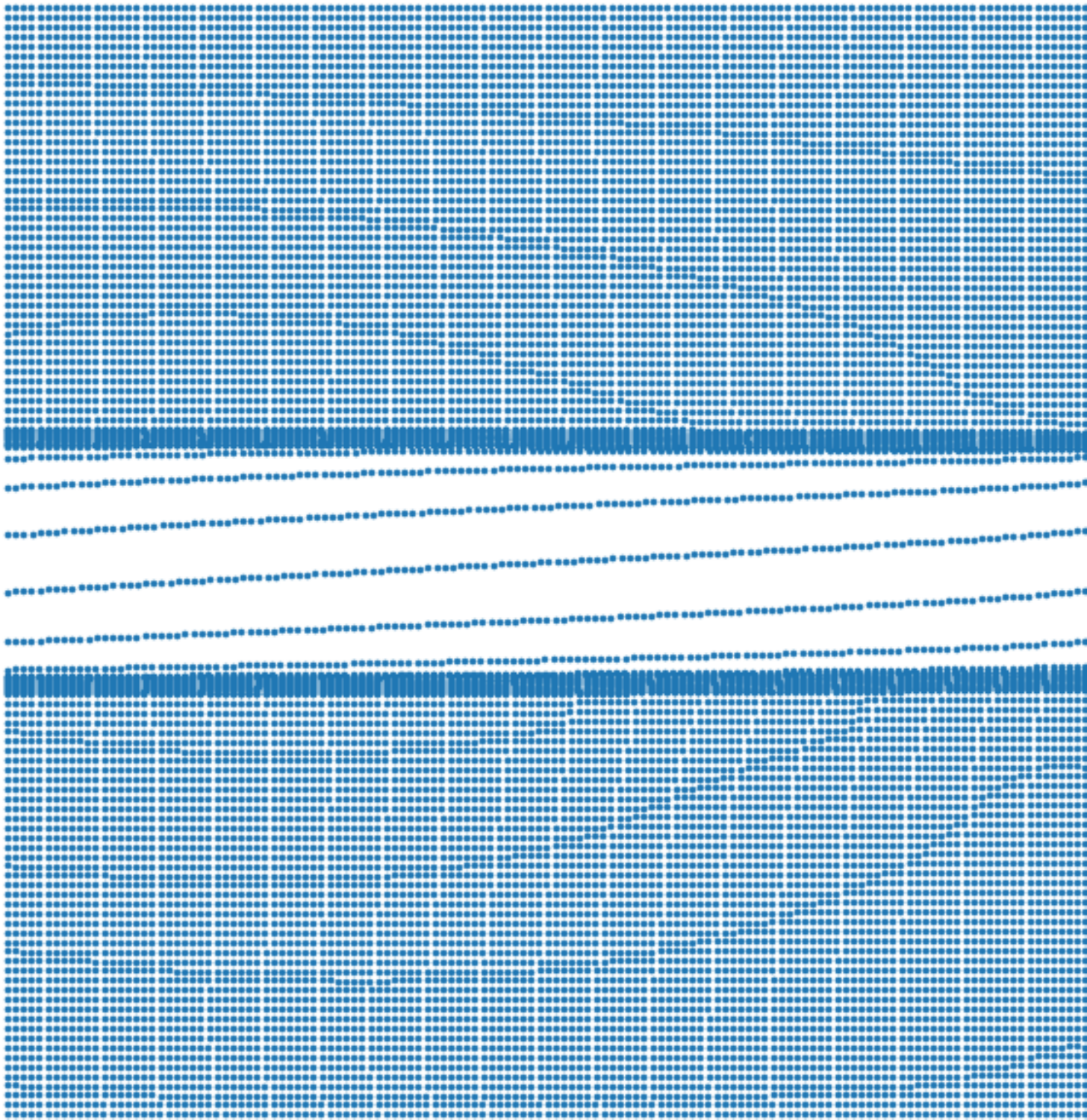}}
    \caption{Scatter plots of the deformed configuration for prescribed displacement of (a) $1\times10^{-3}$, (b) $2\times10^{-3}$, (c) $3\times10^{-3}$, (d) $4\times10^{-3}$, (e) $4.5\times10^{-3}$ and (f) $5\times10^{-3}$.}
    \label{fig:crackflow}
\end{figure}

The propagation of crack at certain selected  displacement are shown in \autoref{fig:phasefieldPlots}.
We note that, unlike the previous example, no analytical solutions exist for this problem. 
Therefore, we compare the results with those available in \cite{natarajan2019fenics}.
To show the crack growth, the scatter plots of the deformed configuration at certain selected displacement step are shown in \autoref{fig:crackflow}.
As expected, with increase in displacement the crack width increases.
The failure load in this case is reported to be $670 N$, which is extremely close to the failure load of $687 N$ reported in \cite{natarajan2019fenics}. Moreover, the load increment used is significantly larger.
We note that while the results reported in \cite{natarajan2019fenics} is obtained by discretizing the plate into 1,31,071 triangular elements, here we have only used 960 subdivisions.

For a comparative assessment, we attempted to generate the results using the residual based PINN.
However, even after several trials by varying the network architecture, the number of 
integration points and the number of iterations, the residual based formulation never converged.
This is probably because of the fact that the residual based formulation is unable to 
capture the sharp discontinuity in the system. 

As the previous example, we also illustrate the advantage of using the Gauss-Legendre rule
instead of uniformly distributed integration points.
It is observed that to obtain results of similar accuracy as those
obtained with Gauss-Legendre rule, we need to modify the neural
network architecture.
To be specific, we require two additional hidden layers
of 50 neurons each.
Moreover, we need to divide the overall domain into three parts: 
(a) upper crack zone ($[0.0,1.0]\times [0.5+2l_0, 1.0]$), (b) crack zone ($[0.0,1.0]\times [0.5-2l_0, 0.5+2l_0]$) and (c) lower crack zone ($[0.0,1.0]\times [0.0, 0.5-2l_0]$) with
each sub-domain having $300 \times 81$ uniformly spaced integration points.
This clearly indicates the advantage of using the Gauss-Legendre rule over the uniformly distributed integration points.

\begin{figure}[t]
    \centering
    \subfigure[]{
    \includegraphics[width = 0.3\textwidth]{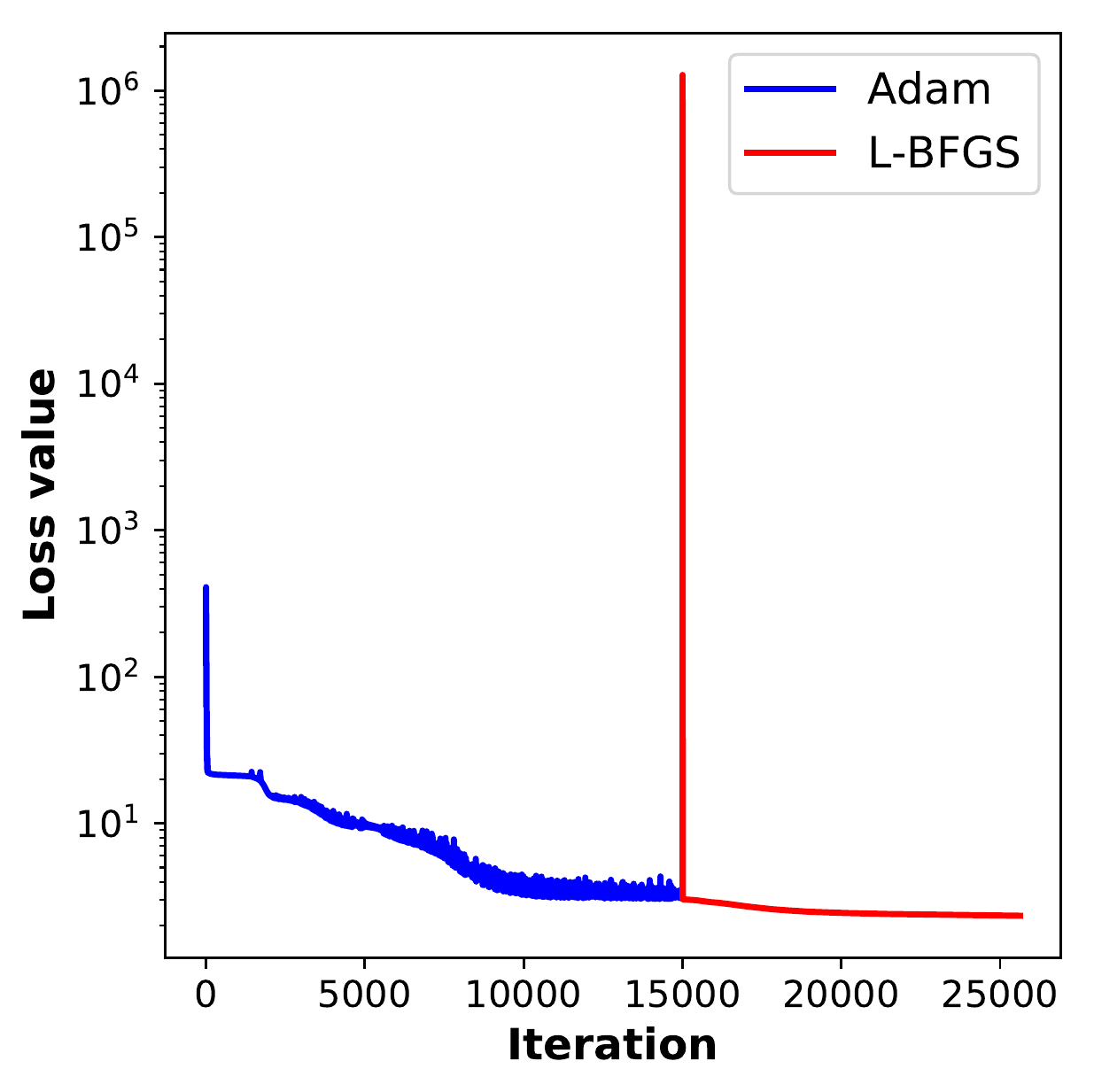}}
    \subfigure[]{
    \includegraphics[width = 0.3\textwidth]{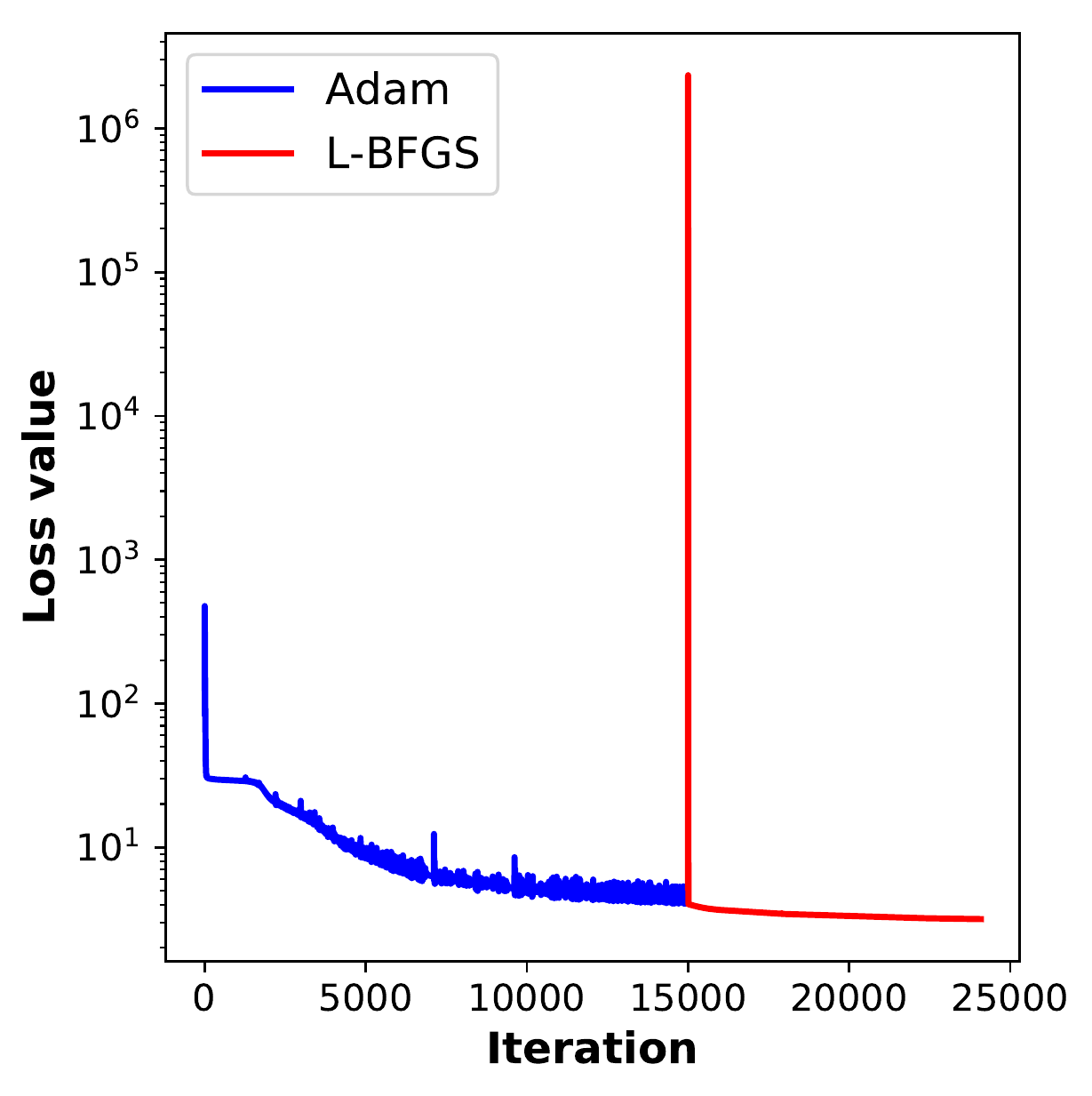}}
     \subfigure[]{
    \includegraphics[width = 0.3\textwidth]{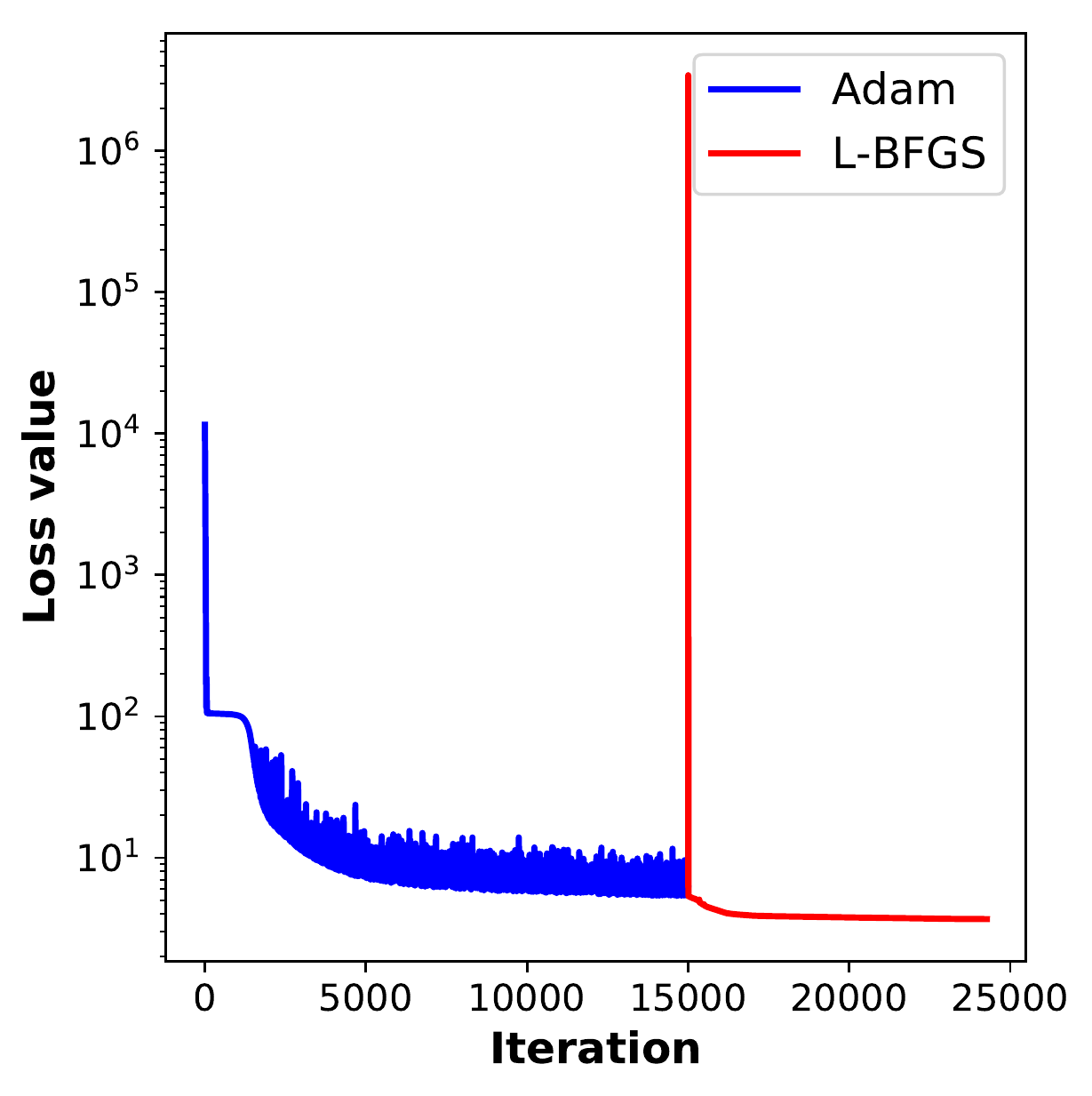}}
    \subfigure[]{
    \includegraphics[width = 0.3\textwidth]{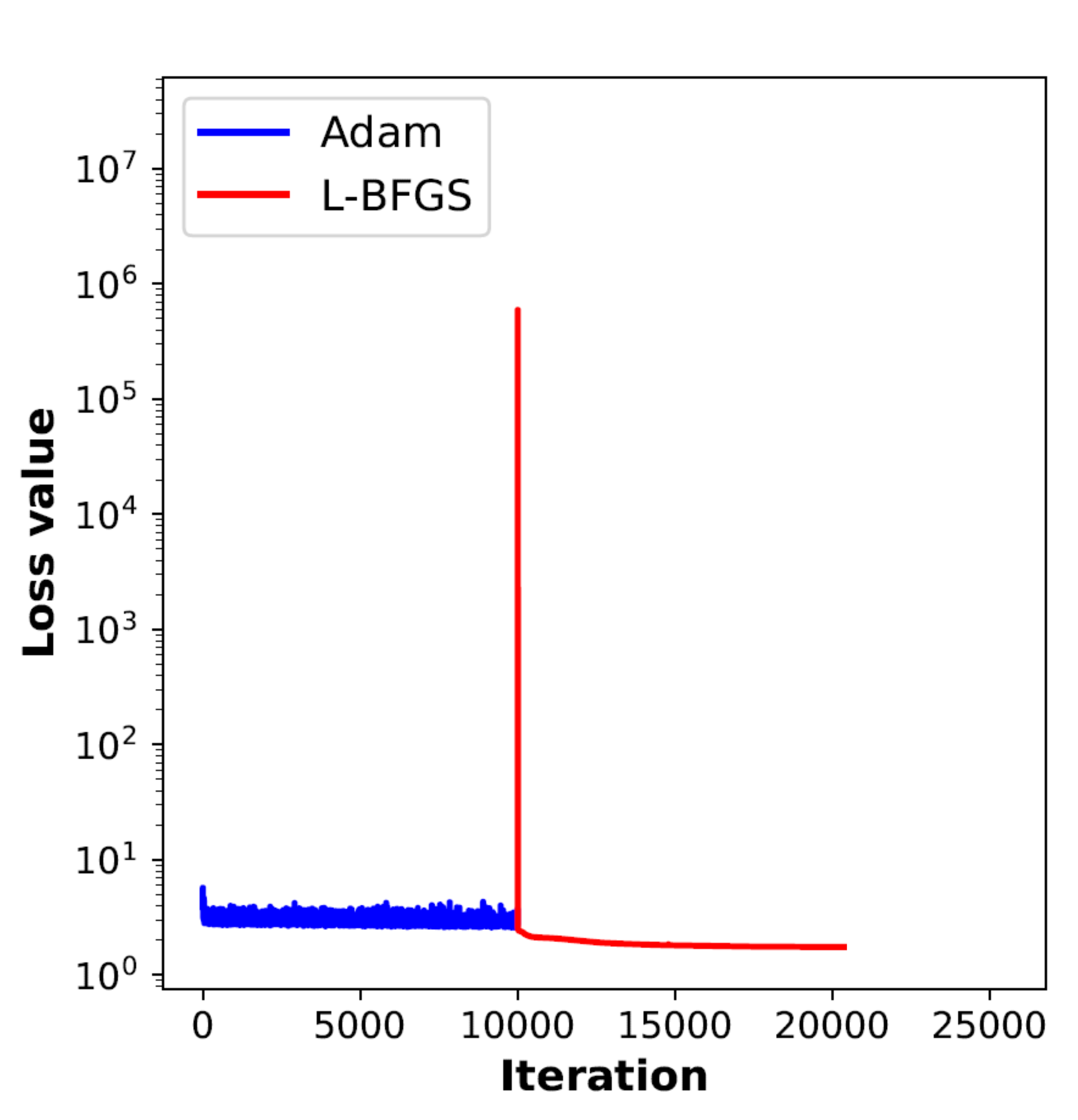}}
    \subfigure[]{
    \includegraphics[width = 0.3\textwidth]{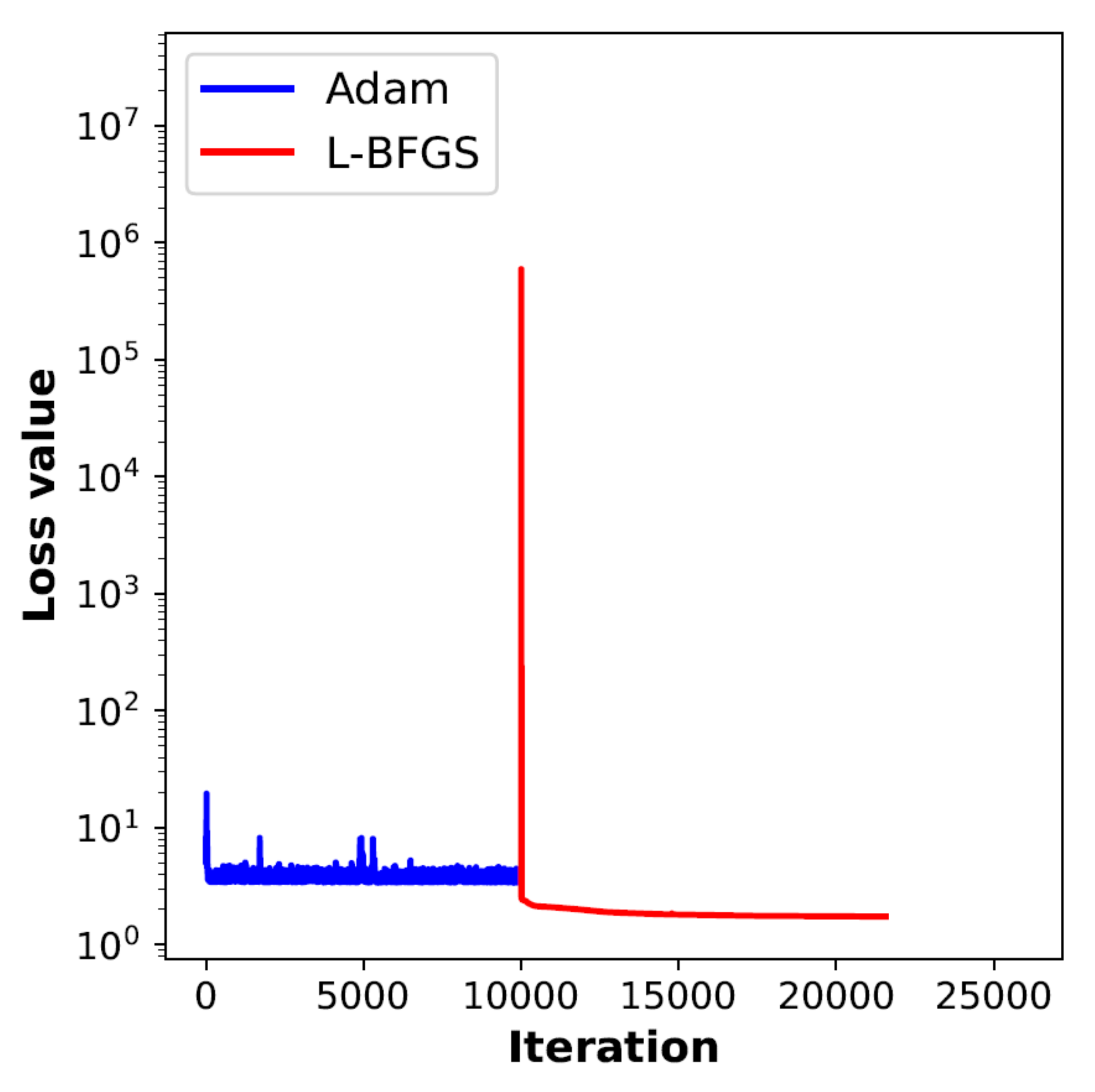}}
     \subfigure[]{
    \includegraphics[width = 0.3\textwidth]{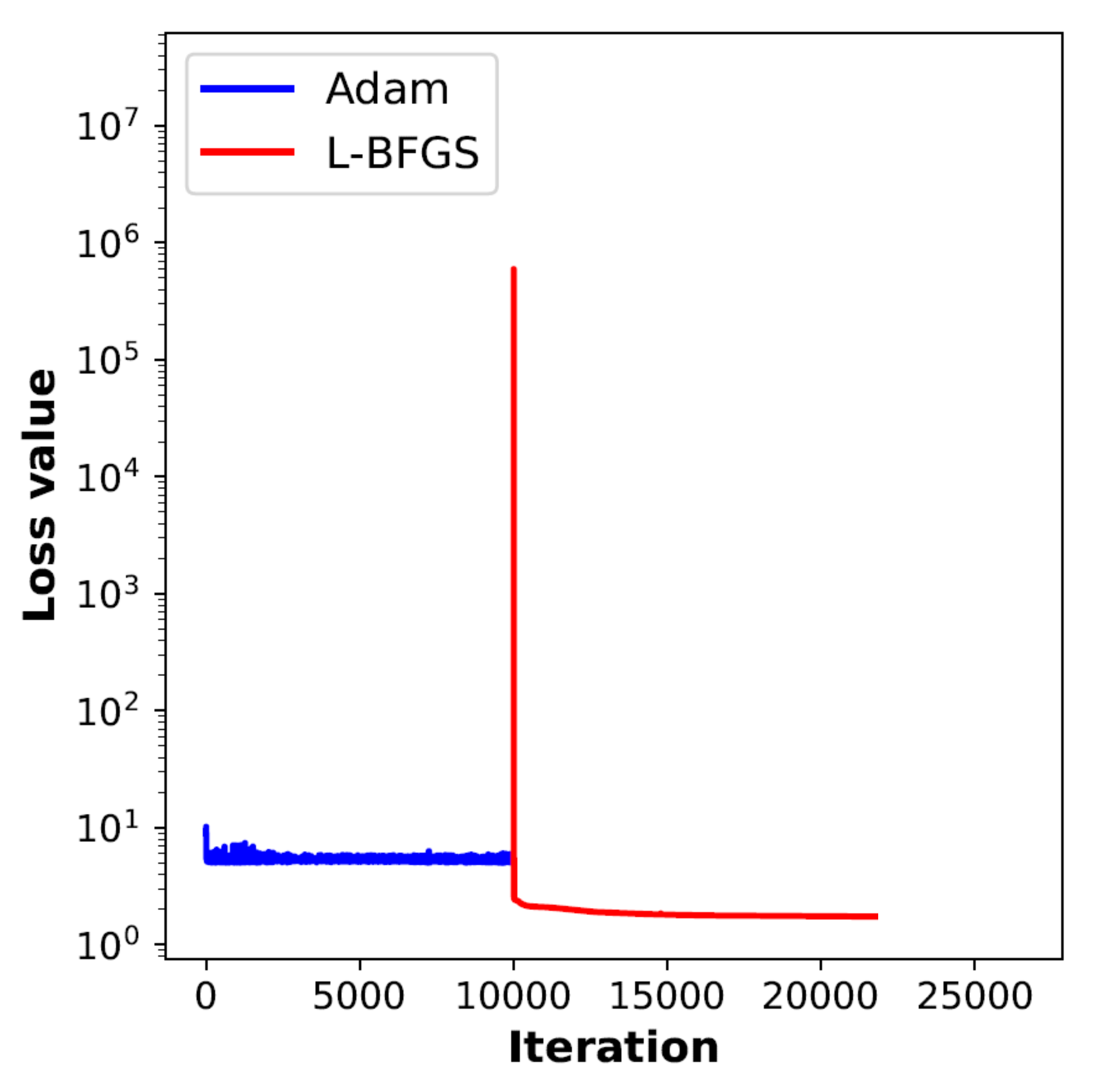}}
    \caption{Plots of convergence of the loss function. The top row shows the plots of convergence without the reuse of the parameters from the previous iteration, the bottom row shows the plots of convergence with the reuse of the parameters. The plots presented are for prescribed displacement of (a) $2\times10^{-3}$, (b) $3\times10^{-3}$ and (c) $5\times10^{-3}$ from left to right.}
    \label{fig:tension_convPlots}
\end{figure}

Finally, we note that the solution of this problem using the proposed PINN
involves repeated training of the neural network.
As already stated, this makes the process computationally expensive.
In order to address this issue, we utilize the transfer learning approach
as discussed in \autoref{sec:phase_pinn}.
To illustrate the advantage gained by using transfer learning, 
the convergence of the loss function, with and without transfer learning,
are shown in \autoref{fig:tension_convPlots}.
We observe that with transfer learning, the algorithm
converges quickly and, on convergence, yields a lower loss function (i.e., variational energy).
Moreover, the computational time per iteration is halved while using transfer learning.

For clarity of readers, the summary of results obtained are shown in \autoref{tab:prob2}.

\begin{table}[htbp!]
    \centering
    \caption{Summary of results for problem 2. We observe among the three PINNs, the proposed approach was the best results.}
    \label{tab:prob2}
    \begin{tabular}{lccc}
    \hline
    \textbf{Methods} & \textbf{PINN Architecture} & \textbf{Integration points} & \textbf{Failure load (N)} \\ \hline
    Benchmark \cite{Miehe2010} & -- & -- & 687 \\ \hline \hline 
    VE-PINN$^*$ & [2, 50, 50, 50, 3] & 61,440 $\left( 960 \times 64 \right)$ & 670 \\
    R-PINN$^{\#}$ & \multicolumn{3}{c}{did not converge} \\
    VE-PINN2$^{\dagger}$ & [2, 50, 50, 50, 50, 50, 3] & 72,900 $\left( 300 \times 81 \times 3 \right)$ & 820 \\ \hline
     \multicolumn{4}{l}{\small $^*$VE-PINN = variational energy based PINN (proposed approach)} \\
         \multicolumn{4}{l}{\small $^{\#}$R-PINN = residual based PINN (conventional approach)} \\
         \multicolumn{4}{l}{\small $^{\dagger}$VE-PINN2 = Same as VE-PINN, but with uniformly distributed integration points} 
    \end{tabular}
\end{table}

\subsection{Perforated and notched asymmetric bending example}
\label{sec:3holes}
\begin{figure}[htbp!]
    \centering
    \centering
    \subfigure[Geometrical setup and boundary conditions.]{
    \includegraphics[width = 0.5\textwidth]{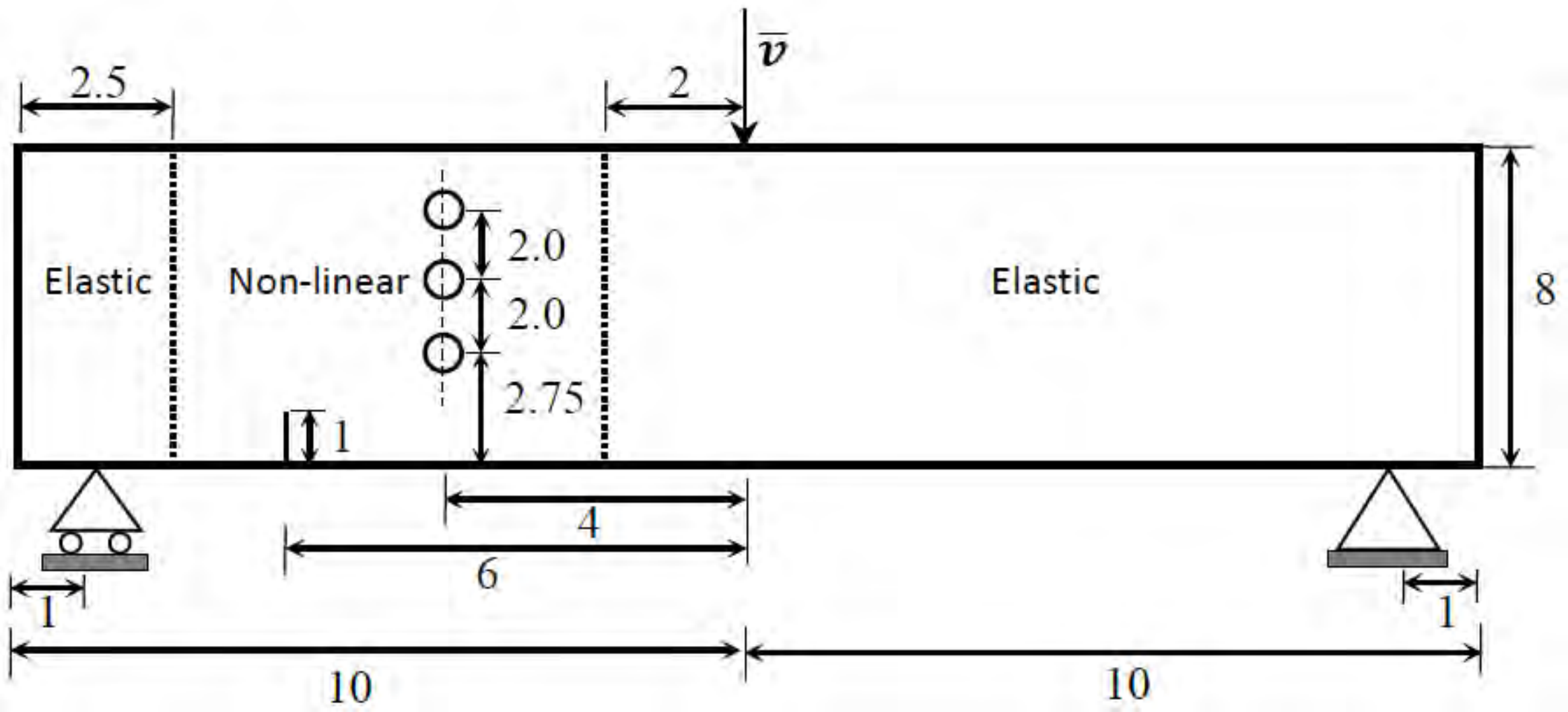}}
    \subfigure[Modeled geometry using NURBS.]{
    \includegraphics[width = 0.45\textwidth]{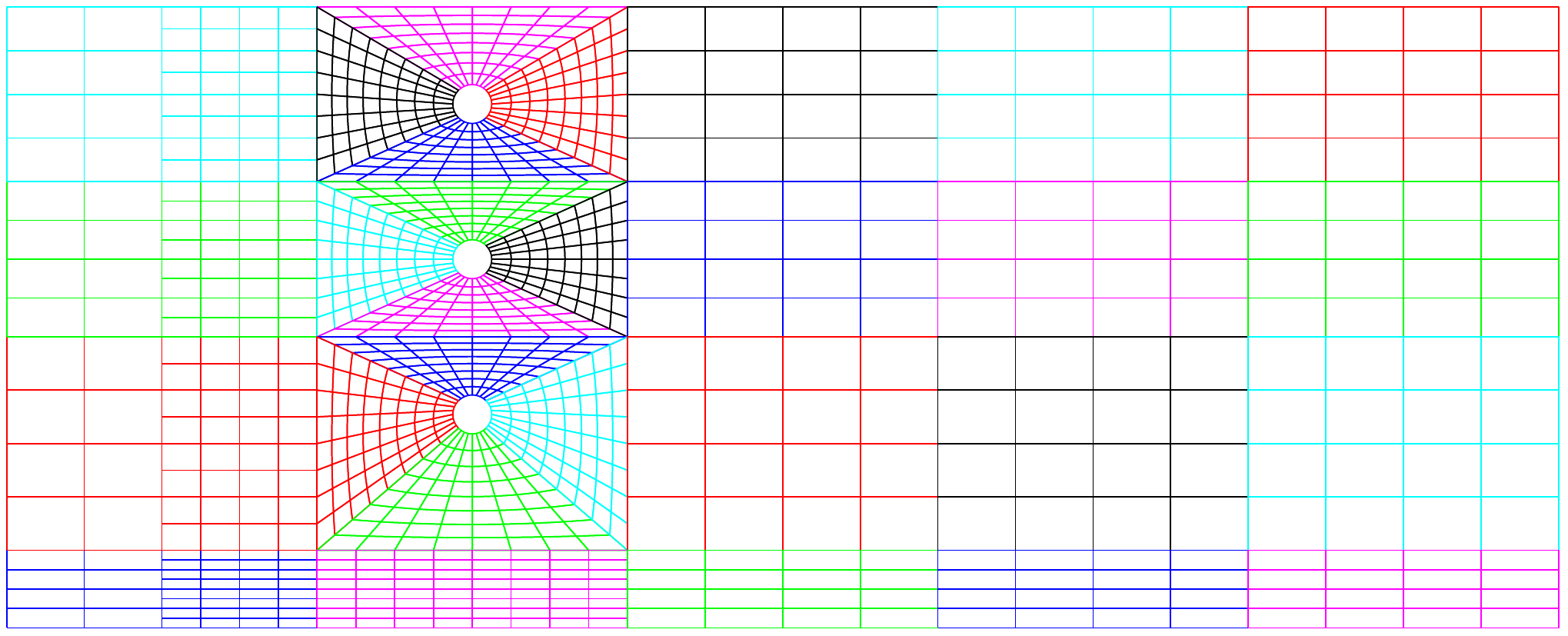}}
    \caption{Perforated and notched asymmetric bending example. The diameter of the holes are 0.5.}
  \label{fig:setup_3holes}
\end{figure}

In this example, we consider the `perforated and notched asymmetric three point bending' problem.
This is a well-known problem and has been previously analyzed both experimentally \cite{bittencourt1996quasi} and numerically \cite{Miehe2010a}.
It concerns an asymmetrically-notched beam with three holes.
The geometrical setup and the boundary conditions for this example are depicted in \autoref{fig:setup_3holes}(a).
This example has been typically chosen to show the performance of the proposed approach in predicting curved crack trajectories.
Also, it illustrates the effectiveness of using NURBS for generating the geometry as the exact boundaries of the holes in the plate
could be efficiently modeled.
The domain is modeled with 29 NURBS patches which are subsequently refined along the path of expected crack propagation.
The model generated using NURBS is shown in \autoref{fig:setup_3holes}(b). The material parameters are considered as $\lambda = $ 12.0 kN/mm$^{2}$, $\mu = $8.0 kN/mm$^{2}$, $G_c = 1 \times 10^{-3}$ kN/mm and $l_0 = 0.25$. 
Similar to \cite{molnar20172d}, the left and right portion of the system are considered to be elastic.

We have considered a fully connected deep neural network with 3 hidden layers.
Each hidden layer has 50 neurons. For the first two layers, we have considered the \texttt{tanh} activation function; whereas for the last layer, linear activation
function has been considered. The plate has been discretized into 1184 elements with each element having 25 Gauss points.
The crack is initiated using the strain 
history functional. The crack path is 
obtained by applying a constant displacement 
increment of $\Delta u$ = $1\times 10^{-2}$. 
The Dirichlet boundary conditions for the 
problem are:
\begin{equation}
    v(1,0) = v(19,0) = u(19,0) = 0, \;\;\; v(10,8)= -\Delta v, 
\end{equation}
where $u$ and $v$ are the solutions of the 
elastic field in \textit{x} and \textit{y}-axis. To ensure that the neural network satisfies the boundary conditions exactly, we define the outputs of the elastic field as:
\begin{equation}
\begin{split}
    u &= \frac{w_2}{(w_2+1)}\hat{u},\\
    v &= \frac{w_1w_2w_3}{(w_1+1)(w_2+1)(w_3+1)}\hat{v} + \frac{y}{8}\Delta v, \\
    \text{where} \;\;\; w_1 & = (x-1)^2 + y^2,\\
    w_2 & = (x-19)^2 + y^2, \\
    w_3 & = (x-10)^2 + (y-8)^2,
\end{split}
\end{equation}
where $\hat{u}$ and $\hat{v}$ are obtained from the neural network.

\begin{figure}[htbp!]
    \centering
    \subfigure[]{
    \includegraphics[width = 0.2\textwidth]{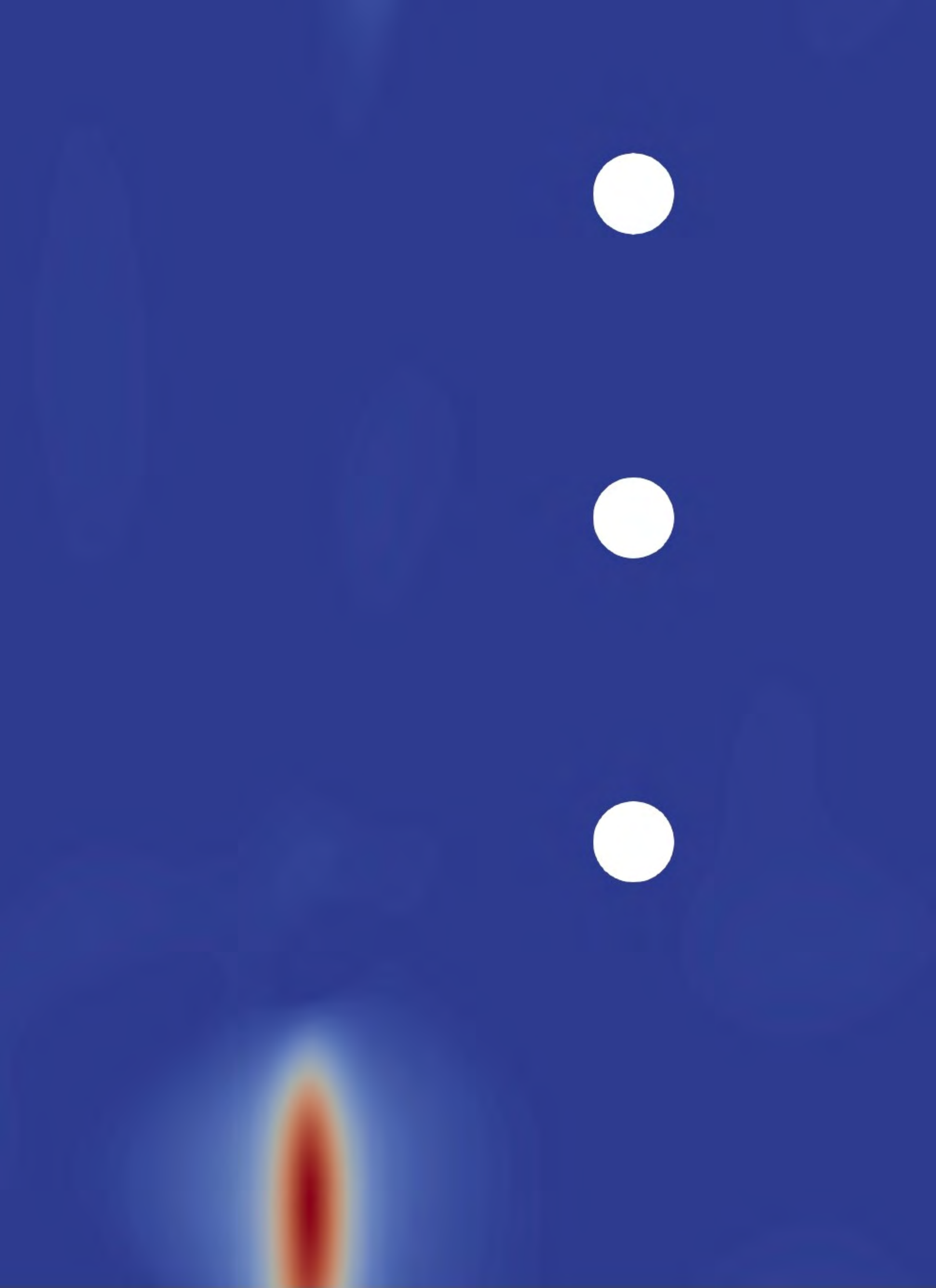}}
    \subfigure[]{
    \includegraphics[width = 0.205\textwidth]{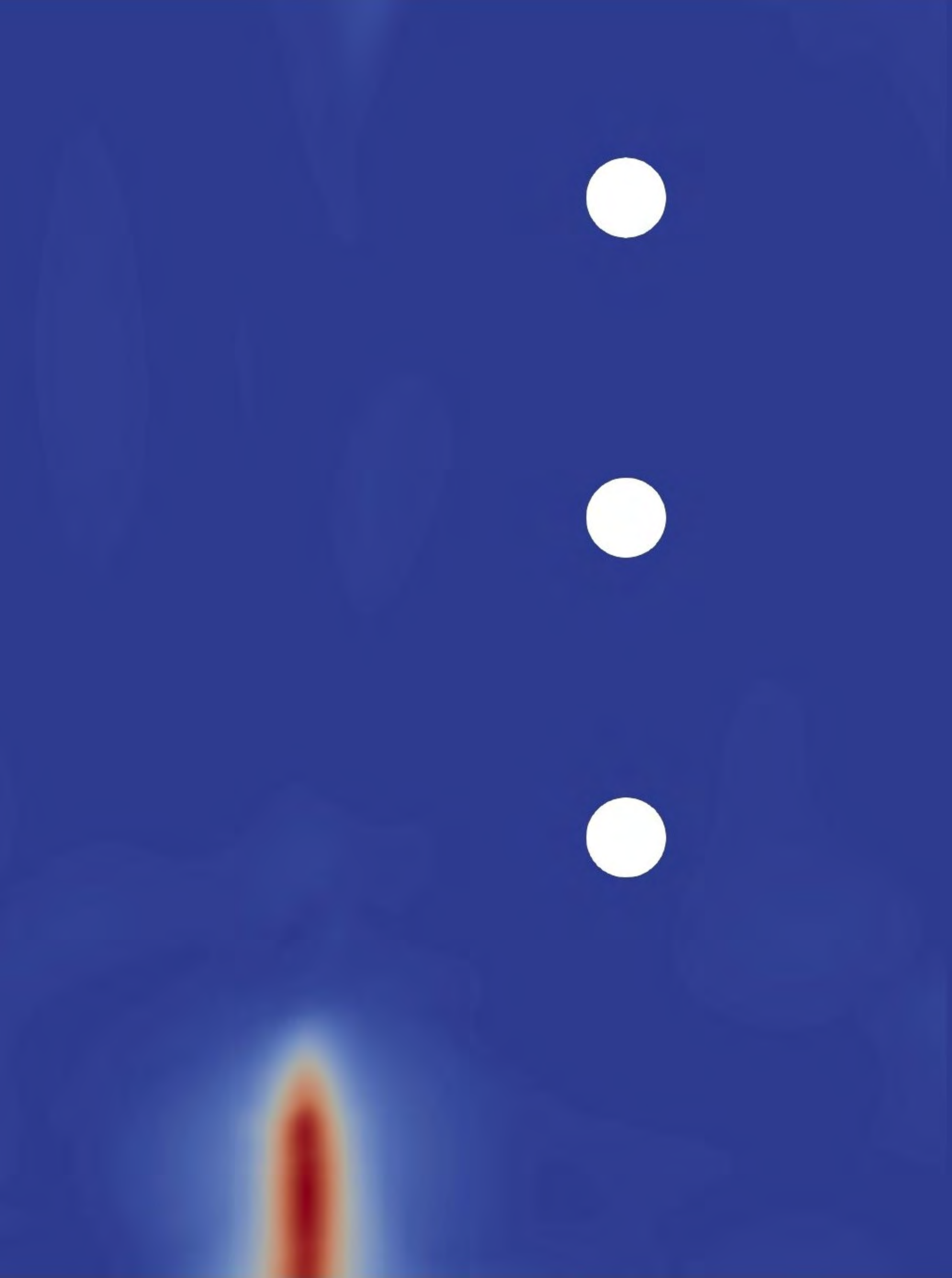}}
    \subfigure[]{
    \includegraphics[width = 0.21\textwidth]{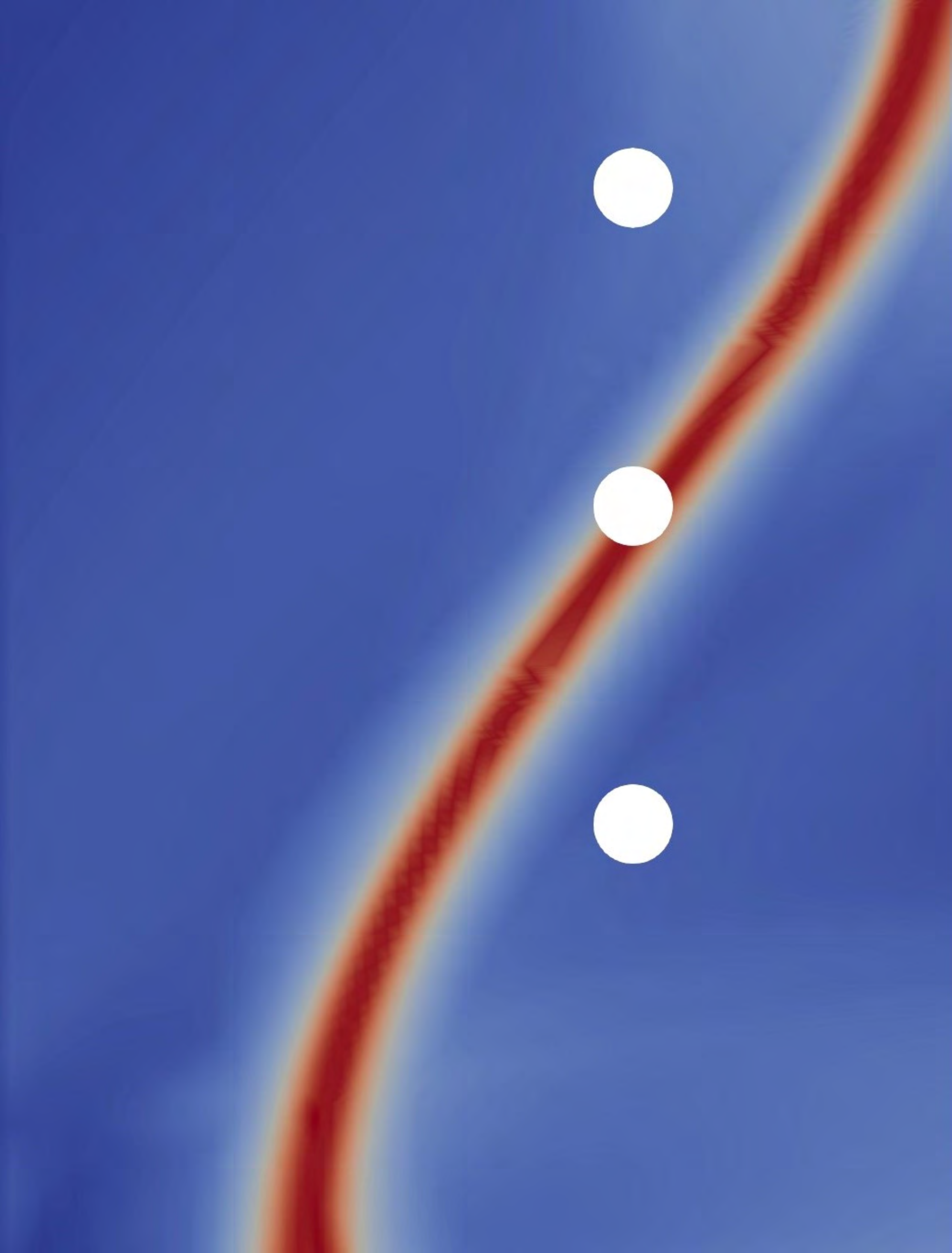}}
    \subfigure{
    \includegraphics[width = 0.4\textwidth]{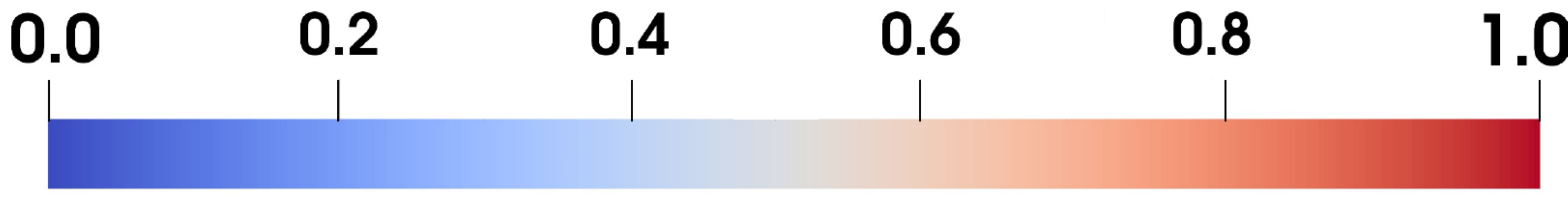}}
    \caption{Plots showing the predicted phase-field for prescribed displacement of (a) initialization of crack, (b) $1\times10^{-2}$, and (c) $16\times10^{-2}$ using the proposed PINN approach.}
    \label{fig:3holes_predphi}
\end{figure}

\begin{figure}[t]
    \centering
    \subfigure[]{
    \includegraphics[width = 0.4\textwidth]{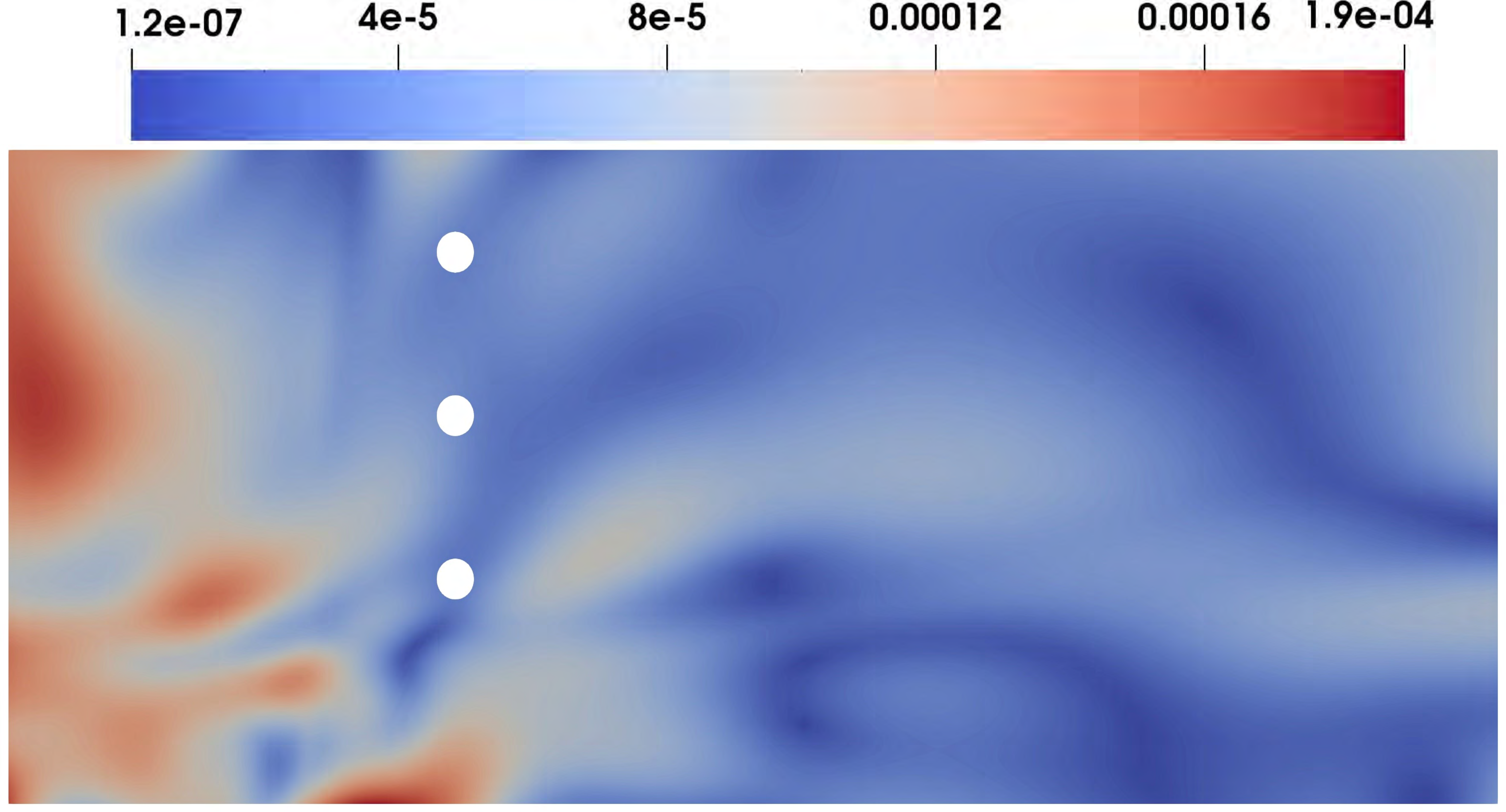}}
    \subfigure[]{
    \includegraphics[width = 0.425\textwidth]{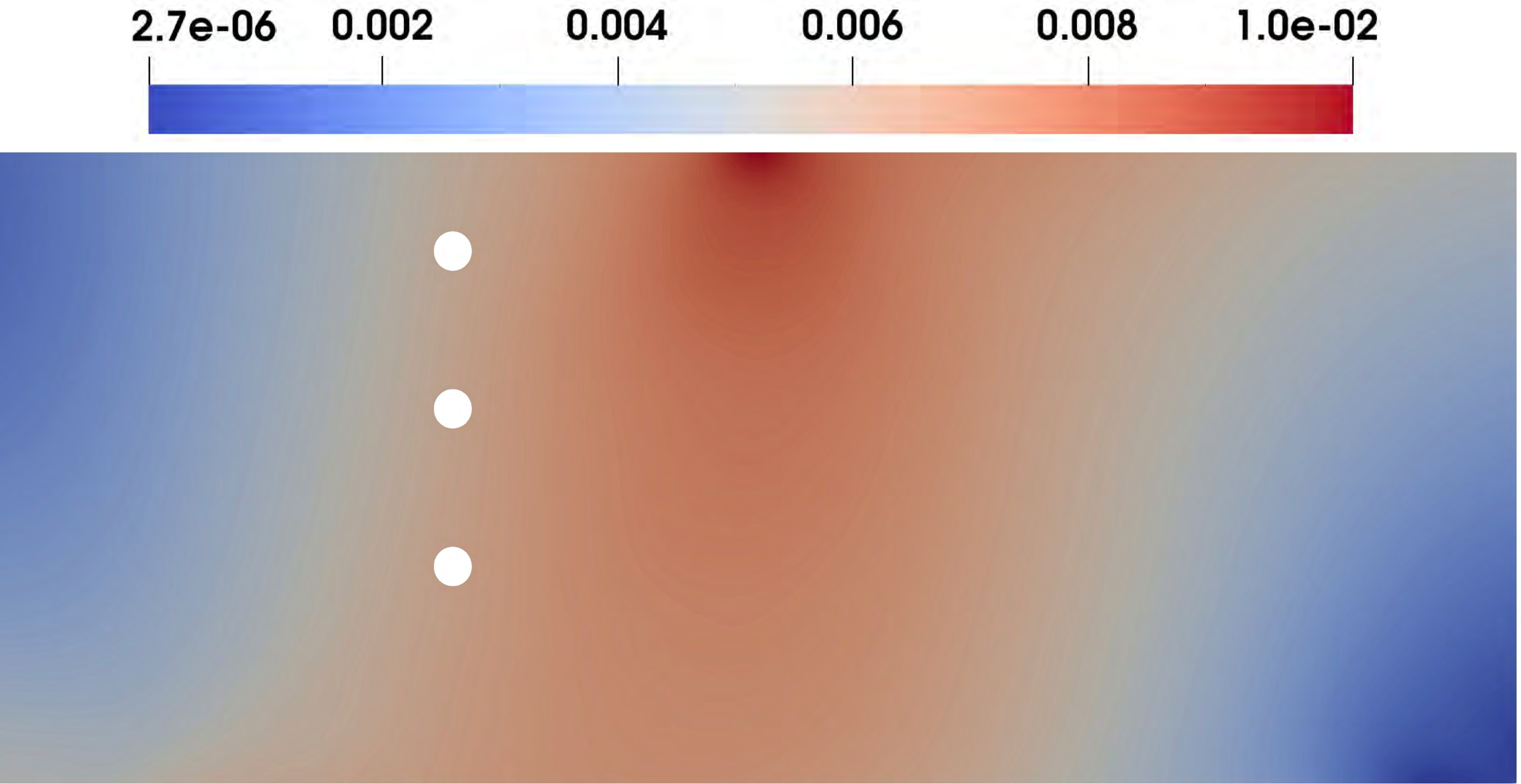}}
    \subfigure[]{
    \includegraphics[width = 0.47\textwidth]{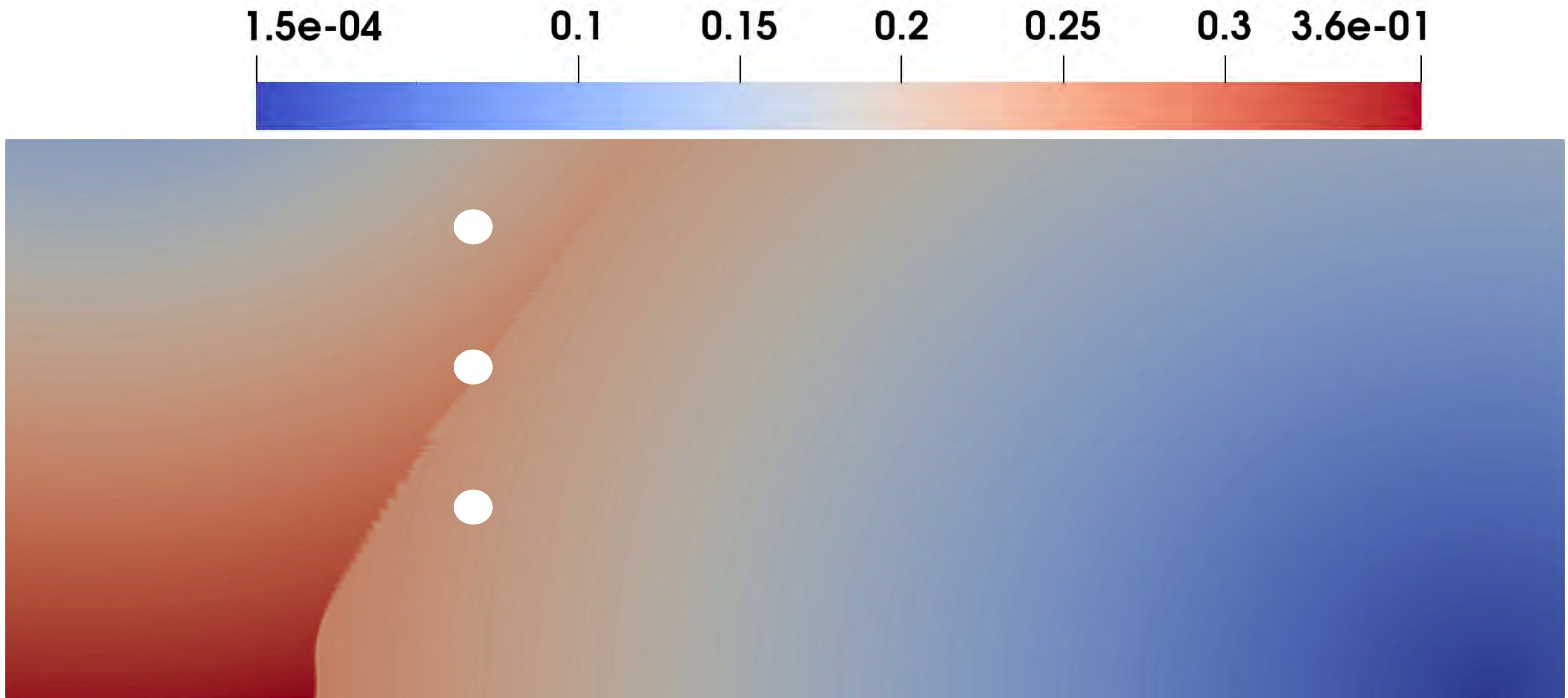}}
    \caption{Plots showing the predicted y-displacement for prescribed displacement of (a) initial crack (without any applied displacement), (b) $1\times10^{-2}$, and (c) $16\times10^{-2}$ using the proposed PINN approach.}
    \label{fig:3holes_predDisp}
\end{figure}

Figs. \ref{fig:3holes_predphi} and \ref{fig:3holes_predDisp} show the phase-field, $\phi$ and the predicted displacement field, $v$, respectively 
at certain displacement steps. 
For phase field, only the nonlinear portion is shown.
Similar to \cite{molnar20172d}, the crack propagates through the second hole. However, propagation of the crack beyond the second hole has not been reported in the literature. We also note that while \cite{molnar20172d} obtained the results with 
60,000 elements, the proposed PINN requires 1,184 elements.

\subsection{Three dimensional mode-I tension test}
\label{sec:3D_tension}
\begin{figure}[htbp!]
    \centering
    \includegraphics[width = 0.35\textwidth]{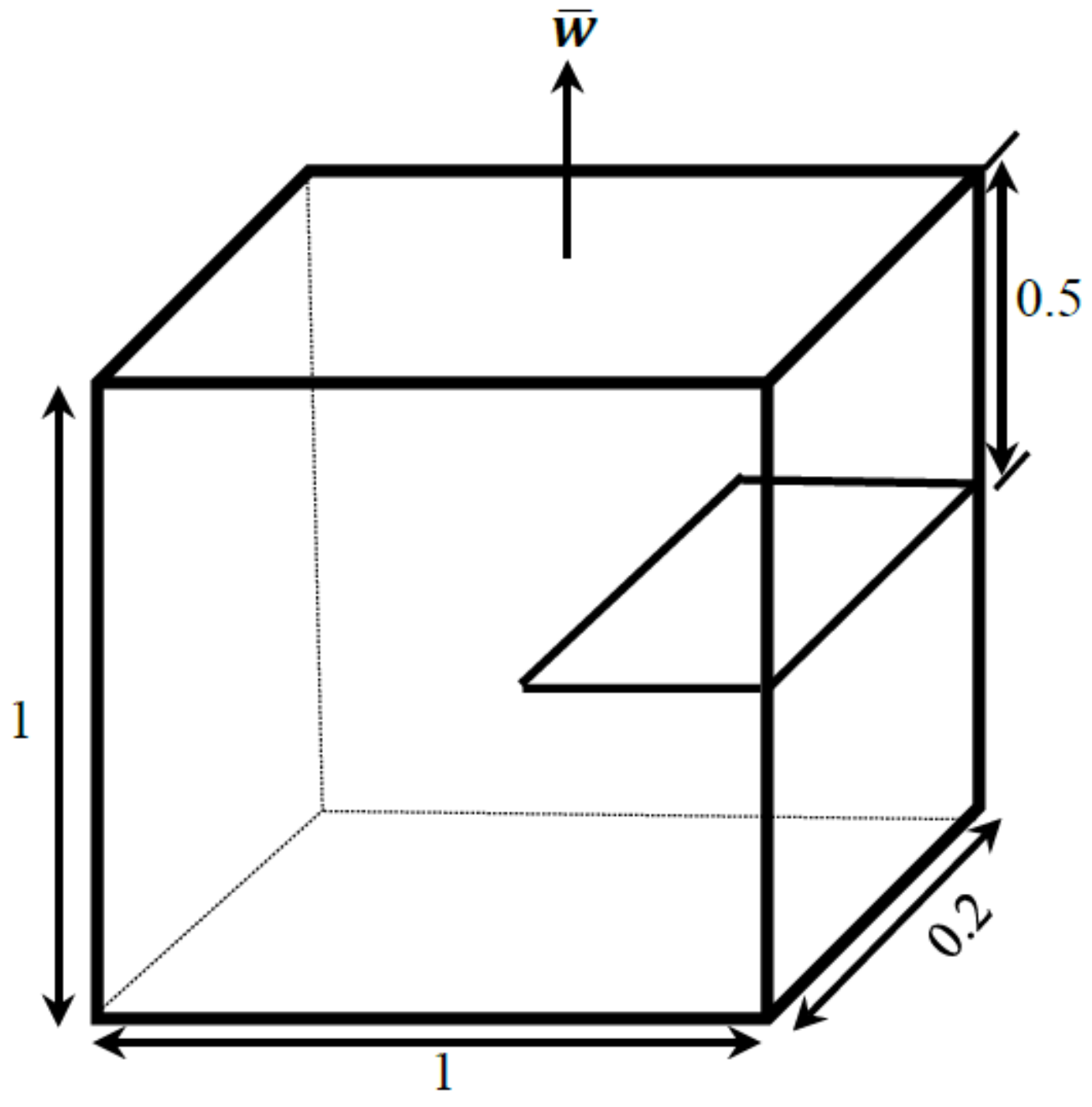}
    \caption{Geometrical setup and boundary conditions of three dimensional mode-I tension test.}
    \label{fig:3Dtension_setup}
\end{figure}

As the last example, we consider a 3D cube subjected to tensile loading.
The purpose of selecting this example is to illustrate the application of the proposed approach for a three dimensional problem. The geometric setup and the associated boundary conditions are shown in \autoref{fig:3Dtension_setup}. 
The NURBS mesh is used to generate the Gauss points. We 
consider $\lambda = $ 12 kN/mm$^{2}$, $\mu = 
$ 8 kN/mm$^{2}$, $G_c =$ 0.5$\times 10^{-3}$ 
kN/mm and $l_0 = 0.0625$. 
To obtain the crack path, a constant displacement increment of $\Delta w$ = $1\times 10^{-3}$ mm has been applied.

\begin{figure}
    \centering
     \subfigure[]{
    \includegraphics[width = 0.25\textwidth]{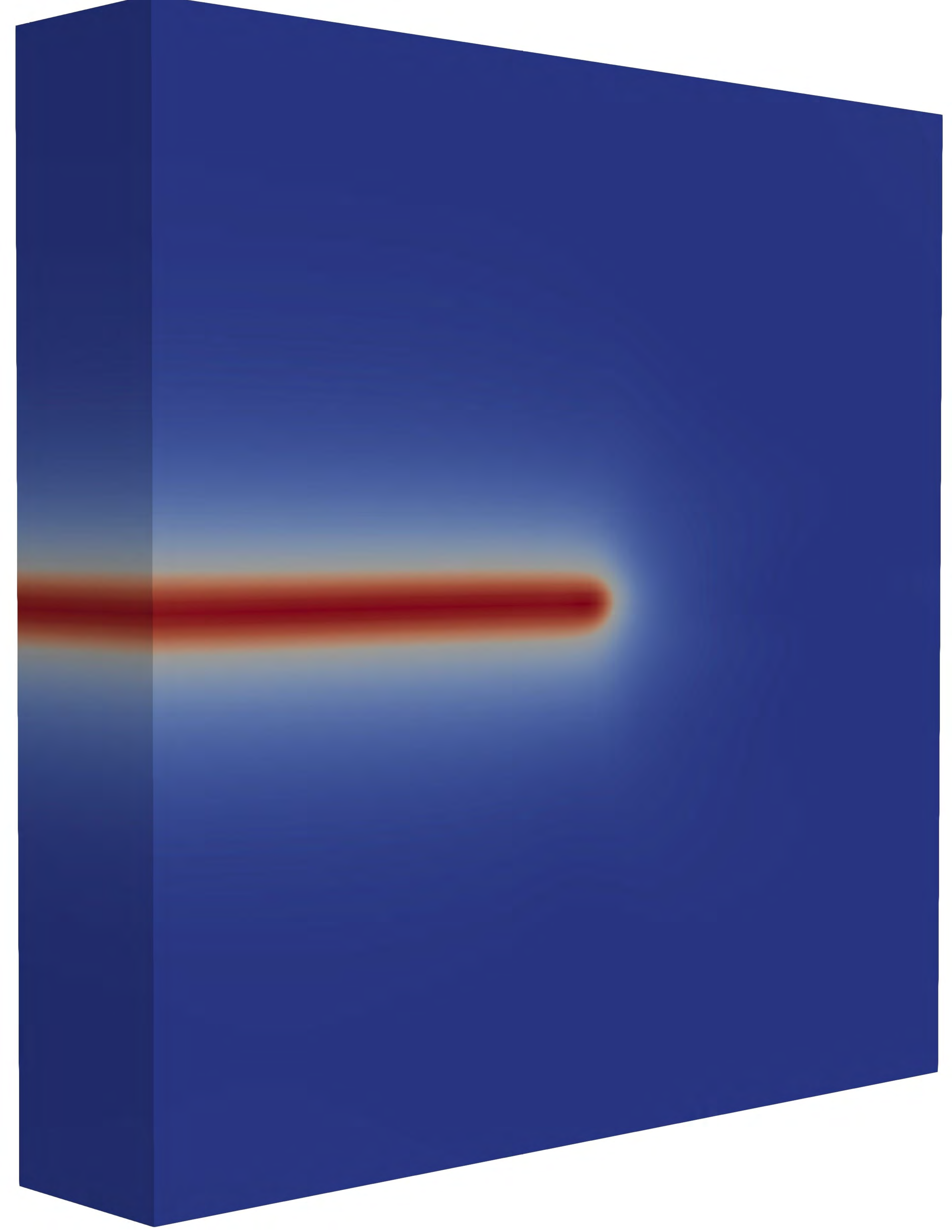}}
     \subfigure[]{
    \includegraphics[width = 0.23\textwidth]{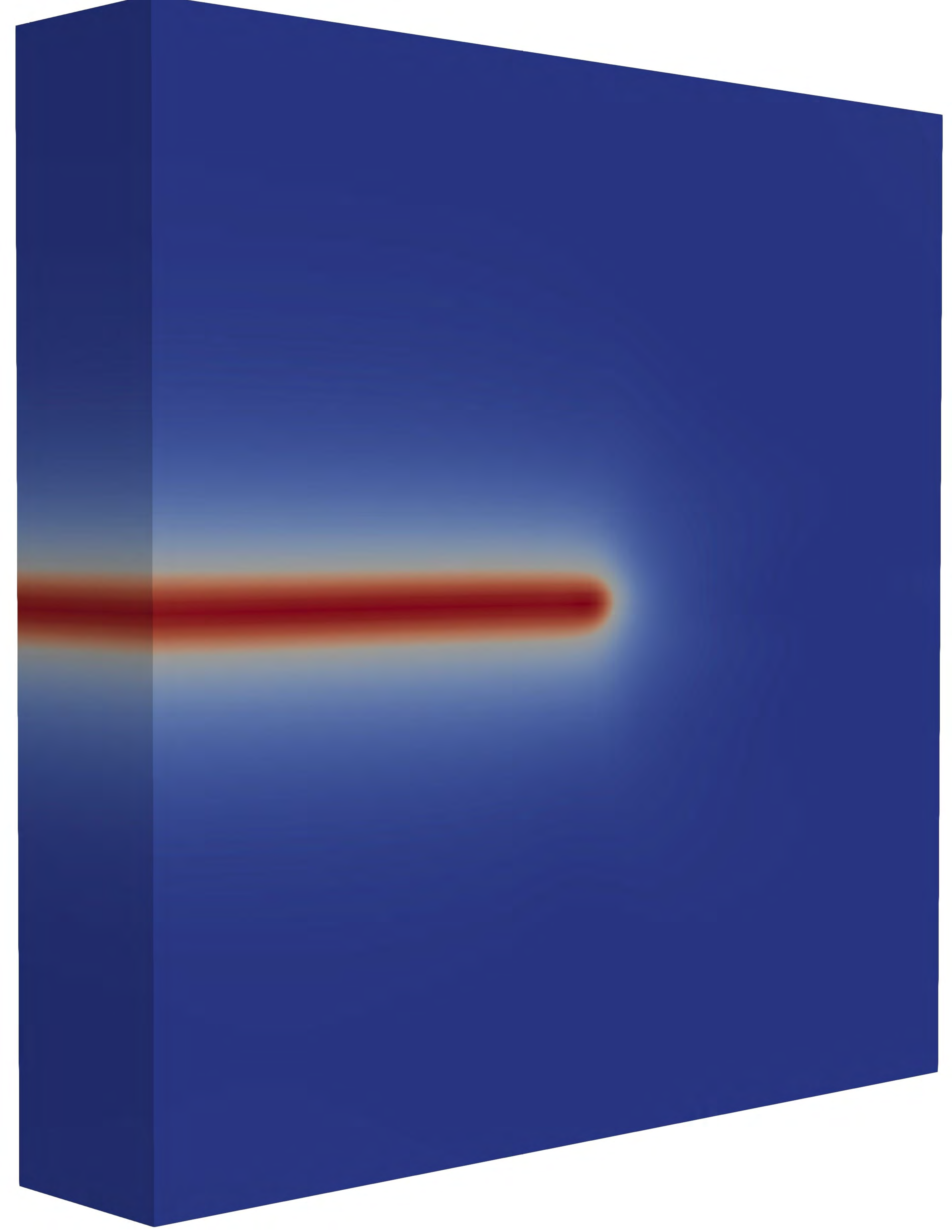}}
    \subfigure[]{
    \includegraphics[width = 0.23\textwidth]{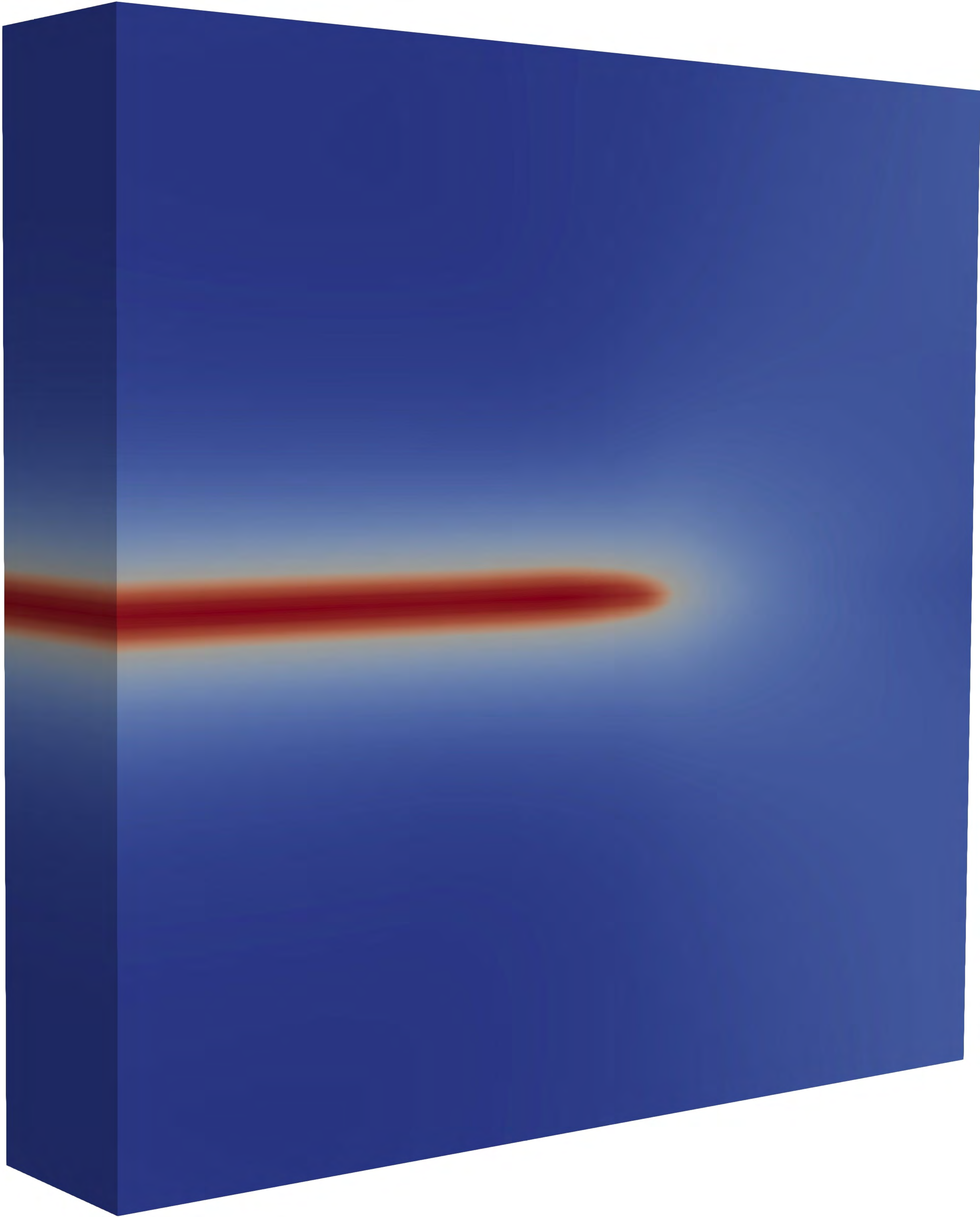}}
     \subfigure[]{
    \includegraphics[width = 0.22\textwidth]{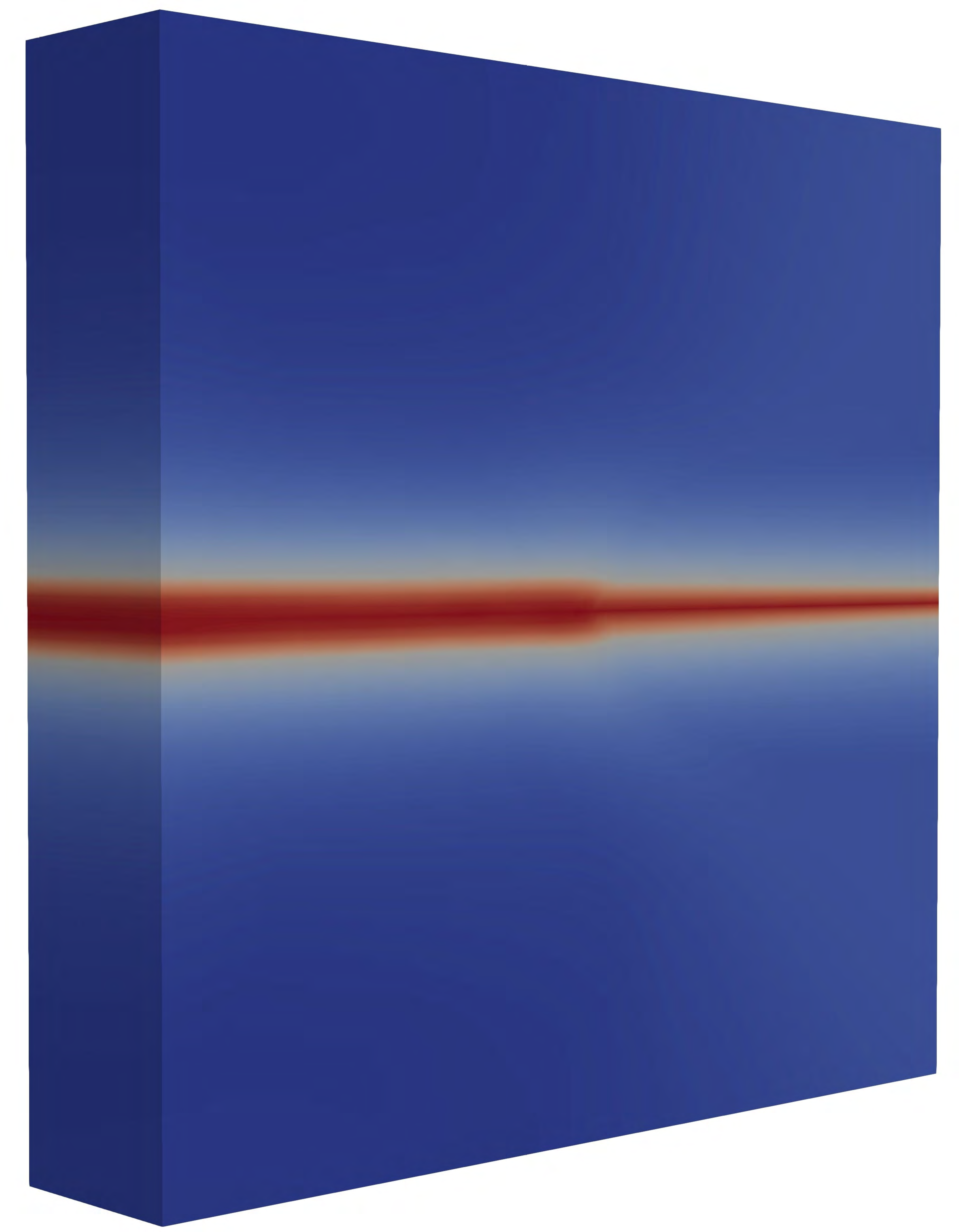}}
    \subfigure{
    \includegraphics[width = 0.4\textwidth]{phiColorBar.pdf}}
    \caption{Plots of cube showing the predicted phase-field for prescribed displacement of (a) initial crack (without any applied displacement), (b) $1\times10^{-3}$, (c) $3\times10^{-3}$, and (d) $6\times10^{-3}$. Plot (a) shows the initialization of the crack.}
    \label{fig:3D_predPhi}
\end{figure}

Similarly to the previous examples, we have considered a fully connected deep neural network with 3 hidden layers, comprising of 50 neurons in each hidden layer.
For the first two layers, we have considered the \texttt{tanh} activation function; 
whereas for the last layer, linear activation
function has been considered.
The overall problem domain has been divided
in to 512 elements with each element having
64 Gauss points.
In this example too, the crack is initiated 
using the strain history functional. The 
Dirichlet boundary conditions for the problem
are:
\begin{equation}
    u(x,y,0) = v(x,y,0) = w(x,y,0) = 0, \;\;\; w(x,y,1)= \Delta w, 
\end{equation}
where $u$, $v$ and $w$ are the solutions of the elastic field in \textit{x}, \textit{y}, and \textit{z}-axis, respectively. To ensure that the boundary conditions are exactly satisfied, we have set
\begin{equation}
\begin{split}
    u &= z\hat{u},\\
    v &= z\hat{v}, \\
    w & = z(z-1)\hat{w} + z \Delta u,
\end{split}
\end{equation}
where $\hat{u}$, $\hat{v}$ and $\hat{w}$ are obtained from the neural network.
The phase-field at certain displacement steps obtained using the proposed PINN approach is shown in \autoref{fig:3D_predPhi}.
The crack propagation pattern is found to be similar to those reported in literature \cite{natarajan2019fenics, molnar20172d}.
We also note that while \cite{natarajan2019fenics}
obtained results with 134,567 elements,
the proposed PINN require only 512 elements.
Lastly, we observe that the crack propagation pattern observed for this example is similar to 
the 2D problem in problem 2.
This is expected as in both the cases, there exists a single-edge notched crack and the plate/cube is subjected to a tensile load.
Overall, the results obtained illustrate that the proposed approach can solve 3D problems as well.

\section{Conclusions and future work}
\label{sec:conclusion}
In this work, we have proposed a new physics 
informed neural network (PINN) algorithm for predicting the 
crack path using the phase-field approach.
Unlike most of the PINN algorithms available 
in literature, we propose to
utilize the variational energy of the 
system as the loss function.
We argue that compared to the residual based loss functions commonly available in the literature,
the variational energy based loss function is easier to minimize and hence, the proposed PINN performs better.
Moreover, we modify the neural network outputs in such a way that the boundary conditions are exactly satisfied and hence, no boundary loss component is present within the loss function.
In order to compute the total variational energy of the system in an efficient manner,
we propose to utilize NURBS and Gauss-Legendre rule.
While NURBS is used for building the problem domain/geometry, 
Gauss-Legendre rule is used for computing the total variational energy by numerical integration.
For efficient numerical integration of the fracture zone, we propose to discretize the problem domain into a number of elements and then generate Gauss points within each element. Moreover, we  utilize the concept of `transfer learning' wherein the network is retrained partially and hence, the computational cost is significantly reduced.

The proposed approach is applied for solving four fracture mechanics examples.
For all the examples, we observe that the 
results obtained using the proposed approach 
match closely with results from the literature.
For the first two examples, we perform a comparative 
assessment between the proposed approach and the  conventional residual based PINN.
For both the problems, the proposed approach is found to be more accurate.
The efficiency gained by using the Gauss-Legendre based integration scheme has also been illustrated for the first two examples.

Despite the excellent performance of the proposed approach for the four problems presented,
it is important to note that the proposed approach is at its early stages and hence,
has certain limitations.
First and foremost, compared to classical computational mechanics techniques like finite element analysis, the advantage of the proposed approach resides in its efficiency. 
This is indicative by the fact that the number of discretized elements required in the proposed approach is significantly smaller as compared to finite element based phase field methods available in the literature.
Secondly, in the current study we have pre-refined the expected path of crack growth.
It will be more useful if the refinement can be carried out in an adaptive manner.
In future works, we will address some of these issues.

The potential application of the proposed approach is to be used as a low-fidelity surrogate  for the high fidelity numerical solvers.
Since the proposed approach is directly trained from the physics of the problem, the high-fidelity solver is not required for training the proposed model.
Such a surrogate will be extremely useful in domains such as  reliability analysis, uncertainty quantification and design optimization.

\section*{Acknowledgement}
SG acknowledges the support of the German Academic Exchange Service (DAAD).
SC acknowledges the support of XSEDE (grant no. DMR180088) and Center for Research Computing, University of Notre Dame for providing computational resources required for carrying out this work.

\bibliographystyle{unsrt}



\end{document}